\documentclass{article}

\PassOptionsToPackage{numbers, compress}{natbib}
\usepackage[final]{neurips_2022}
\usepackage[utf8]{inputenc} % allow utf-8 input
\usepackage[T1]{fontenc}    % use 8-bit T1 fonts
\usepackage{hyperref}       % hyperlinks
\usepackage{url}            % simple URL typesetting
\usepackage{booktabs}       % professional-quality tables
\usepackage{amsfonts}       % blackboard math symbols
\usepackage{nicefrac}       % compact symbols for 1/2, etc.
\usepackage{microtype}      % microtypography
\usepackage{xcolor}         % colors
\usepackage{amssymb}
\usepackage{amsmath}
\usepackage{algorithm}
\usepackage{algpseudocode} 
\usepackage{graphicx}
\usepackage{subfigure}
\usepackage{wrapfig}
\usepackage{makecell}
\usepackage{mathtools}
\usepackage{amsthm}
\usepackage{array}
\usepackage{diagbox}
\usepackage{makecell}
\usepackage{booktabs}

\theoremstyle{plain}
\newtheorem{theorem}{Theorem}[section]

\theoremstyle{definition}

\theoremstyle{remark}
\newtheorem{remark}[theorem]{Remark}

\title{Scalable Multi-agent Covering Option Discovery \\ based on Kronecker Graphs}

\author{%
  Jiayu Chen \\
  Purdue University\\
  West Lafayette, IN 47907\\
  \texttt{chen3686@purdue.edu} \\
   \And
  Jingdi Chen \\
  The George Washington University\\
  Washington, DC 20052\\
  \texttt{jingdic@gwu.edu} \\
   \And
  Tian Lan \\
  The George Washington University\\
  Washington, DC 20052\\
  \texttt{tlan@gwu.edu} \\
   \And
  Vaneet Aggarwal \\
  Purdue University\\
  West Lafayette, IN 47907\\
  \texttt{vaneet@purdue.edu} \\
}

\begin{document}

\maketitle

\begin{abstract}
Covering skill (a.k.a., option) discovery has been developed to improve the exploration of RL in single-agent scenarios with sparse reward signals, through connecting the most distant states in the embedding space provided by the Fiedler vector of the state transition graph. Given that joint state space grows exponentially with the number of agents in multi-agent systems, existing researches still relying on single-agent skill discovery either become prohibitive or fail to directly discover joint skills that improve the connectivity of the joint state space. In this paper, we propose multi-agent skill discovery which enables the ease of decomposition. Our key idea is to approximate the joint state space as a Kronecker graph, based on which we can directly estimate its Fiedler vector using the Laplacian spectrum of individual agents' transition graphs. Further, considering that directly computing the Laplacian spectrum is intractable for tasks with infinite-scale state spaces, we further propose a deep learning extension of our method by estimating eigenfunctions through NN-based representation learning techniques. The evaluation on multi-agent tasks built with simulators like Mujoco, shows that the proposed algorithm can successfully identify multi-agent skills, and significantly outperforms the state-of-the-art. Codes are available at: \href{https://github.itap.purdue.edu/Clan-labs/Scalable_MAOD_via_KP}{https://github.itap.purdue.edu/Clan-labs/Scalable\_MAOD\_via\_KP}.
\end{abstract}

\section{Introduction}

Option discovery \cite{DBLP:journals/ai/SuttonPS99} enables temporally-abstract actions to be constructed and utilized in RL, which can significantly improve the performance of RL agents. Among recent developments, \textit{Covering Option Discovery} \cite{DBLP:conf/icml/JinnaiAHLK19, DBLP:conf/icml/JinnaiPAK19} has been shown to be an effective approach to improve the exploration in environments with sparse reward signals. In particular, it first computes the second smallest eigenvalue $\lambda_{2}$ and the corresponding eigenvector $F$ (i.e., Fiedler vector \cite{fiedler1973algebraic}) of the Laplacian matrix extracted from the state transition process in RL. Then, options are built to connect the states corresponding to the minimum or maximum in $F$, which greedily improves the algebraic connectivity of the state space \cite{fast_graphs} and accelerate exploration within it. In this paper, we consider constructing and utilizing covering options in MARL. Due to the exponentially-large state space in multi-agent scenarios, a commonly-adopted way to solve this problem \cite{DBLP:conf/atal/AmatoKK14, amato2019modeling, DBLP:conf/atal/ChakravortyWRCB20, DBLP:conf/iclr/LeeYL20} is to construct individual options as if in a single-agent environment first, and then learn to collectively leverage these individual options to tackle multi-agent tasks. This method fails to consider the coordination among agents in the option discovery process, and thus can suffer from poor behavior in multi-agent collaborative tasks. 

To this end, we propose a framework that makes novel use of Kronecker graphs to directly construct multi-agent options and adopt them to accelerate the joint exploration of agents in MARL. In particular, the joint state space is approximated as the Kronecker product of the state spaces of individual agents. Then, based on the theorem we propose, we can directly estimate the Fiedler vector of the joint state space using the Laplacian spectrum of individual ones. Next, the joint options can be constructed to explore the joint states corresponding to the minimum or maximum in the Fiedler vector, resulting in a greedy improvement of the joint state space's connectivity. However, calculating the Laplacian spectrum can be expensive in a matrix base. As a scalable extension, we leverage a deep neural network approximation to learn the Laplacian spectrum so that our method can be adopted to tasks with infinite-scale state spaces. Our main contributions are: (1) We propose \textit{Multi-agent Covering Option Discovery} -- a decentralized manner to discover multi-agent options which can greatly aid the joint exploration of agents in MARL. (2) We propose that multi-agent options can be adopted to MARL in either a decentralized or centralized manner, which means that agents can either jointly decide on their options, or choose their options independently and select different options to execute simultaneously. (3) We scale \textit{Multi-agent Covering Option Discovery} to infinite-scale state spaces using SOTA representation learning techniques, and propose that the discovered options can be adopted with deep MARL methods for continuous tasks. (4) We empirically show that agents using our multi-agent options significantly outperform agents with single-agent options or no options.

\section{Related Work} \label{related}

\noindent \textbf{Option Discovery: }The option framework was proposed in \cite{DBLP:journals/ai/SuttonPS99}, which extends the usual notion of actions to include options — the closed-loop policies for taking actions over a period of time. A lot of option discovery algorithms have been proposed. Some of them are based on task-related reward signals \cite{DBLP:conf/icml/McGovernB01, Menache02q-cut-, DBLP:conf/nips/MankowitzMM16, Harb2018WhenWI}. Specifically, they directly define or learn the options through gradient descent that can lead the agent to the rewarding states, and then utilize these trajectory segments (options) to compose the completed trajectory toward the goal state. These methods rely on dense reward signals, which are hard to acquire in real-life tasks. Other works define the sub-goal states (termination states of the options) based on the visitation frequency of the states. For example, in \cite{stolle2002learning, DBLP:conf/icml/SimsekWB05, DBLP:conf/nips/SimsekB08}, they discover the options by recognizing the bottleneck states in the environment, through which the agent can transfer between the sub-areas that are loosely connected in the state space, and they define these options as betweenness options. 
Recently, there are some state-of-the-art option generation methods based on the Laplacian spectrum of the state-transition graph, such as \cite{DBLP:conf/icml/JinnaiAHLK19, DBLP:conf/icml/JinnaiPAK19, DBLP:journals/corr/MachadoBB17, DBLP:conf/iclr/MachadoRGLTC18, DBLP:conf/iclr/JinnaiPMK20}, since the eigenvectors of the Laplacian of the state space can provide embeddings in lower-dimensional space, based on which we can obtain good measurements of the accessibility from one state to another. Especially, in \cite{DBLP:conf/icml/JinnaiPAK19, DBLP:conf/iclr/JinnaiPMK20}, they propose covering options and prove that their option generation method has higher exploration speed and better performance compared with the previous option discovery methods. Note that all the approaches mentioned above are for single-agent scenarios, we will extend the construction and adoption of covering options to MARL in this paper.

\noindent\textbf{Adopting options in multi-agent scenarios: }Most of the researches on adopting options in MARL \cite{DBLP:conf/atal/AmatoKK14, amato2019modeling, DBLP:conf/atal/ChakravortyWRCB20, DBLP:conf/iclr/LeeYL20, shen2006multi, DBLP:conf/atal/YangBZ20} try to first learn the options for each agent with the option discovery methods we mentioned above, and then learn to collaboratively utilize these individual options. Therefore, the options they use are still single-agent options, and the coordination among agents can not be utilized in the option discovery process. In this paper, we propose directly constructing multi-agent covering options of the joint state transition graphs based on Laplacian spectrum of the individual ones, and explore how to utilize the multi-agent options in MARL tasks effectively. Further, to scale our algorithm, we extend a tabular algorithm considered in \cite{9847387}, adopt SOTA representation learning techniques to obtain the Laplacian spectrum for option discovery, and integrate the discovered options with deep MARL algorithms, to aid the learning performance in scenarios with infinite-scale state spaces, which is testified on multiple complex Grid and Mujoco tasks.

% In \cite{9847387}, we propose a primary algorithm for multi-agent option discovery based on tabular reinforcement learning.

\section{Background} \label{background}

\textbf{Kronecker Product of Graphs: }Let $G_1=(V_{G_1}, E_{G_1})$ and $G_2=(V_{G_2}, E_{G_2})$ be two state transition graphs (defined in Appendix \ref{notation}), corresponding to the individual state space $\mathcal{S}_{1}$ and $\mathcal{S}_{2}$. The Kronecker product of them denoted by $G_1 \otimes G_2$ is a graph defined on the set of vertices $V_{G_1} \times V_{G_2}$, such that \cite{weichsel1962kronecker}: \textit{Two vertices of $G_1 \otimes G_2$, namely $(g,h)$ and $(g',h')$, are adjacent if and only if $g$ and $g'$ are adjacent in $G_1$ and $h$ and $h'$ are adjacent in $G_2$.} Thus, the Kronecker graph can capture the joint transitions of agents in their joint state space very well. In Section \ref{theory}, we propose to use the Kronecker graph as an effective approximation of the joint state transition graph, so that we can discover the joint options based on the factor graphs. Further, $A_1 \otimes A_2$ is an $|\mathcal{S}_{1}||\mathcal{S}_{2}| \times |\mathcal{S}_{1}||\mathcal{S}_{2}|$ matrix with elements defined by $(A_1 \otimes A_2){(I,J)} =  A_1{(i,j)}A_2{(k,l)}$ with Equation (\ref{equ:-1}), where $A_1$ and $A_2$ are the adjacency matrices of $G_1$ and $G_2$, $A_1{(i,j)}$ is the element lies on the $i$-th row and $j$-th column of $A_1$ (indexed from 1).
\begin{equation} \label{equ:-1}
    \setlength{\abovedisplayskip}{1pt}
    \setlength{\belowdisplayskip}{1pt}
    \begin{aligned}
        I = (i-1) \times |\mathcal{S}_{2}| + k, \ \ 
        J = (j-1) \times |\mathcal{S}_{2}|+ l
    \end{aligned}
\end{equation}

\textbf{Covering Option Discovery: } As defined in \cite{DBLP:journals/ai/SuttonPS99}, an option $\omega$ consists of three components: an intra-option policy $\pi_{\omega}: \mathcal{S} \text{ x } \mathcal{A} \rightarrow [0,1]$, a termination condition $ \beta_{\omega}: \mathcal{S} \rightarrow \{0,1\}$, and an initiation set $I_{\omega} \subseteq \mathcal{S}$. An option $<I_{\omega}, \pi_{\omega}, \beta_{\omega}>$ is available in state $s$ if and only if $s \in I_{\omega}$. If the option $\omega$ is taken, actions are selected according to $\pi_{\omega}$ until $\omega$ terminates stochastically according to $\beta_{\omega}$ (i.e., $\beta_{\omega}=1$). Recently, the authors of \cite{DBLP:conf/icml/JinnaiPAK19} proposed \textit{Covering Option Discovery} -- discovering options by minimizing the upper bound of the expected cover time of the state space. First, they compute the Fiedler vector $F$ of the Laplacian matrix of the state transition graph. Then, they collect the states $s_{i}$ and $s_{j}$ with the largest $(F_i-F_j)^2$ ($F_i$ is the $i$-th element in $F$), based on which they construct two symmetric options: $\omega_{ij}=<I_{\omega_{ij}}=\{s_{i}\},\pi_{\omega_{ij}},\beta_{\omega_{ij}}=\{s_{j}\}>, \omega_{ji}=\ <I_{\omega_{ji}}=\{s_{j}\},\  \pi_{\omega_{ji}}, \ \beta_{\omega_{ji}}=\{s_{i}\}>$ to connect these two subgoal states, where $\pi_{\omega_{ij}}$ is defined as the optimal path from $s_i$ to $s_j$. 
% This whole process is repeated until they get the required number of options. 

The intuition of this method is as follows. The authors of \cite{fast_graphs} prove that $(F_i-F_j)^2$ gives the first order approximation of the increase in $\lambda_{2}(L)$ (i.e., algebraic connectivity) by connecting $(s_i, s_j)$, and propose a greedy heuristic to improve the algebraic connectivity of a graph: adding a certain number of edges one at a time, and each time connecting $(s_i, s_j)$ corresponding to the largest $(F_i-F_j)^2$. Further, in \cite{DBLP:conf/icml/JinnaiPAK19}, they prove that the larger the algebraic connectivity is, the smaller the upper bound of the expected cover time would be and the easier the exploration tends to be. Therefore, applying this greedy heuristic to the state transition graph can effectively improve the exploration in the state space.

\section{Proposed Algorithm} \label{alg}
\subsection{System Model} \label{example}

In this paper, we consider to compute covering options in multi-agent scenarios, with $n$ being the number of agents, $\widetilde{\mathcal{S}}=\mathcal{S}_1 \times \mathcal{S}_2 \times \cdots \times \mathcal{S}_n$ being the set of joint states, $\widetilde{\mathcal{A}}=\mathcal{A}_1 \times \mathcal{A}_2 \times \cdots \times \mathcal{A}_n$ being the set of joint actions, $\mathcal{S}_i$ and $\mathcal{A}_i$ being the individual state space and action space of agent $i$. The size of the joint state space grows exponentially with the number of agents. Thus, it is prohibitive to directly compute the covering options based on the joint state transition graph using the approach introduced in Section \ref{background} for a large $n$. 

A natural method to tackle this problem is to compute the options for each agent by considering only its own state transitions, and then learn to collaboratively leverage these individual options. However, it fails to directly recognize joint options composed of multiple agents' temporal action sequences for encouraging the joint exploration of all the agents. The algebraic connectivity of the joint state space may not be improved with these single-agent options. We illustrate this with a simple example.

\begin{figure}[t]
	\centering
	\subfigure[Joint state transition graph of agent 1 and 2]{
	\label{fig:-1(a)}
 \includegraphics[height=0.6in, width=2.6in]{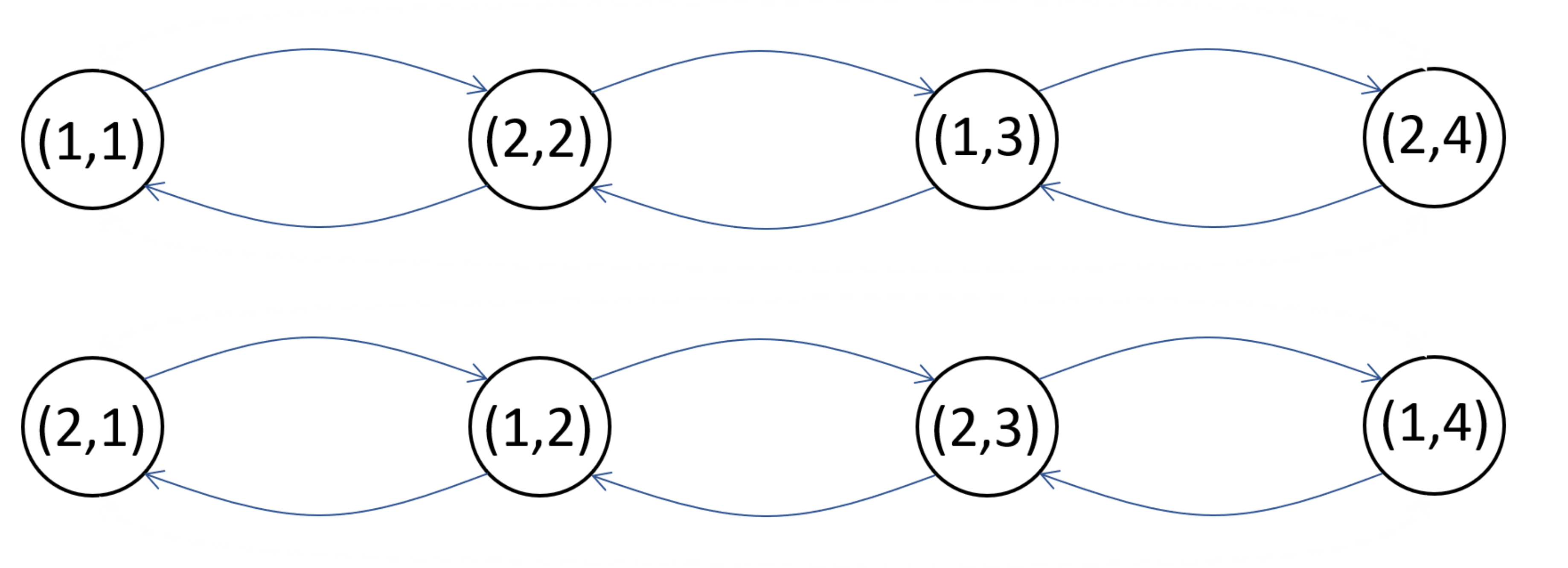}}
	\subfigure[Joint state transition graph with individual options]{
	\label{fig:-1(b)}
\includegraphics[height=0.6in, width=2.6in]{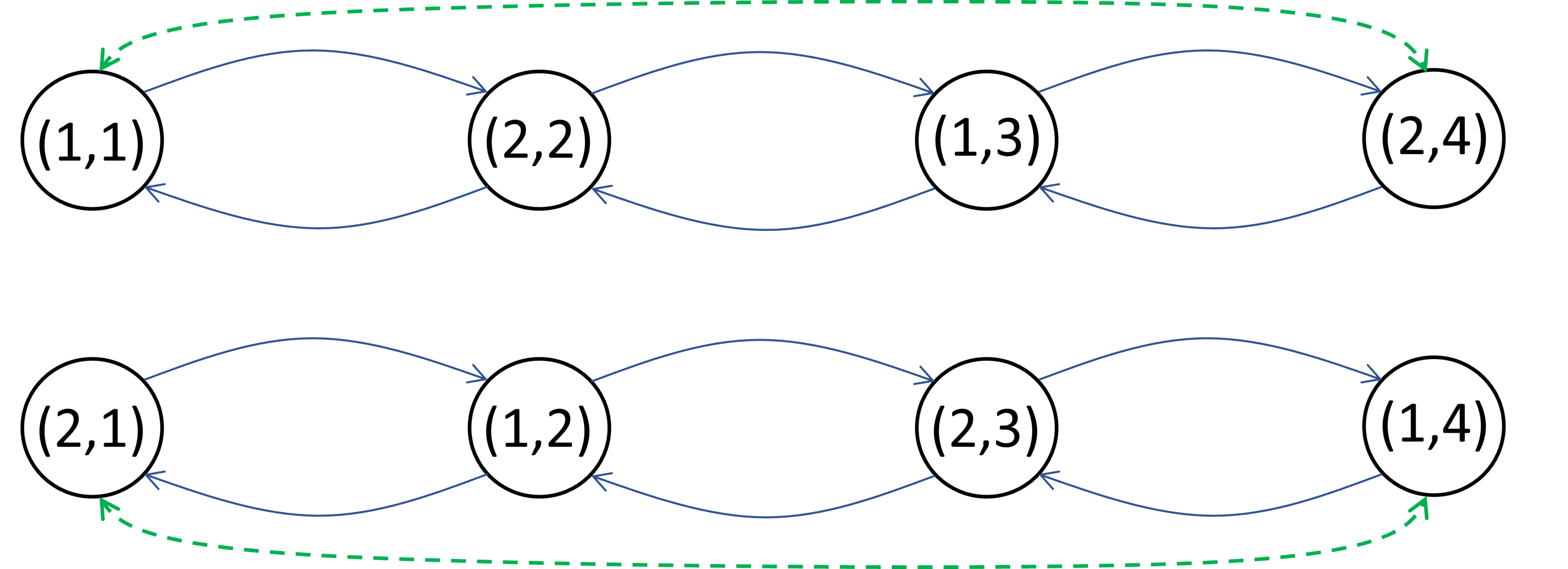}}
	\caption{An illustrative example: limitations of utilizing single-agent options alone for MARL.}
	\label{fig:-1}

\end{figure}

\textbf{Illustrative example: }Figure \ref{fig:-1(a)} shows a joint state transition graph $\widetilde{G}$ of two agents, where $\mathcal{S}_1=\{1,2\}$ and $\mathcal{S}_2=\{1,2,3,4\}$. In order to compute the individual options, we can restrict our attention to the state transition graph of each agent, namely $G_1$ and $G_2$. The Fiedler vectors (not normalized) of $G_1$ and $G_2$ are: $v^{G_1}=[-1, 1], v^{G_2}=[-1, -\sqrt{2}+1, \sqrt{2}-1, 1]$. Then, according to the option discovery approach described in Section \ref{background}, we can get the individual options for agent 1 to connect its state 1 (minimum) and state 2 (maximum), and individual options for agent 2 to connect its state 1 and state 4. With these options, the updated joint state space is Figure \ref{fig:-1(b)}. The straightforward decomposition of option discovery for MARL fails to create a connected graph. Thus, utilizing single-agent options alone may not be sufficient for efficient exploration.

Therefore, we propose to build Multi-agent Covering Options to enhance the connectivity of the joint state space. We can represent it as a tuple: $<I_{\omega}, \pi_{\omega}, \beta_{\omega}>$, where $I_{\omega} \subseteq \widetilde{\mathcal{S}}$ is the set of initiation joint states, $ \beta_{\omega}: \widetilde{\mathcal{S}} \rightarrow \{0,1\}$ indicates the joint states to terminate, $\pi_{\omega}=(\pi_{\omega}^{1}, \cdots, \pi_{\omega}^{n}) (\pi_{\omega}^{i}: \mathcal{S}_i \times \mathcal{A}_i \rightarrow [0,1]),$ is the joint intra-option policy that can lead the agents from the initiation states to the termination states. The key challenge is to calculate the Fiedler vector of the joint state space according to which we can define $<I_{\omega}, \pi_{\omega}, \beta_{\omega}>$. Given that $|\widetilde{\mathcal{S}}|$ grows exponentially with $n$, we propose to estimate the joint Fiedler vector based on the individual state spaces in the next section.

\subsection{Theory results} \label{theory}

% This section shows the theoretical foundations of our approach. 
We propose to use the Kronecker graph to decompose the calculation of the joint Fielder vector to single-agent state spaces, because: (1) the Kronecker product of individual state transition graphs $\otimes_{i=1}^{n}G_{i}=G_1\otimes\cdots\otimes G_n$ provides a good approximation of the joint state transition graph $\widetilde{G}$; (2) The Fielder vector of $\otimes_{i=1}^{n}G_{i}$ can be estimated with its factor graphs $G_i(i=1,\cdots,n)$.

The use of $\otimes_{i=1}^{n}G_{i}$ as a factorized approximation of $\widetilde{G}$ introduces noise, since $\widetilde{G}=\otimes_{i=1}^{n}G_{i}$ becomes exact only in the case where agents' transitions are not influenced by the others. However, for the purpose of option discovery, we only need to identify areas in the state space with relatively low or high values in the Fielder vector, so an exact calculation of $\widetilde{G}$ and its Fiedler vector is not necessary. Moreover, the state transition influence among agents, e.g., collisions and blocking, would most likely result in local perturbations of the transition graph and thus is inconsequential to global properties of $\widetilde{G}$ (e.g., the algebraic connectivity), especially for large-scale state spaces. Therefore, approximating $\widetilde{G}$ by $\otimes_{i=1}^{n}G_{i}$ allows efficient option discovery.
%that $\otimes_{i=1}^{n}G_{i}$ can be used as an approximation of $\widetilde{G}$ for computing multi-agent options and $\widetilde{G}=\otimes_{i=1}^{n}G_{i}$ becomes exact if the state transitions of an agent would not be influenced by the others.  Given that we only need to identify areas in the state space with relatively low or high values in the Fielder vector and connect them with options, exact calculation of $\widetilde{G}$ and its Fiedler vector is not necessary. Also, the state transition influence among agents most likely would result in local perturbations, e.g., collisions, in $\widetilde{G}$ compared with $\otimes_{i=1}^{n}G_{i}$, and these local noise will not greatly impact the global property of $\widetilde{G}$ especially for large-scale state spaces. Further, w
We empirically show in Figure~\ref{fig:5} that superior exploration can still be achieved under such approximation noise, even if the transition influence is heavy (affecting 63\% of the episode length), numerically validating the robustness of our proposed approach to the approximation error. We further provide a quantitative study on the approximation error in Appendix \ref{QS}. Next, we show how to effectively approximate the Fiedler vector of $\otimes_{i=1}^{n}G_{i}$ based on the Laplacian spectrum of the factor graphs, to achieve an effective decomposition. Inspired by \cite{basic2021estimation} that proposed an estimation of the Laplacian spectrum of the Kronecker product of two factor graphs, we prove Theorem \ref{thm:2} for an arbitrary number of factor graphs.
% We note that to discover multi-agent options, we only need to identify the most distant states in the joint state space which are corresponding to the minimum and maximum of its Fiedler vector. Thus, we do not need to get the exact joint state transition graph $\widetilde{G}$ or the exact value of its Fiedler vector.

\begin{figure}[t]
	\centering
	\subfigure[Joint state transition graph updated with option $\omega_1$]{
	\label{fig:-3(a)}
 \includegraphics[height=0.7in, width=2.6in]{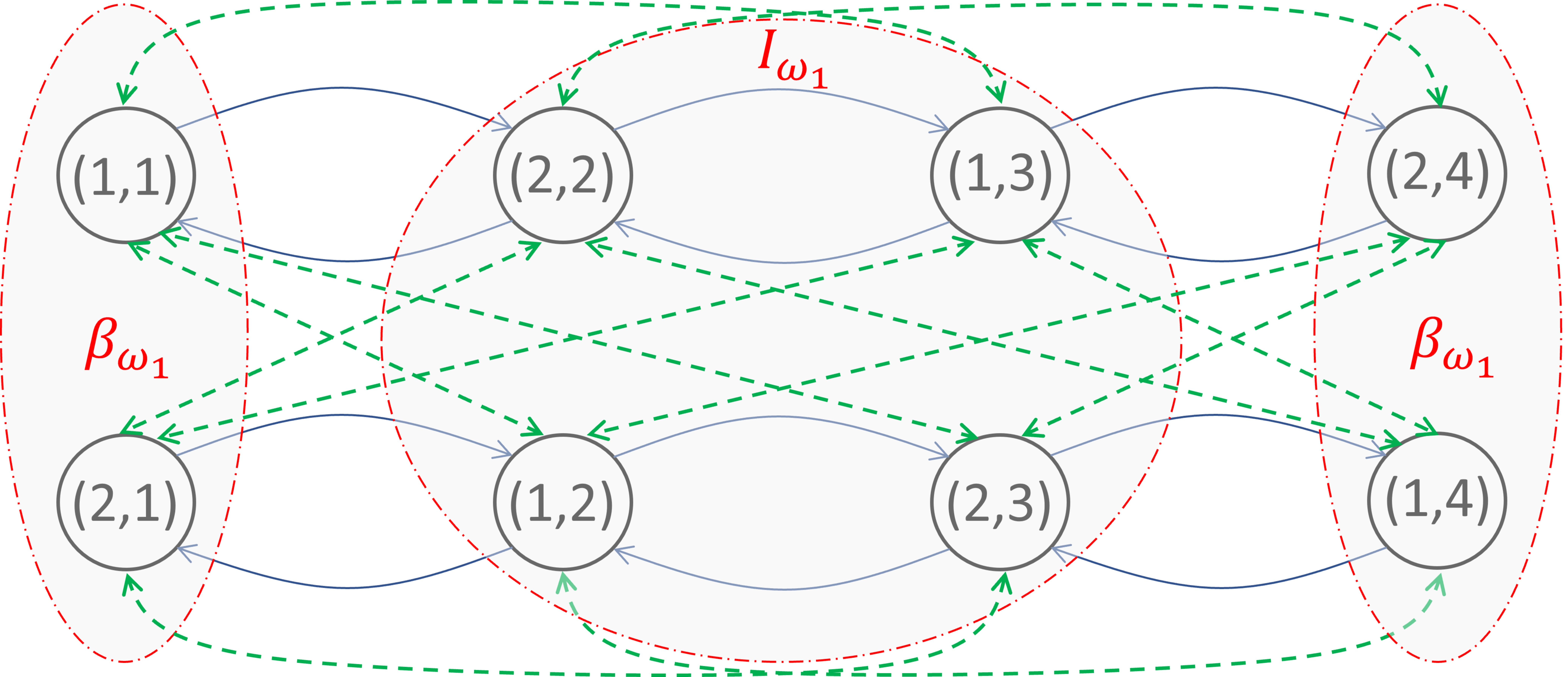}}
	\subfigure[Joint state transition graph updated with option $\omega_2$]{
	\label{fig:-3(b)}
 \includegraphics[height=0.7in, width=2.6in]{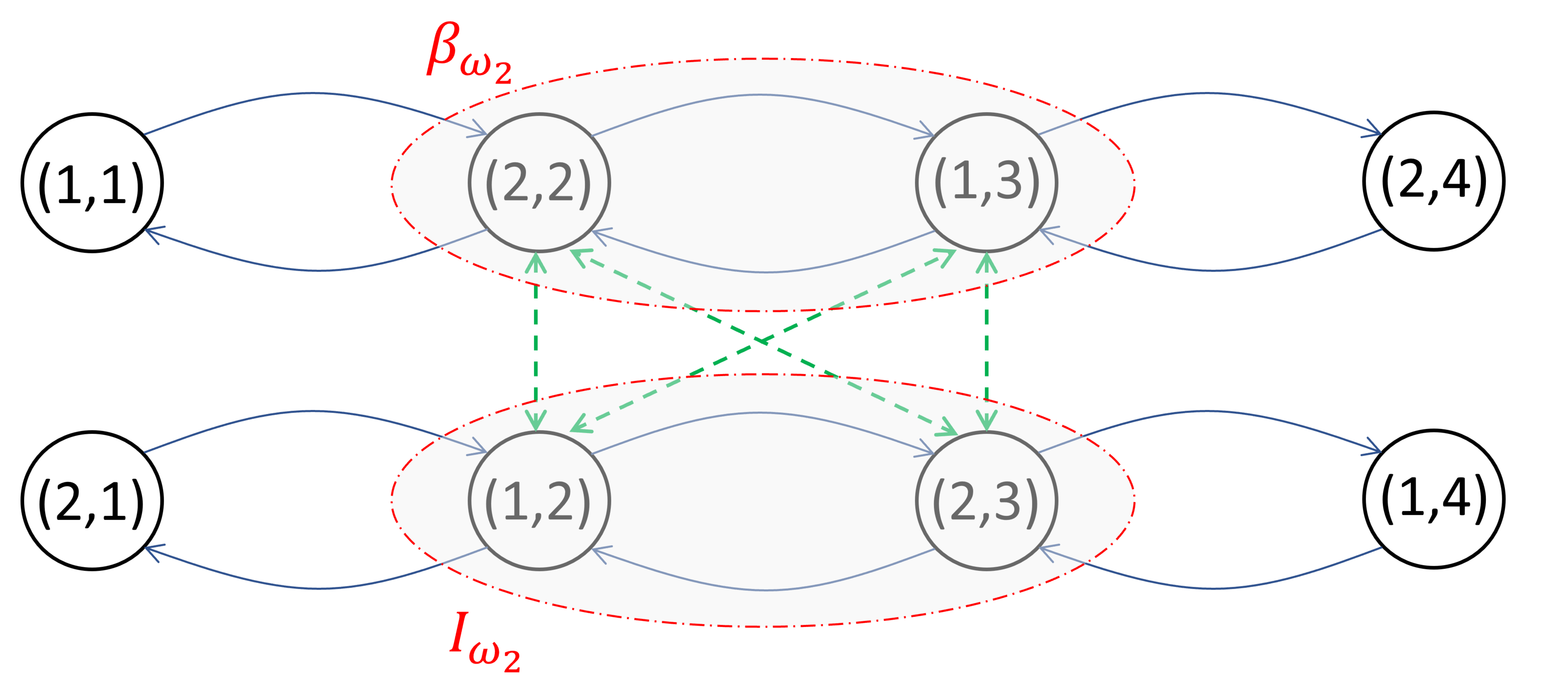}}

	\caption{The joint state transition graph updated with the detected multi-agent options}

	\label{fig:-3}

\end{figure}

\begin{theorem}
\label{thm:2}

For graph $\widetilde{G}=\otimes_{i=1}^{n}G_{i}$, we can approximate the eigenvalues $\mu$ and eigenvectors $v$ of its Laplacian $L$ by:
\begin{equation} \label{equ:mu}
    \setlength{\abovedisplayskip}{1pt}
    \setlength{\belowdisplayskip}{1pt}
    \begin{aligned}
    \mu_{k_1,\ldots,k_n} =\left\{\left[1-\prod_{i=1}^{n}(1-\lambda_{k_i}^{G_i})\right] \prod_{i=1}^{n}d_{k_i}^{G_i}\right\}, \ v_{k_1,\ldots,k_n}=\otimes_{i=1}^{n}v_{k_i}^{G_i}
    \end{aligned}
\end{equation}

% \begin{equation} \label{equ:v}
%     \setlength{\abovedisplayskip}{1pt}
%     \setlength{\belowdisplayskip}{1pt}
%     v_{k_1,\ldots,k_n}=\otimes_{i=1}^{n}v_{k_i}^{G_i}
% \end{equation}
where $\lambda_{k_i}^{G_i}$ and $v_{k_i}^{G_i}$ are the $k_i$-th smallest eigenvalue and corresponding eigenvector of $\mathcal{L}_{G_i}$ (normalized Laplacian of $G_i$), $d_{k_i}^{G_i}$ is the $k_i$-th smallest diagonal entry of $D_{G_{i}}$ (degree matrix of $G_i$).
\end{theorem}

The proof of Theorem \ref{thm:2} is provided in Appendix \ref{thmproof}. Through enumerating $(k_1, \cdots, k_n)$, we can collect the eigenvalues  and corresponding eigenvectors of $\otimes_{i=1}^{n}G_{i}$ by Equation (\ref{equ:mu}). Then, the eigenvector $v_{\hat{k}_1,\cdots,\hat{k}_n}$ corresponding to the second smallest eigenvalue $\mu_{\hat{k}_1,\cdots,\hat{k}_n}$ is the estimated joint Fiedler vector $F_{\widetilde{G}}$. We can define the joint states corresponding to the maximum or minimum in $F_{\widetilde{G}}$ as the initiation or termination joint states. As discussed in Section \ref{background}, connecting these two joint states with options can greedily improve the algebraic connectivity of the joint state space.
%  and accelerate the joint exploration

\textbf{Illustrative example: }Now we reconsider Figure \ref{fig:-1(a)}, where $\widetilde{G} = G_{1} \otimes G_{2}$. We approximate $F_{\widetilde{G}}$ using Theorem \ref{thm:2} and get two approximations (details are in Appendix \ref{example2}):
\begin{equation} \label{v11}
\setlength{\abovedisplayskip}{1pt}
\setlength{\belowdisplayskip}{1pt}
v_{11}=\left[\frac{1}{\sqrt{2}},\ 1,\ 1,\ \frac{1}{\sqrt{2}},\ \frac{1}{\sqrt{2}},\ 1,\ 1,\ \frac{1}{\sqrt{2}}\right]^{T}, \ v_{24}=\left[-\frac{1}{\sqrt{2}},\ 1,\ -1,\ \frac{1}{\sqrt{2}},\ \frac{1}{\sqrt{2}},\ -1,\ 1,\ -\frac{1}{\sqrt{2}}\right]^{T}
\end{equation}
% \begin{equation}
% \setlength{\abovedisplayskip}{1pt}
% \setlength{\belowdisplayskip}{1pt}
% v_{24}=\left[-\frac{1}{\sqrt{2}},\ 1,\ -1,\ \frac{1}{\sqrt{2}},\ \frac{1}{\sqrt{2}},\ -1,\ 1,\ -\frac{1}{\sqrt{2}}\right]^{T}
% \end{equation}
Based on them and the indexing relationship between $\widetilde{G}$ and its factor graphs (Equation (\ref{equ:-1})), we can get two sets of multi-agent options: \{$I_{\omega_1}$=\{(1,2),\ (1,3),\ (2,2),\ (2,3)\}, $\beta_{\omega_1}$=\{(1,1), (1,4),\ (2,1),\ (2,4)\}\} and \{$I_{\omega_2}$=\{(1,2),\ (2,3)\}, $\beta_{\omega_2}$=\{(1,3),\ (2,2)\}\}, where we set the joint states corresponding to the maximum or minimum as the initiation or termination states respectively. For example, in $v_{11}$, the $7$-th element (indexed from 1) is a maximum, so the $7$-th joint state is in $I_{\omega_1}$ and denoted as $(2,3)$ (Equation (\ref{equ:-1})). As shown in Figure \ref{fig:-3}, both of the two options can lead to a connected graph. Thus, the adoption of multi-agent options has the potential to encourage efficient exploration in the joint state space. Next, we will formalize our algorithm.

% , and we can discover multi-agent options based on individual agents' state spaces to enjoy the ease of decomposition.

\subsection{Scalable Multi-agent Covering Option Discovery} \label{smacod}

In this section, we first introduce how to construct multi-agent options based on individual state transition graphs of each agent (represented as adjacency matrices $A_{1:n}$), of which the detailed pseudo codes are in Appendix \ref{pseudo-code}. Then, we propose a scalable extension of our algorithm to infinite-scale state space, by which we don't need to explicitly construct the state transition graph or compute its Laplacian spectrum through eigendecomposition.

First, we need to calculate the estimation of $F_{\widetilde{G}}$ through Theorem \ref{thm:2} based on $A_{1:n}$, and collect the joint states corresponding to the minimum or maximum in $F_{\widetilde{G}}$ as the lists of subgoals $MIN$ and $MAX$. Then, we split each joint state into a list of individual states. For example, we convert ($s_{min}, s_{max}$) into ($(s_{min}^{1}, \cdots, s_{min}^{n})$, $(s_{max}^{1}, \cdots, s_{max}^{n})$), so that we can connect ($s_{min}, s_{max}$) in the joint state space by connecting each ($s_{min}^{i}, s_{max}^{i}$) in the corresponding individual space. After decentralizing the joint states, we can define the multi-agent options as follows: (1) For each option $\omega$, we define $I_{\omega}$ as the explored joint states, and $\beta_{\omega}$ as a joint state in $MIN \cup MAX$ or the unexplored area. Option $\omega$ is available in state $s$ if and only if $s \in I_{\omega}$. Therefore, instead of constructing a point option between ($s_{min}, s_{max}$), e.g., setting \{$s_{min}$\} as $I_{\omega}$ and \{$s_{max}$\} as $\beta_{\omega}$, we extend $I_{\omega}$ to the known area to increase the accessibility of $\omega$. (2) As for the intra-option policy $\pi_{\omega}$ used for connecting the initiation and termination joint state, we divide it into a list of single-agent policies $\pi_{\omega}^{i}\ (i=1,\cdots,n)$, where $\pi_{\omega}^{i}$ can be trained with any single-agent RL algorithm aiming at leading agent $i$ from its own initiation state to the termination state $s_{min}^{i}$ ($s_{max}^{i}$). To sum up, the proposed algorithm first discovers the joint states that need to be explored most, then projects each sub-goal into individual state spaces and train the policy for each agent to visit the projection.
% build multi-agent options to encourage agents to visit these sub-goals. More precisely, we project each sub-goal into the individual state spaces and train the policy for each agent to visit the projection of the sub-goal state in its individual state space.

Here is the computational complexity analysis: Consider an MDP with $n$ agents and $m$ states for each agent. To compute the Fiedler vector, which is the computation bottleneck of the algorithm, directly from the joint state transition graph would require time complexity $\mathcal{O}(m^{3n})$, since there are $m^{n}$ joint states and the time complexity of eigenvalue decomposition is cubic with the number of joint states. While, with our approach, we can decompose the original problem into computing eigenvectors of the individual state transition graphs, of which the overall time complexity is $\mathcal{O}(nm^{3})$. Thus, our solution significantly reduces the problem complexity from $\mathcal{O}(m^{3n})$ to $\mathcal{O}(nm^{3})$. As for problems with large state space (i.e., $m$ is large), our option discovery approach could be directly integrated with sample-based techniques for eigenfunction estimation \cite{DBLP:conf/iclr/WuTN19, DBLP:conf/icml/WangZZSHF21}, which makes it highly-scalable.

\textbf{Extension to tasks with infinite-scale state space.} According to \cite{DBLP:conf/icml/WangZZSHF21}, the $k$ smallest eigenvalues $\lambda_{1:k}$ and corresponding eigenvectors $v_{1:k}$ of the normalized Laplacian $\mathcal{L}$ can be estimated by: 
\begin{equation} \label{opt}
    \setlength{\abovedisplayskip}{1pt}
    \setlength{\belowdisplayskip}{1pt}
    \begin{aligned}
    \min\limits_{v_1,\cdots,v_{k}} \sum_{i=1}^{k}(k-i+1)v_{i}^{T}\mathcal{L}v_{i} 
    , \ s.t.\  v_{i}^{T}v_{j} = \delta_{ij}, \forall\ i,j = 1, \cdots, k
    \end{aligned}
\end{equation}
For the large-scale state space, the eigenvectors can be represented as a neural network that takes a state $s$ as input and outputs a $k$-dimension vector $[f_1(s), \cdots, f_k(s)]$ as an estimation of $[v_1(s), \cdots, v_k(s)]$. Accordingly, the objective in Equation (\ref{opt}) can be expressed as: (please refer to \cite{DBLP:conf/icml/WangZZSHF21} for details)
\begin{equation} \label{obj}
    \setlength{\abovedisplayskip}{1pt}
    \setlength{\belowdisplayskip}{1pt}
    \begin{aligned}
    G(f_{1},\cdots,f_{k})=\frac{1}{2}\mathbb{E}_{(s,s')\sim\mathcal{T}}\left[\sum_{l=1}^{k}\sum_{i=1}^{l}(f_{i}(s)-f_{i}(s'))^2\right]
    \end{aligned}
\end{equation}
where $\mathcal{T}$ is a set of state-transitions collected by interacting with the environment randomly. Further, the orthonormal constraints in Equation (\ref{opt}) is implemented as a penalty term:
\begin{equation} \label{const}
\setlength{\abovedisplayskip}{1pt}
\setlength{\belowdisplayskip}{1pt}
    \begin{aligned}
    P(f_{1},\cdots,f_{k})=\beta\mathbb{E}_{s\sim\rho, s'\sim\rho}\left[\sum_{l=1}^{k}\sum_{i=1}^{l}\sum_{j=1}^{l} (f_{i}(s)f_{j}(s)-\delta_{ij})(f_{i}(s')f_{j}(s')-\delta_{ij})\right]
    \end{aligned}
\end{equation} 
where $\beta$ is the weight term and $\rho$ is the distribution of states in $\mathcal{T}$. To sum up, the eigenfunctions can be trained as NN by minimizing the loss function: $L(f_{1},\cdots,f_{k}) = G(f_{1},\cdots,f_{k}) + P(f_{1},\cdots,f_{k})$.
% \begin{equation} \label{loss}
%     \setlength{\abovedisplayskip}{1pt}
%     \setlength{\belowdisplayskip}{1pt}
%     \begin{aligned}
%     L(f_{1},\cdots,f_{k}) = G(f_{1},\cdots,f_{k}) + P(f_{1},\cdots,f_{k})
%     \end{aligned}
% \end{equation}

After getting estimation of $v_{1:k}$, we can estimate the corresponding eigenvalues $\lambda_{1:k}$ by: 
\begin{equation}
    \setlength{\abovedisplayskip}{0.5pt}
    \setlength{\belowdisplayskip}{1pt}
    \begin{aligned}
    \lambda_{i} = v_{i}^{T}\mathcal{L}v_{i} = \frac{1}{2}\mathbb{E}_{(s,s')\sim\mathcal{T}}\left[(f_{i}(s)-f_{i}(s'))^2\right],\ i=1,\cdots,k
    \end{aligned}
\end{equation}
As for the degree $d$ of each state in the individual state transition graph, we can obtain them through $f_1$ which is the eigenvector corresponding to the smallest eigenvalue of $\mathcal{L}$ and satisfies $f_{1}(s_{1})/f_{1}(s_{2})=\sqrt{d(s_{1})} / \sqrt{d(s_{2})}$ \cite{chung1997spectral}. Therefore, we can collect the $k$-smallest eigenvalues and corresponding eigenvectors, and the degree list for each agent, after which we can utilize Theorem \ref{thm:2} to estimate the joint Fielder vector. Note that since we only care about the second smallest eigenvalue, we don't need to enumerate all the eigenvalues of the factor graphs, and we only need to know the relative value of the degrees. 
After getting the joint Fiedler vector, we define the subgoal joint states corresponding to the minimum or maximum (e.g., $s_{min}$=($s_{min}^{1}$, $\cdots$, $s_{min}^{n}$)) as the termination set, and the other joint states as the initiation set. As for the intra-option policy, it can be trained with any deep RL algorithms for each agent $i$ to reach its corresponding individual termination state (e.g., $s_{min}^{i}$). In this way, we can obtain the multi-agent options in infinite-scale state spaces.

\subsection{Adopting Multi-agent Options in MARL} \label{framework}
\begin{wrapfigure}{r}{6.3cm}
	\centering
 \includegraphics[height=1.1in, width=2.3in]{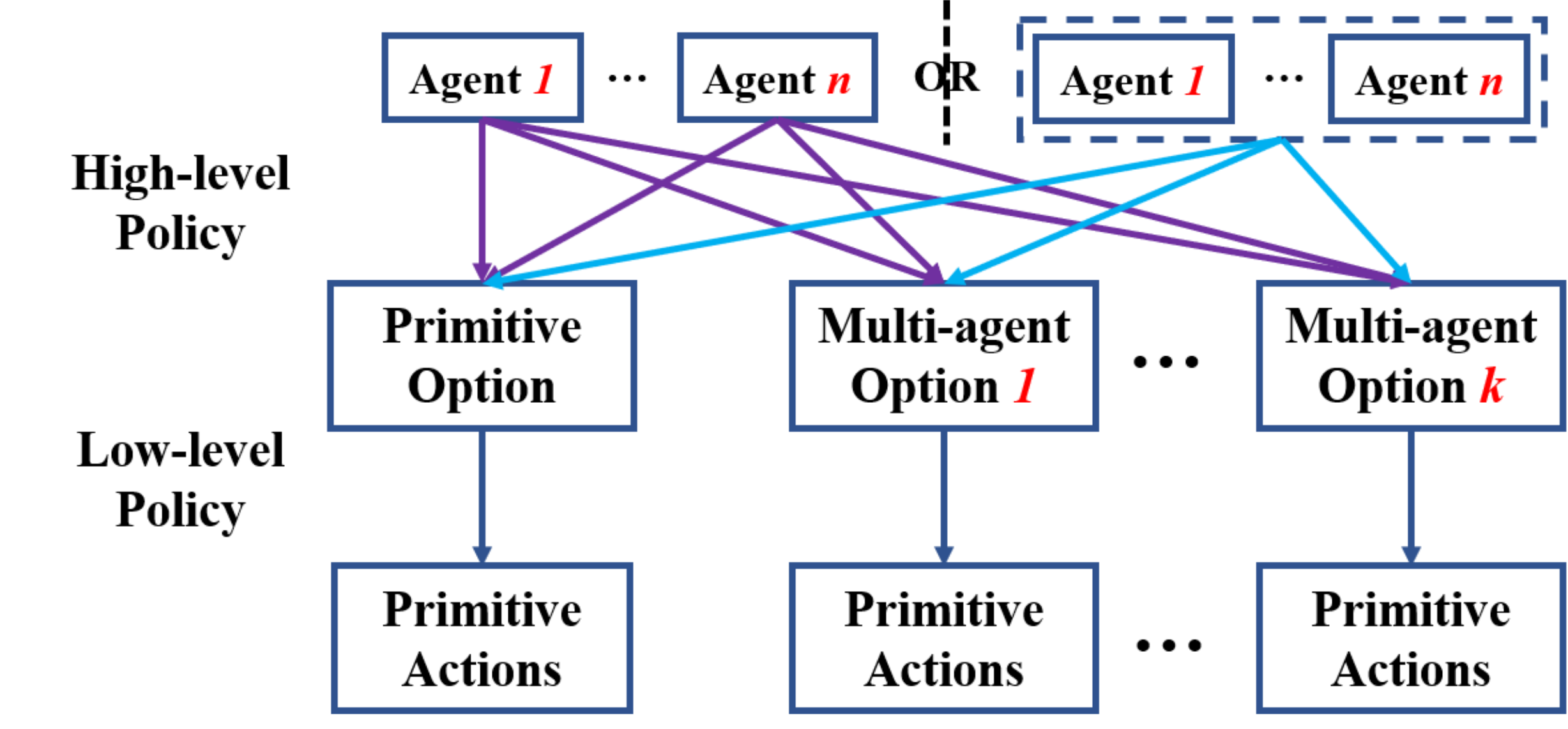}
	\caption{Hierarchical algorithm framework}
	\label{fig:1}
\end{wrapfigure}

To utilize options in MARL, we adopt a hierarchical algorithm framework shown in Figure \ref{fig:1}. Typically, we train a RL agent to select among the primitive actions, aiming to maximize the accumulated reward. We view this agent as a one-step option -- primitive option. As shown in Figure \ref{fig:1}, when getting new observations, the hierarchical agent first decides on which option $\omega$ to use with the high-level policy, and then decides on the primitive action to take based on the corresponding intra-option policy $\pi_{\omega}$. The agent does not decide on a new option with the high-level policy until the current option terminates.

For a multi-agent option $<I_{\omega}, \pi_{\omega}=(\pi_{\omega}^{1}, \cdots, \pi_{\omega}^{n}), \beta_{\omega}=\{(s_{1}, \cdots, s_{n})\}>$, it can be adopted either in a decentralized or centralized manner. As the purple arrows in Figure \ref{fig:1}, the agents choose their own options independently, and they may choose different options to execute in the meantime. In this case, if agent $i$ selects option $\omega$, it will execute $\pi_{\omega}^{i}$ until reaching its termination state $s_{i}$ or exceeding the execution step limit. On the other hand, we can force the agents to execute the same multi-agent option simultaneously. As shown by the blue arrows in Figure \ref{fig:1}, we view the $n$ agents as a whole, which takes the joint state as the input and chooses the same multi-agent option to execute at a time. Once a multi-agent option $\omega$ is chosen, agents $1:n$ will execute $\pi_{\omega}^{1:n}$ until they reach $(s_{1}, \cdots, s_{n})$ or exceed the step limit. The decentralized way is more flexible but has a larger search space. While, the centralized way fails to consider all the possible solutions but makes it easier for the agents to visit the sub-goal joint states, since the agents select the same joint option which won't terminate within the time limit until the agents arrive at the sub-goal. 

Further, the centralized manner may not be applicable when the number of agents $n$ is large, since the input space will grow exponentially with $n$. Thus, we propose to partition the agents into sub-groups, and then learn the joint options within each sub-group. The intuition is that a multi-agent task can usually be divided into sub-tasks, each of which can be completed by a sub-group of agents. For each sub-group, we can learn a list of joint options, and then the agents of this group can utilize these options in a decentralized or centralized way as mentioned above. Further, if there is no way to divide the agents based on sub-tasks, we can still partition them randomly to some two-agent or three-agent sub-groups. Agents within the same sub-group will co-explore their joint state space sharing their views. More intelligent grouping methods, e.g., attention mechanism \cite{DBLP:conf/nips/VaswaniSPUJGKP17}, can be easily integrated into our solution to boost the performance, which is not the focus here (multi-agent option discovery). 
% In Section \ref{evaluation}, we show that the grouping techniques can greatly accelerate the exploration and improve the scalability.

\section{Evaluation and Results} \label{evaluation}

\begin{figure}[t]
\centering
\subfigure[Grid Room]{
\label{fig:2(b)} 
\includegraphics[width=1.2in, height=0.9in]{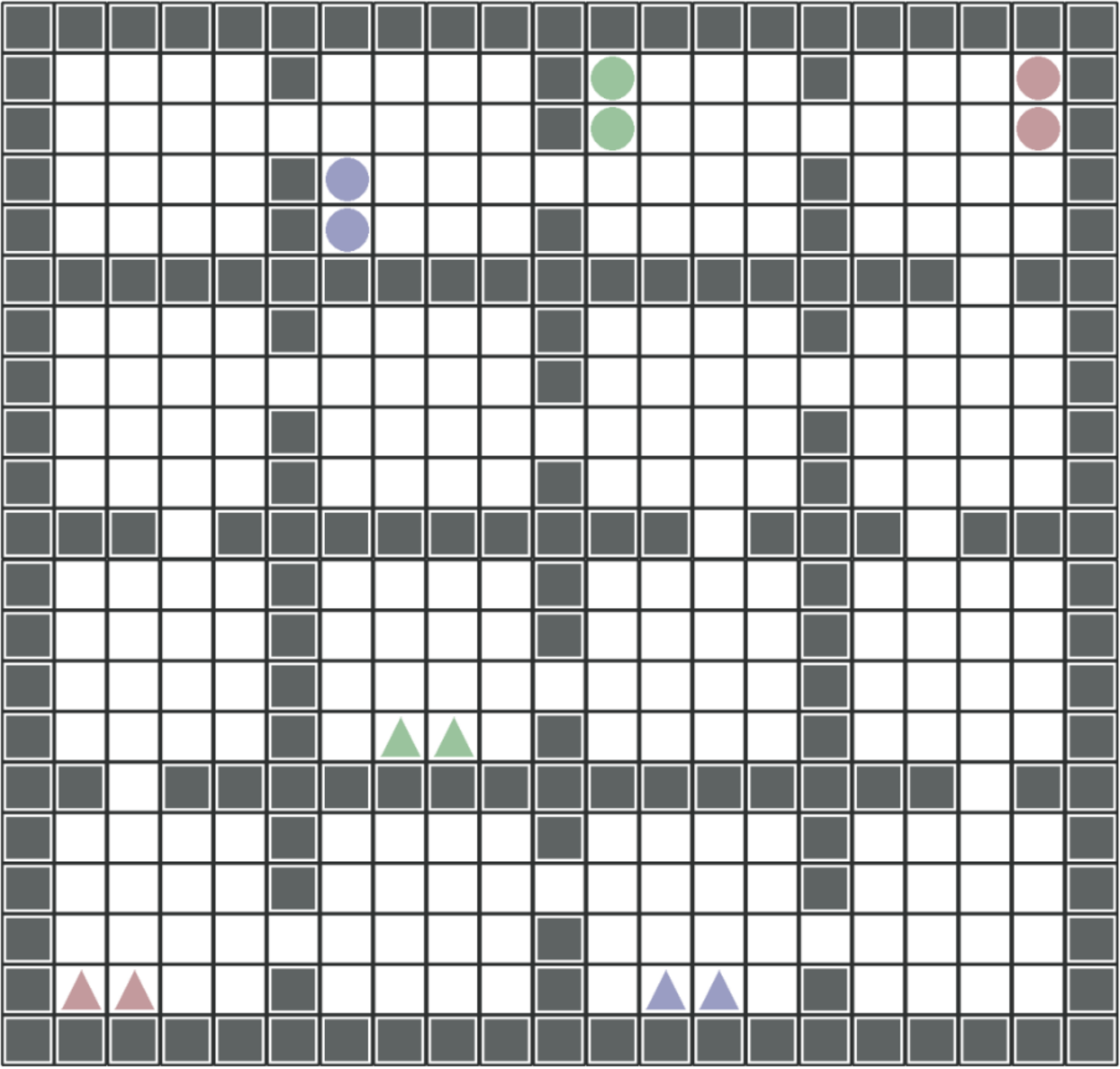}}
\subfigure[Grid Maze]{
\label{fig:2(c)} 
\includegraphics[width=1.2in, height=0.9in]{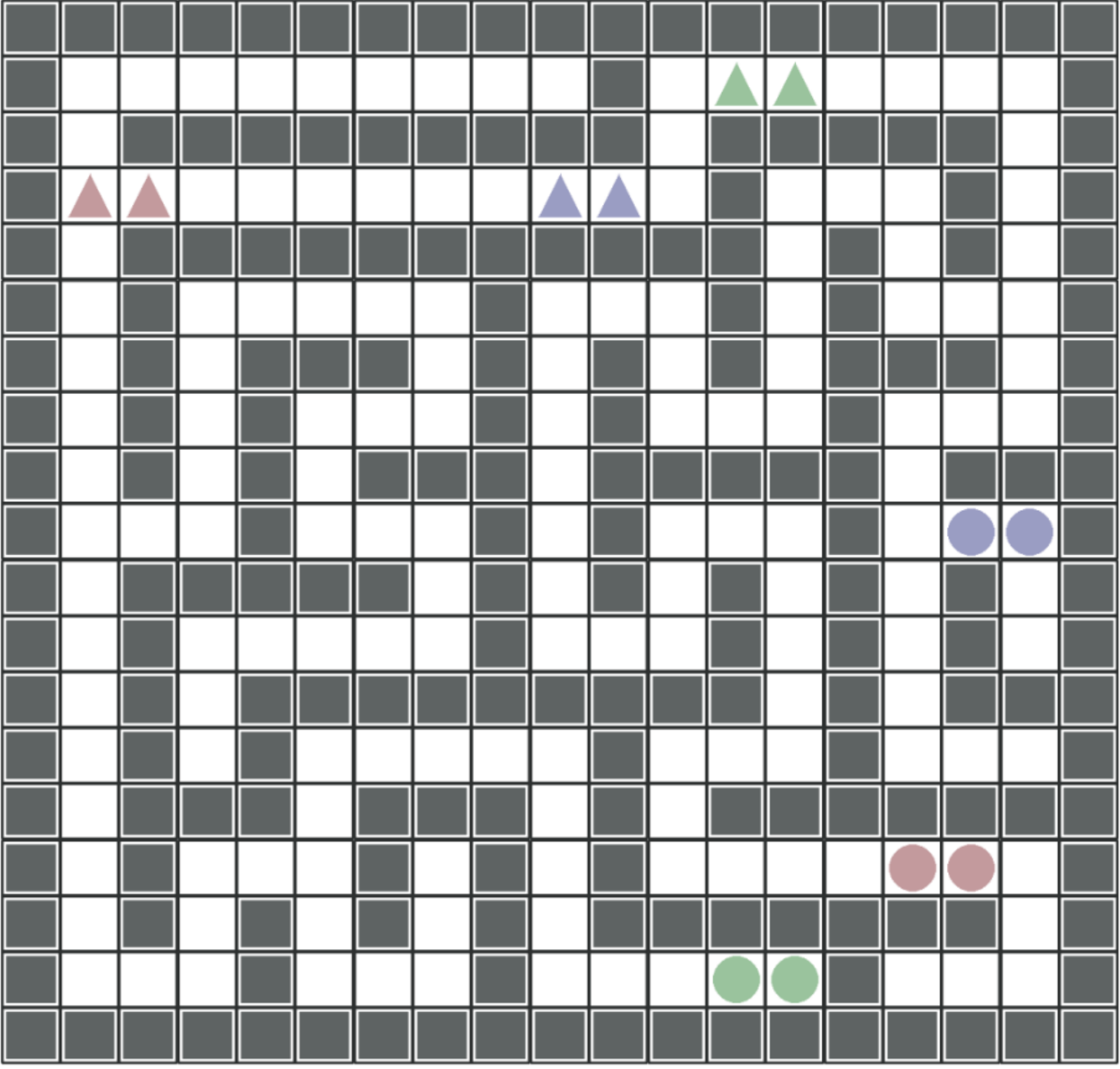}}
\subfigure[Mujoco Room]{
\label{fig:2(d)} 
\includegraphics[width=1.2in, height=0.9in]{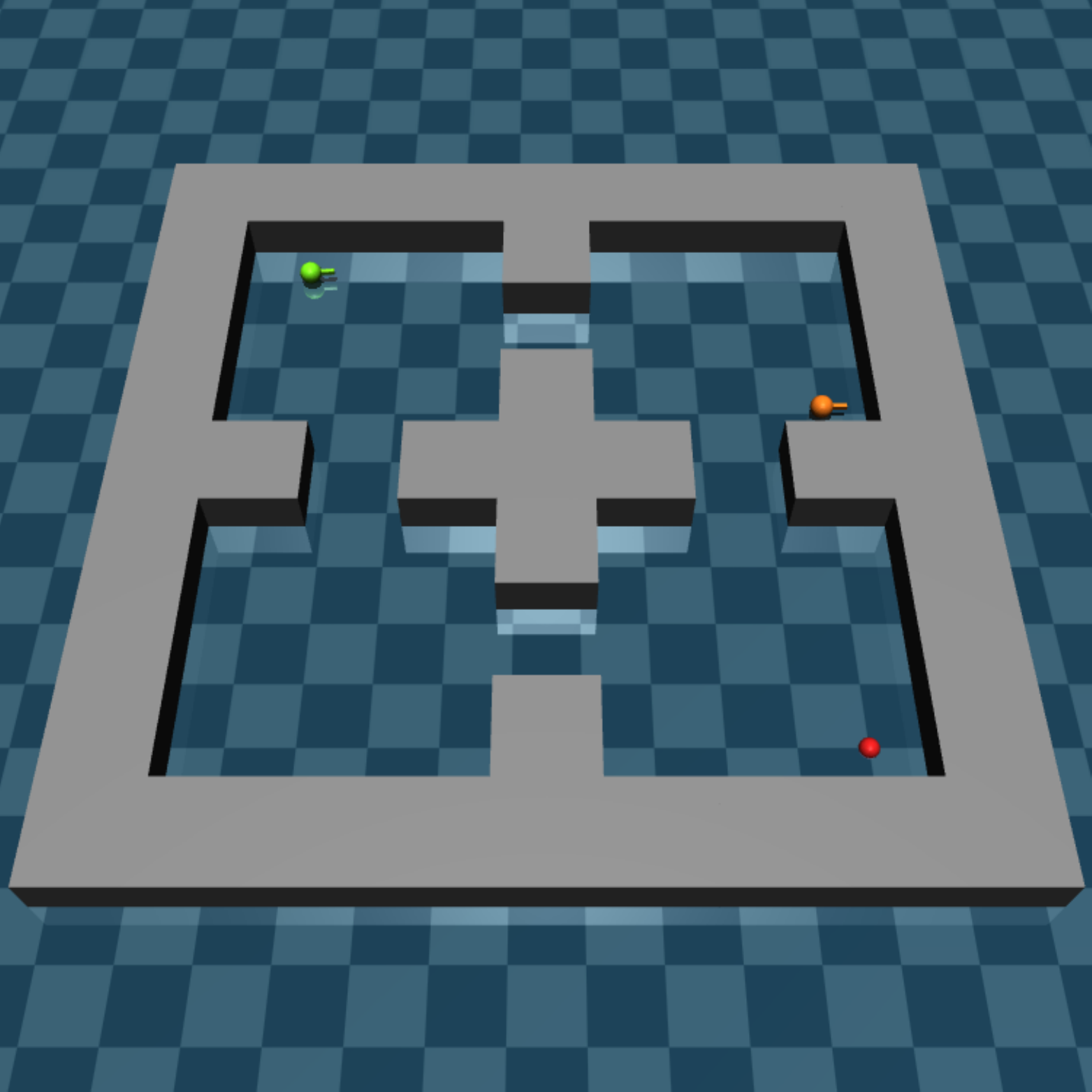}}
\subfigure[Mujoco Maze]{
\label{fig:2(e)} 
\includegraphics[width=1.2in, height=0.9in]{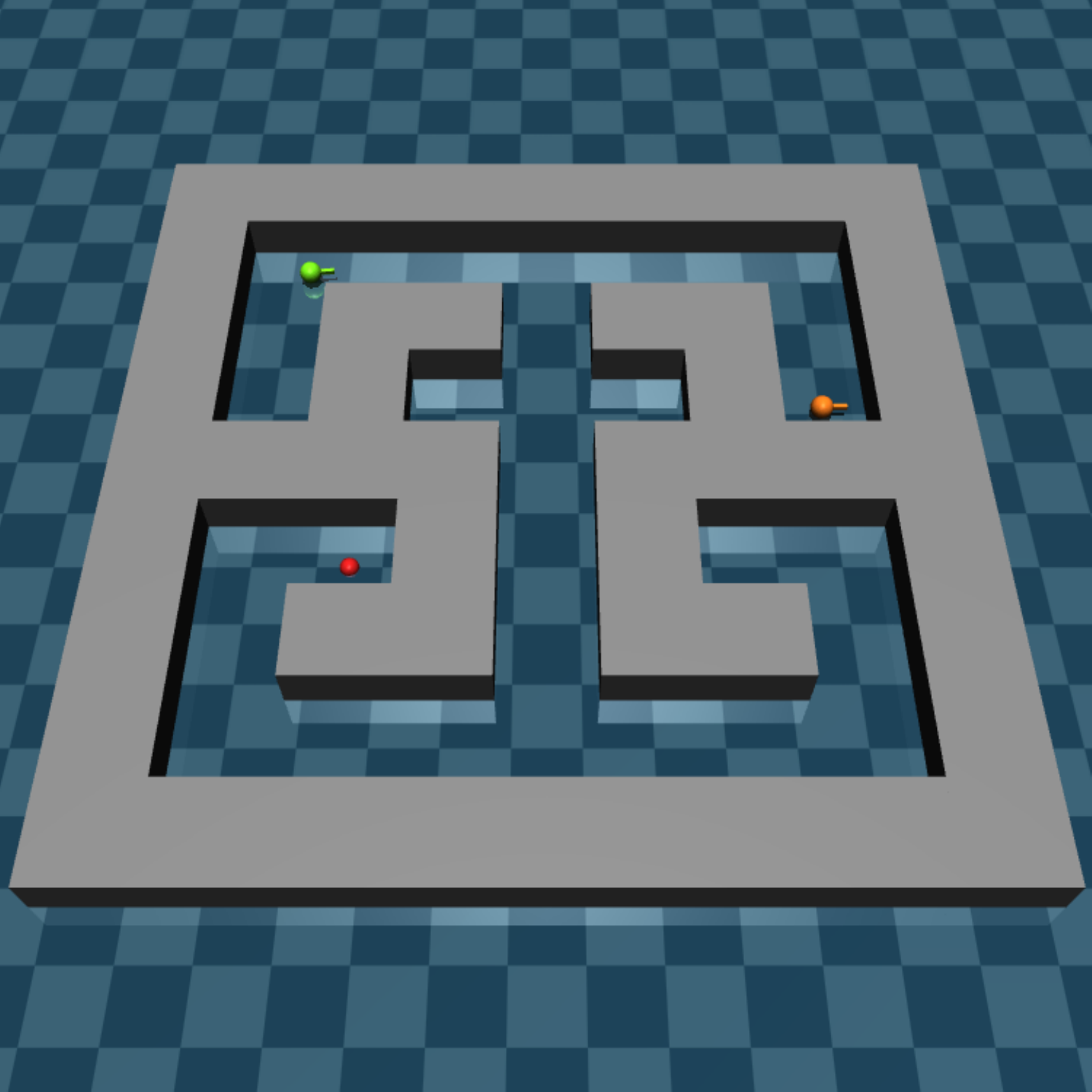}}

\caption{Simulators for Evaluation}
\label{fig:2} %% label for entire figure

\end{figure}

\subsection{Experiment Setup}

\begin{figure}[t]
\centering
\subfigure[Grid Maze with 2 agents]{
\label{fig:3(b)} 
\includegraphics[width=1.7in, height=0.85in]{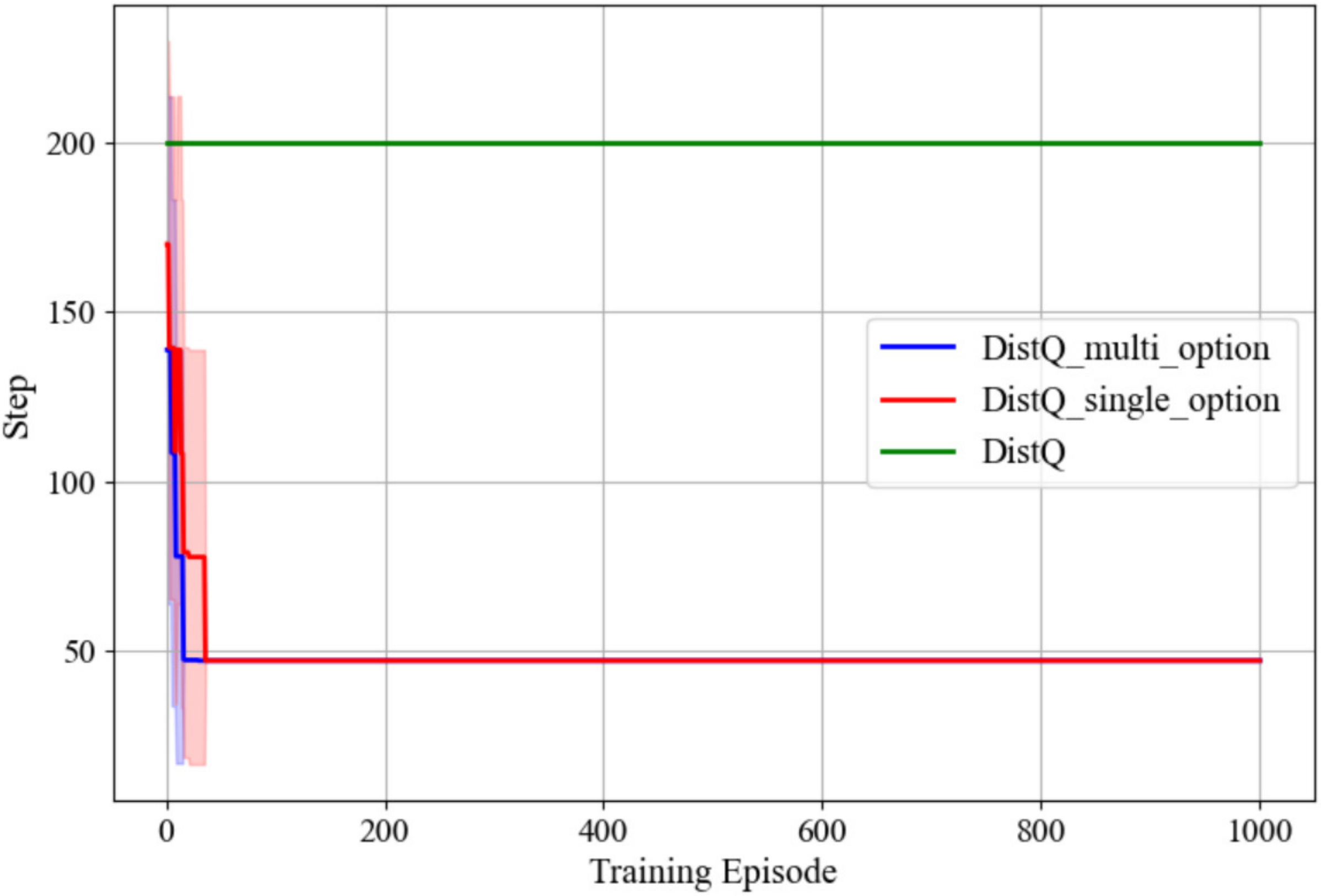}}
\subfigure[Grid Maze with 3 agents]{
\label{fig:3(c)} 
\includegraphics[width=1.7in, height=0.85in]{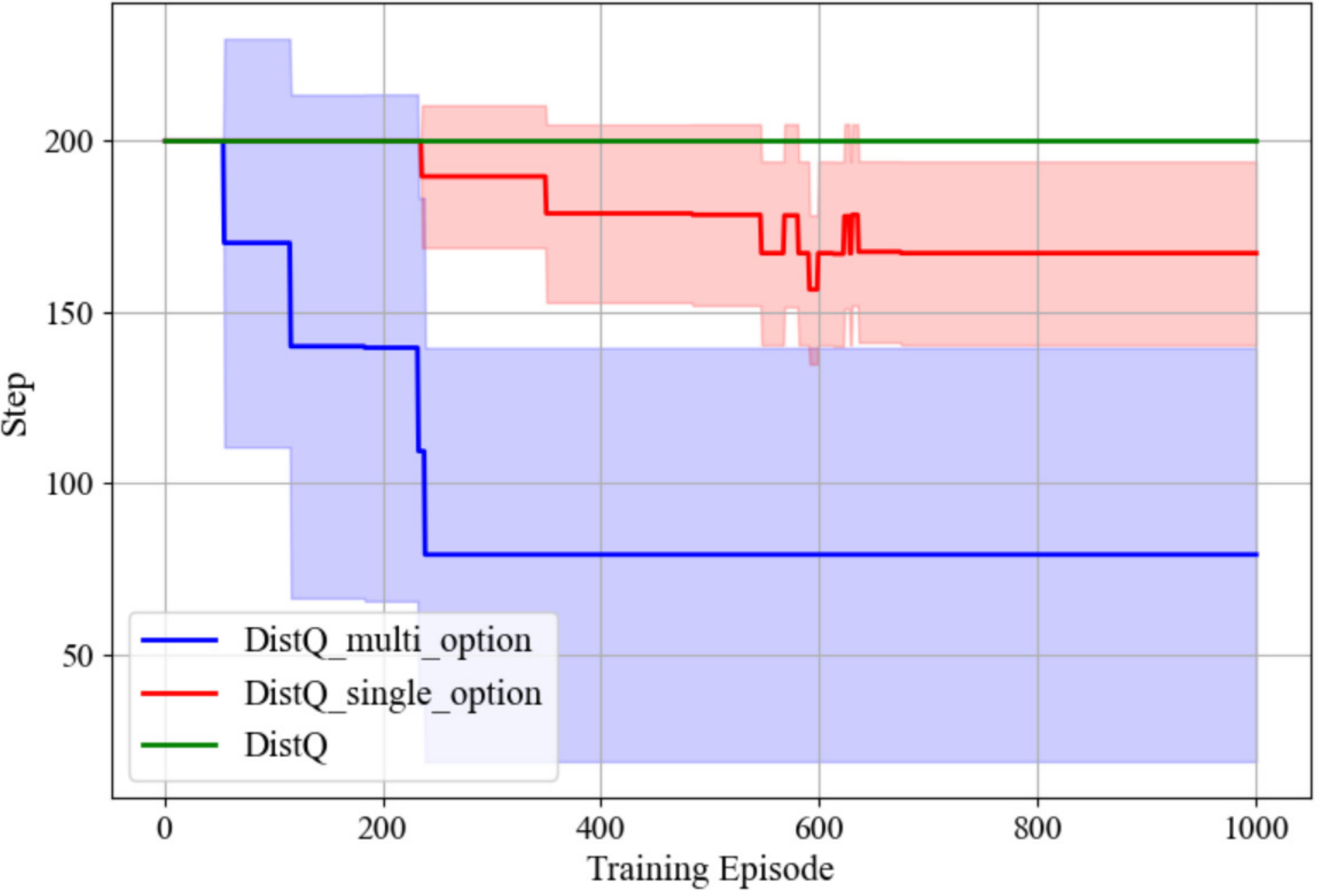}}
\subfigure[Grid Maze with 4 agents]{
\label{fig:3(d)} 
\includegraphics[width=1.7in, height=0.85in]{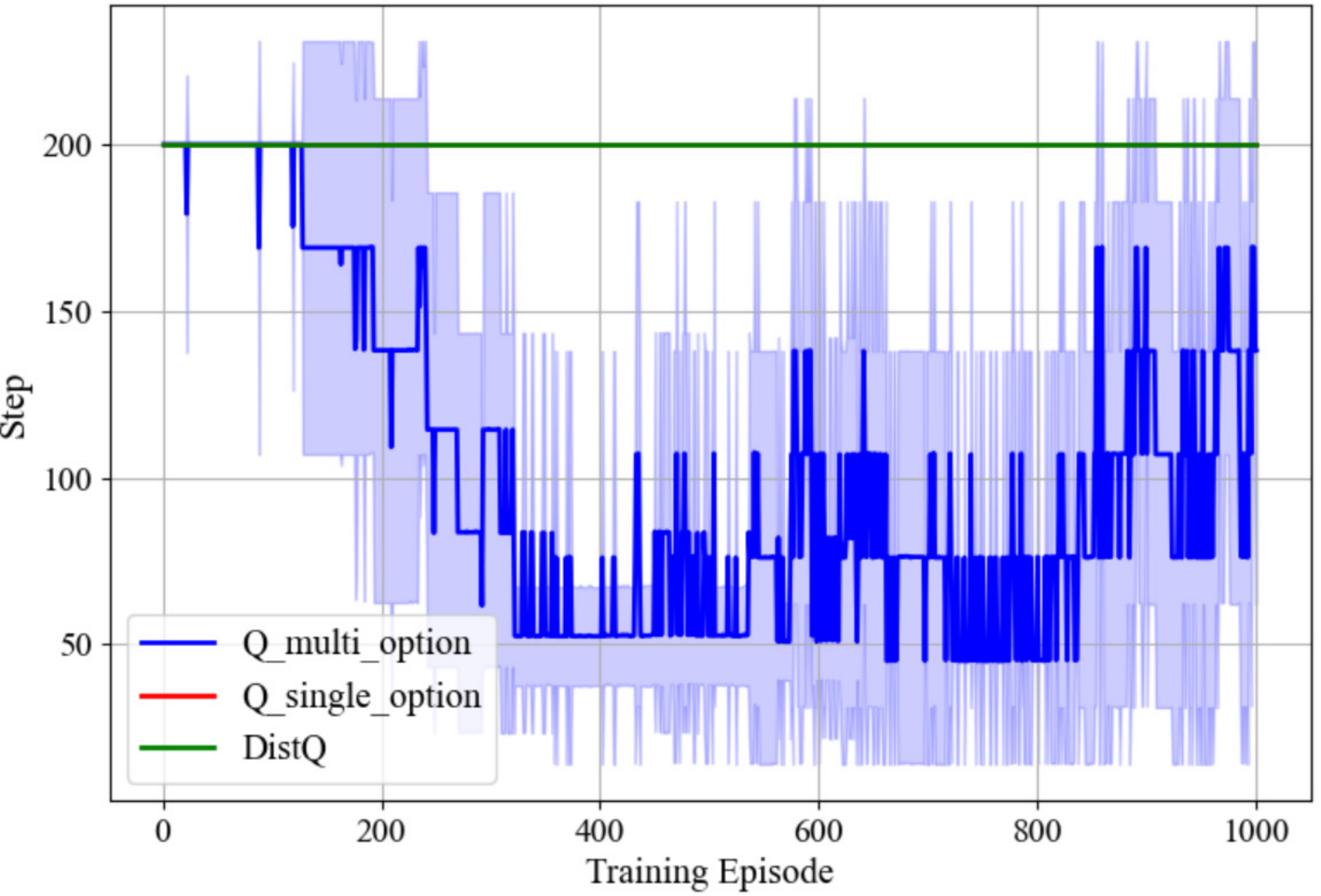}}
\subfigure[Grid Maze with 2 agents]{
\label{fig:3(e)} 
\includegraphics[width=1.7in, height=0.85in]{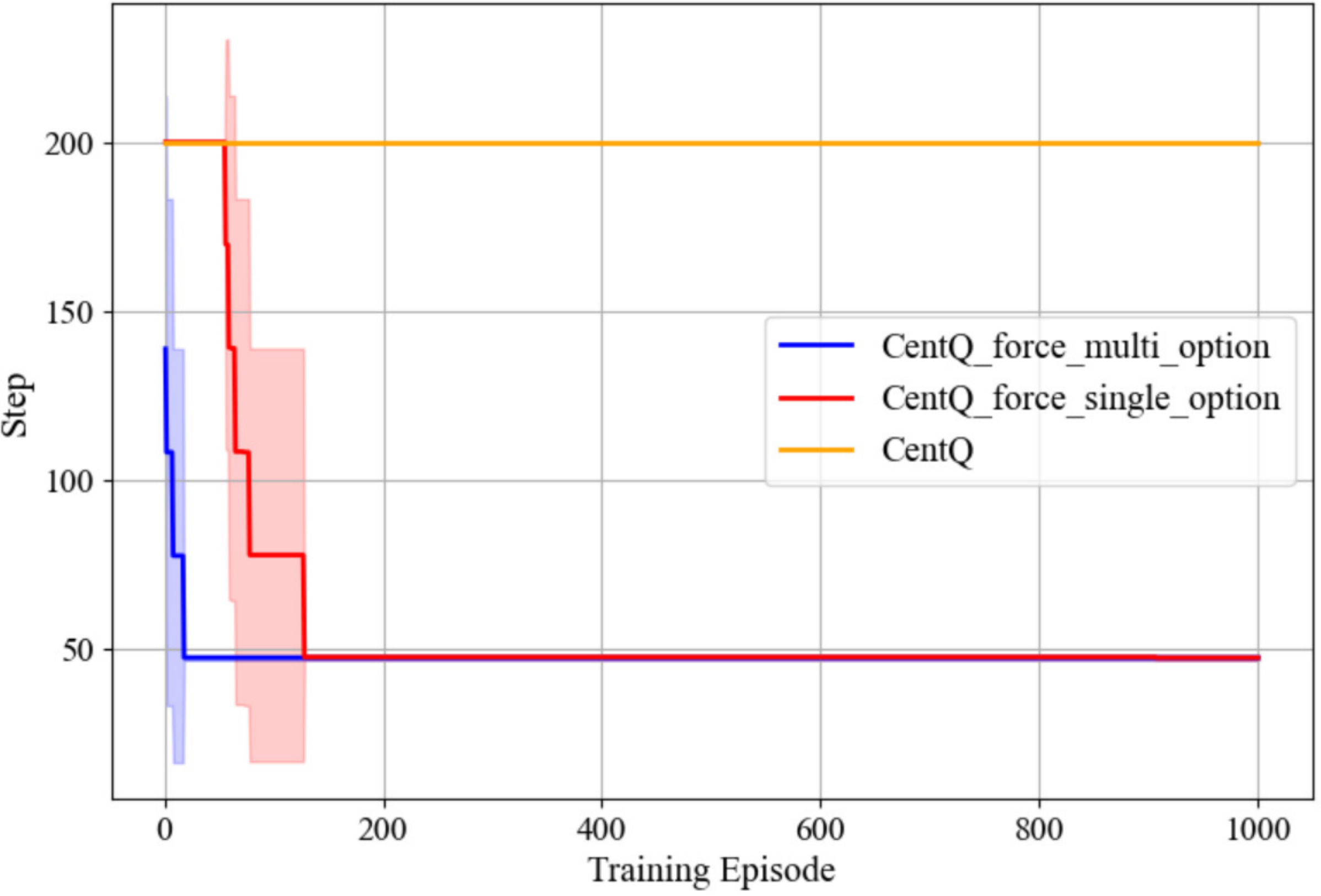}}
\subfigure[Grid Maze with 3 agents]{
\label{fig:3(f)} 
\includegraphics[width=1.7in, height=0.85in]{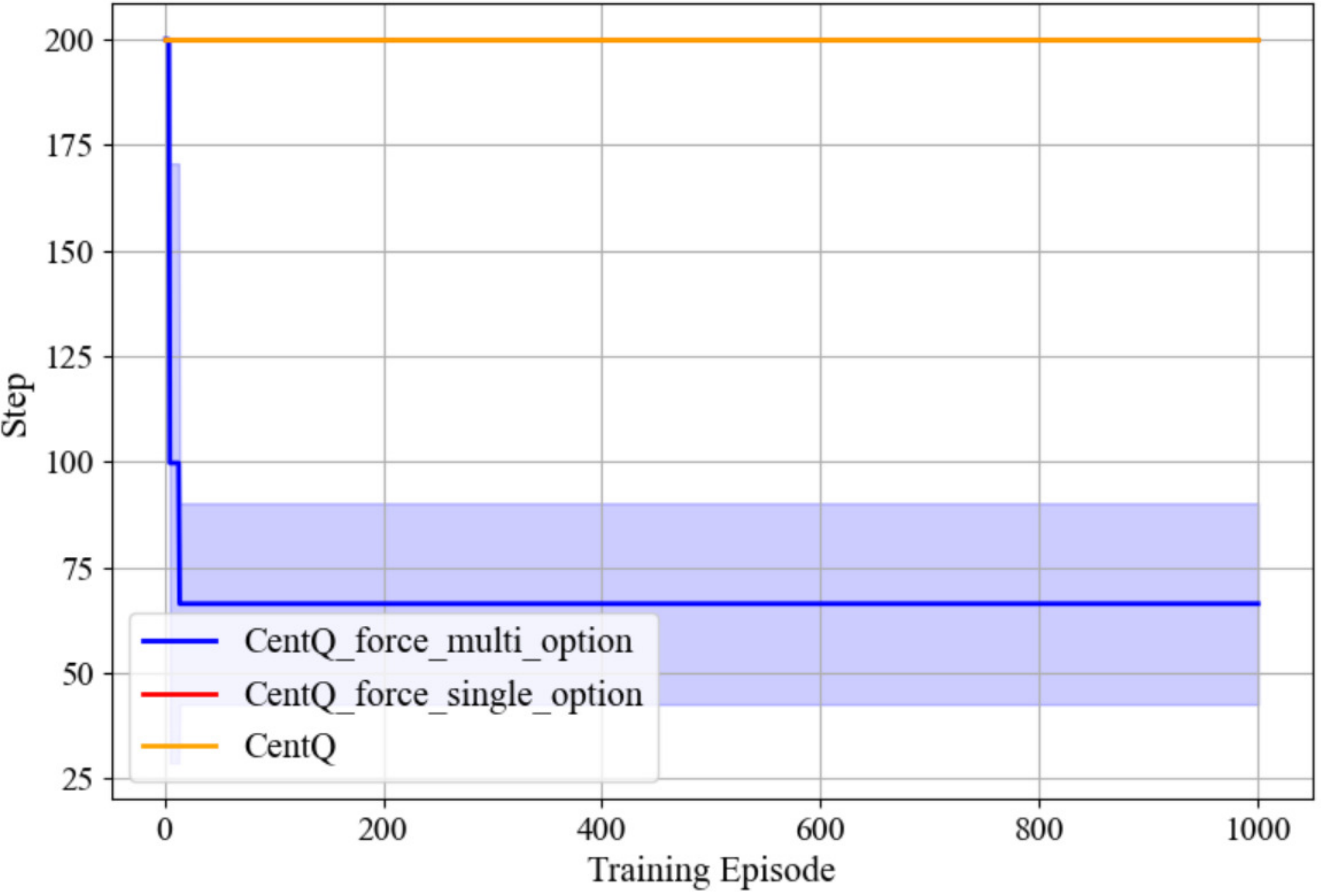}}
\subfigure[Grid Maze with 4 agents]{
\label{fig:3(g)} 
\includegraphics[width=1.7in, height=0.85in]{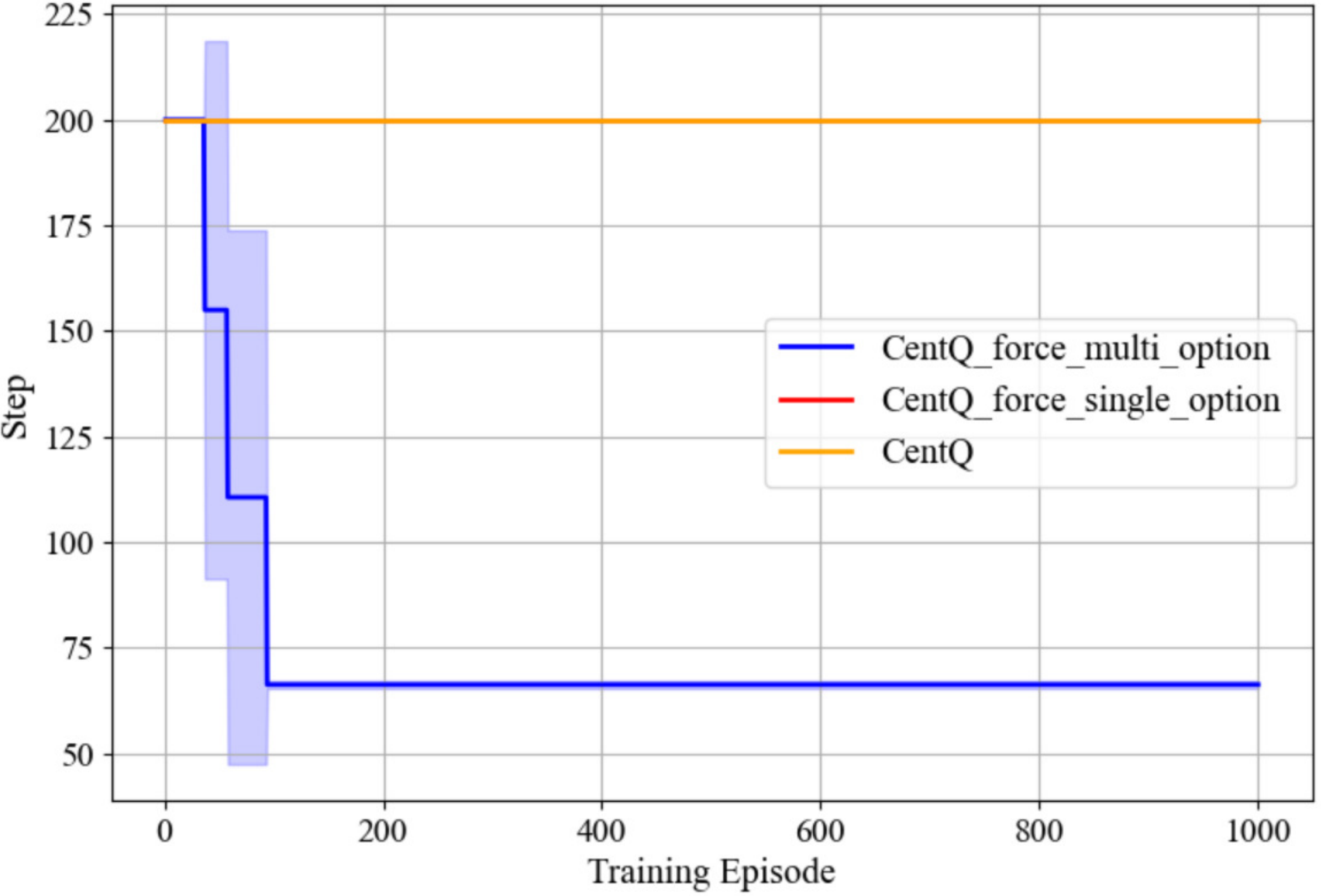}}

\caption{Evaluation on $n$-agent Grid Maze tasks: (a)-(c): Distributed Q-Learning; (d)-(f) Centralized Q-Learning + Force. The performance improvement of our approach are more and more significant as the number of agents increases. The centralized way to utilize the $n$-agent options performs better.}

\label{fig:3} 

\end{figure}

As shown in Figure \ref{fig:2}, the proposed approach is evaluated on four multi-agent goal-achieving tasks. (1) For tasks shown as Figure \ref{fig:2(b)} and \ref{fig:2(c)}, $n$ (2$\sim$8) agents (triangles) must reach the goal area (circles) at the same time to complete this task; Or even harder, $n \times m$ agents are divided into $m$ groups and each group of agents has a special goal labeled with the same color, the $n \times m$ agents should get to their related goals simultaneously to complete this task, without knowing which goal is related to them at first. Hence, highly coordinated policy is required. (2) To test the performance of the deep learning extension of our algorithm, we build the tasks shown as Figure \ref{fig:2(d)} and \ref{fig:2(e)} based on Mujoco \cite{todorov2012mujoco}, where two point agents need to reach the target area (red) simultaneously to complete the task. Both the state and action spaces of the point agent are continuous and infinite-scale.  Note that these tasks are quite challenging: (1) Agents’ observations do not include any information about the goal area, so they need to fully explore their joint state space to find the rewarding state. (2) The reward space is highly sparse and delayed: for all the tasks, only when the agents complete the task can they receive a reward signal $r=1.0$ (shared by all the agents); otherwise, they will receive $r=0.0$. Hence, agents without highly-efficient exploration strategies like ours cannot complete these tasks. In Section \ref{results}, we conduct experiments with tasks of increasing complexity (e.g., Figure \ref{fig:3}), showing that the more difficult the task is, the more advantageous our approach becomes.

% The observation of a point agent includes its position, orientation and corresponding velocities: $(x, y, \theta, \dot{x}, \dot{y}, \dot{\theta})$, which are continuous. Based on that, the point agent needs to decide on the change of its orientation and distance to move forward, which are also continuous.

There are a high-level and low-level policy in the hierarchical framework (Figure \ref{fig:1}). The low-level policy is decomposed into single-agent policies for each agent to reach its individual termination state, so it can be trained with single-agent RL. (1) For discrete tasks (e.g., Figure \ref{fig:2(b)} and \ref{fig:2(c)}), we adopt Distributed Q-Learning \cite{DBLP:conf/icml/LauerR00} (\textbf{decentralized manner}: each agent decides on its own option based on the joint state) or Centralized Q-Learning + Force (\textbf{centralized manner}: viewing $n$ agents as a whole, adopting Q-Learning to this joint agent and forcing them to choose the same joint option at a time) to train the high-level policy and adopt Value Iteration \cite{sutton2018reinforcement} for the low-level policy training. (2) For continuous control tasks (e.g., Figure \ref{fig:2(d)} and \ref{fig:2(e)}), the tabular RL algorithms mentioned above cannot work. Instead, to improve the scalability of the our method, we adopt MAPPO \cite{DBLP:journals/corr/abs-2103-01955} to train the high-level policy of the joint options, and Soft Actor-Critic \cite{DBLP:conf/icml/HaarnojaZAL18} to train the low-level policy, which are SOTA deep MARL and RL algorithms respectively.
% While, the low level policy for the primitive option is adopted to complete the whole multi-agent task, so it should be trained with MARL algorithms. 
% We assume the access to joint states within a subgroup, since the multi-agent option discovery is defined on the joint state space. This can be realized through communication mechanisms \cite{DBLP:conf/nips/FoersterAFW16} in MARL which is a separate line of work.

We compare our approach -- agents with multi-agent options, with two baselines: (1) Agents without options: SOTA MARL algorithms are adopted to train the agents without using options. For discrete tasks, we use Distributed and Centralized Q-Learning. These tabular Q-learning algorithms which update the Q-value for all the state-action pairs in each training episode, usually outperform the NN-based algorithms. We show this through comparison with COMA \cite{DBLP:conf/aaai/FoersterFANW18}, MAVEN \cite{DBLP:conf/nips/MahajanRSW19},  Weighted QMIX \cite{DBLP:conf/nips/RashidFPW20}. For continuous tasks, we use MAPPO, MADDPG \cite{DBLP:conf/nips/LoweWTHAM17} and MAA2C as comparisons, which have been proven as strong MARL baselines for continuous control \cite{DBLP:conf/nips/PapoudakisC0A21, DBLP:journals/corr/abs-2006-07869}.  Comparisons with this baseline can show the effectiveness of using options to aid exploration. (2) Agents with single-agent options: we first construct covering options for each agent based on their individual state spaces, and then utilize these options in MARL, like what they do in \cite{DBLP:conf/atal/AmatoKK14, amato2019modeling, shen2006multi, DBLP:conf/atal/ChakravortyWRCB20, DBLP:conf/iclr/LeeYL20}. As for the option discovery method, we adopt the SOTA algorithm proposed in \cite{DBLP:conf/icml/JinnaiPAK19, DBLP:conf/iclr/JinnaiPMK20}, which claims to outperform previous option discovery algorithms for sparse reward scenarios, like \cite{,DBLP:journals/corr/MachadoBB17, DBLP:conf/iclr/EysenbachGIL19}. Comparisons with this baseline can show the superiority of our approach to directly identify and adopt joint options in multi-agent tasks. The number of single- and multi-agent options for each agent to select is the same. 
% Also, we extend the initiation set of each single-agent option to the known area to increase accessibility, like what we do with multi-agent options.

Last, regarding the setup of the option discovery phase, we collect $1\times10^4$ transitions (i.e., $\{(s, a, s')\}$) for the discrete tasks and $5\times10^4$ transitions for the continuous control tasks in order to build the state transition graphs, based on which we can extract single- or multi-agent options. In discrete tasks, we use a random walk policy to collect the data. While, in Mujoco tasks, the data collection is in stages. Specifically, the agents explore the environment through a random walk in the first stage. Then, in the next stage, the agents explore with their primitive actions and options that are extracted from the samples collected in the first stage. Our results guarantee that these options can significantly improve the joint exploration in the second stage. As for the number, we learn 16 multi-agent options in the Grid tasks and 8 multi-agent options in the Mujoco tasks. 
% All the options are discovered in an unsupervised manner, which is not based on any reward signal and not fine-tuned.

\begin{figure}[t]
\centering
\subfigure[2$\times$2 agents]{
\label{fig:6(a)} 
\includegraphics[width=1.31in, height=0.85in]{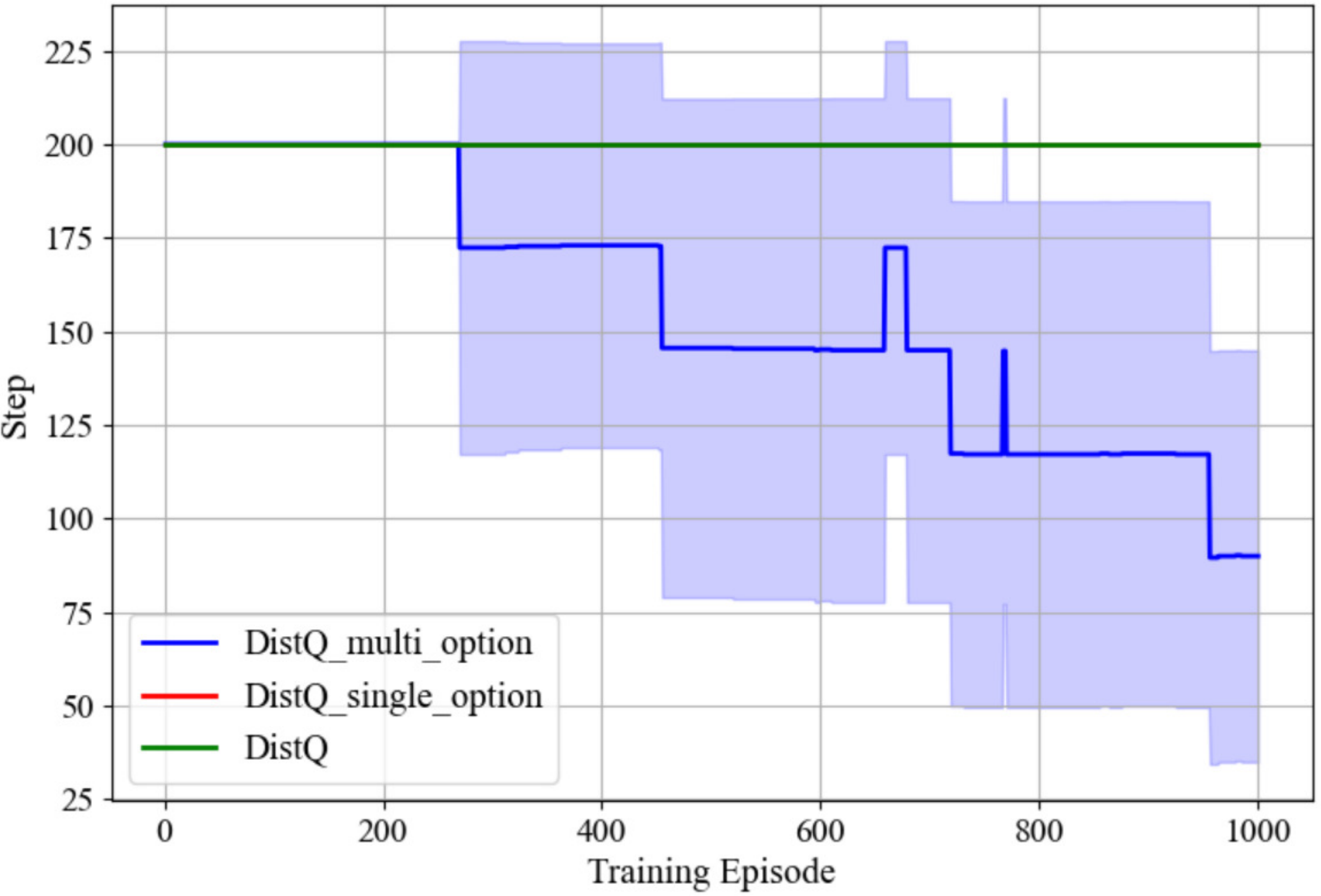}}
\subfigure[3$\times$2 agents]{
\label{fig:6(b)} 
\includegraphics[width=1.31in, height=0.85in]{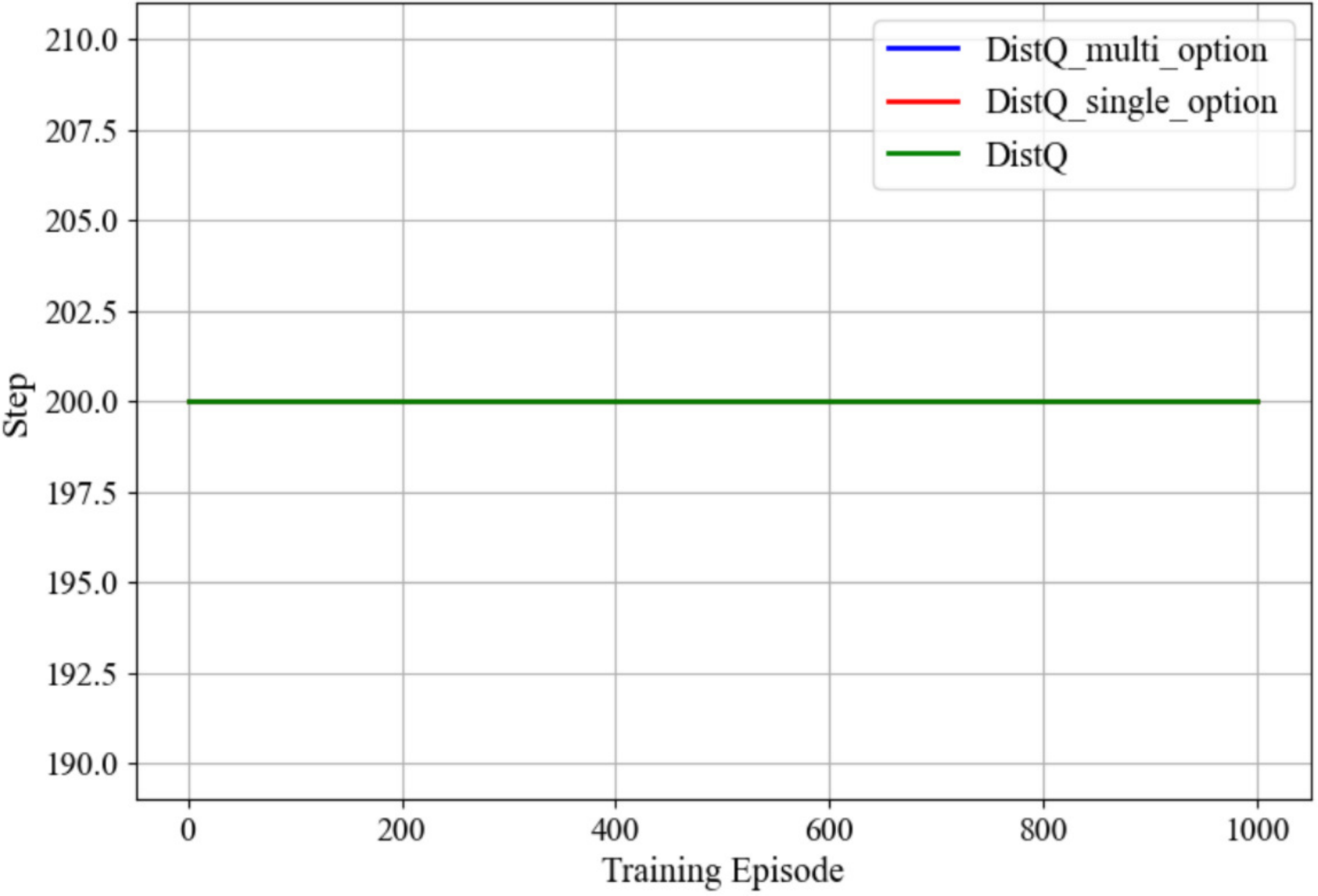}}
\subfigure[2$\times$2 agents]{
\label{fig:6(c)} 
\includegraphics[width=1.31in, height=0.85in]{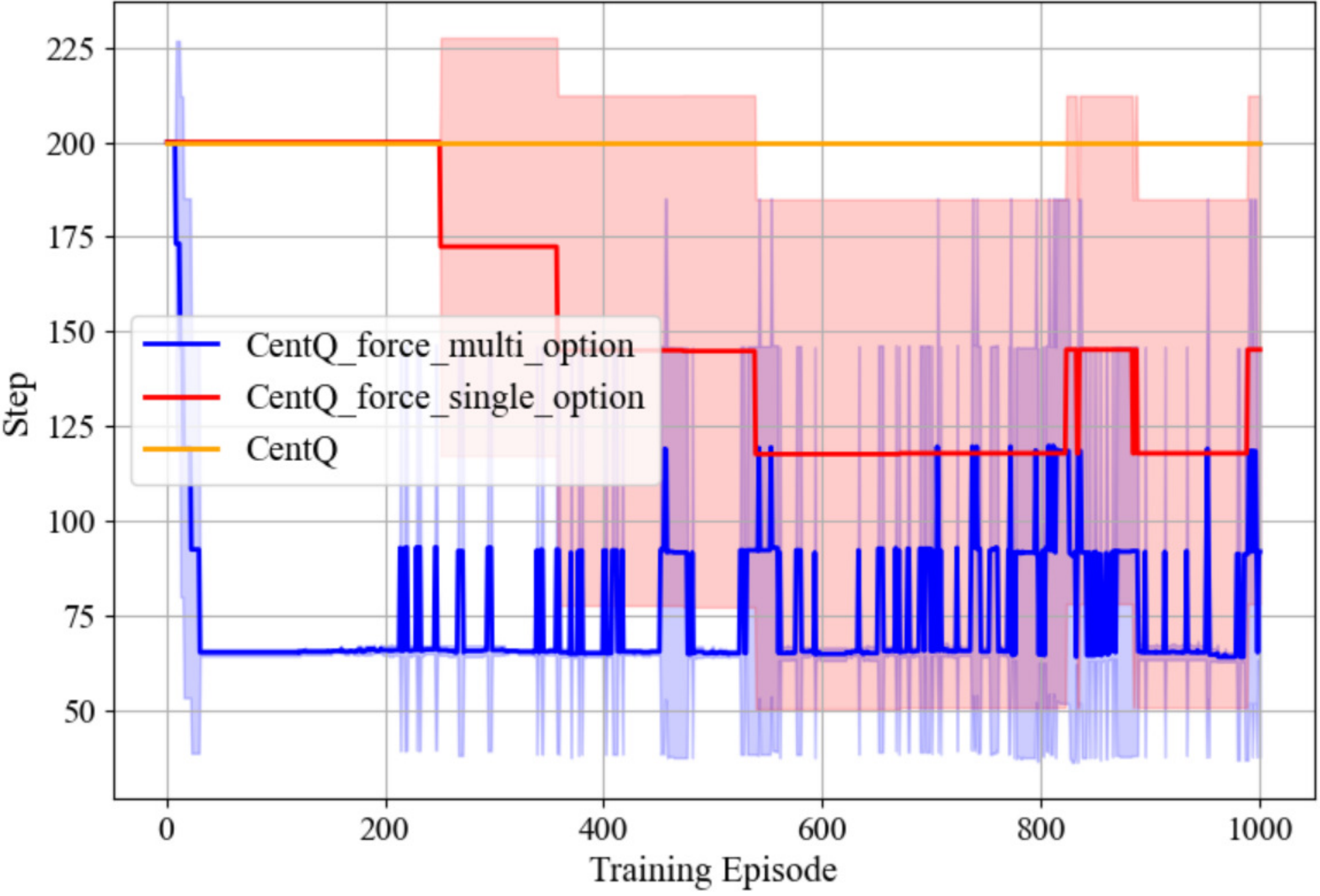}}
\subfigure[3$\times$2 agents]{
\label{fig:6(d)} 
\includegraphics[width=1.31in, height=0.85in]{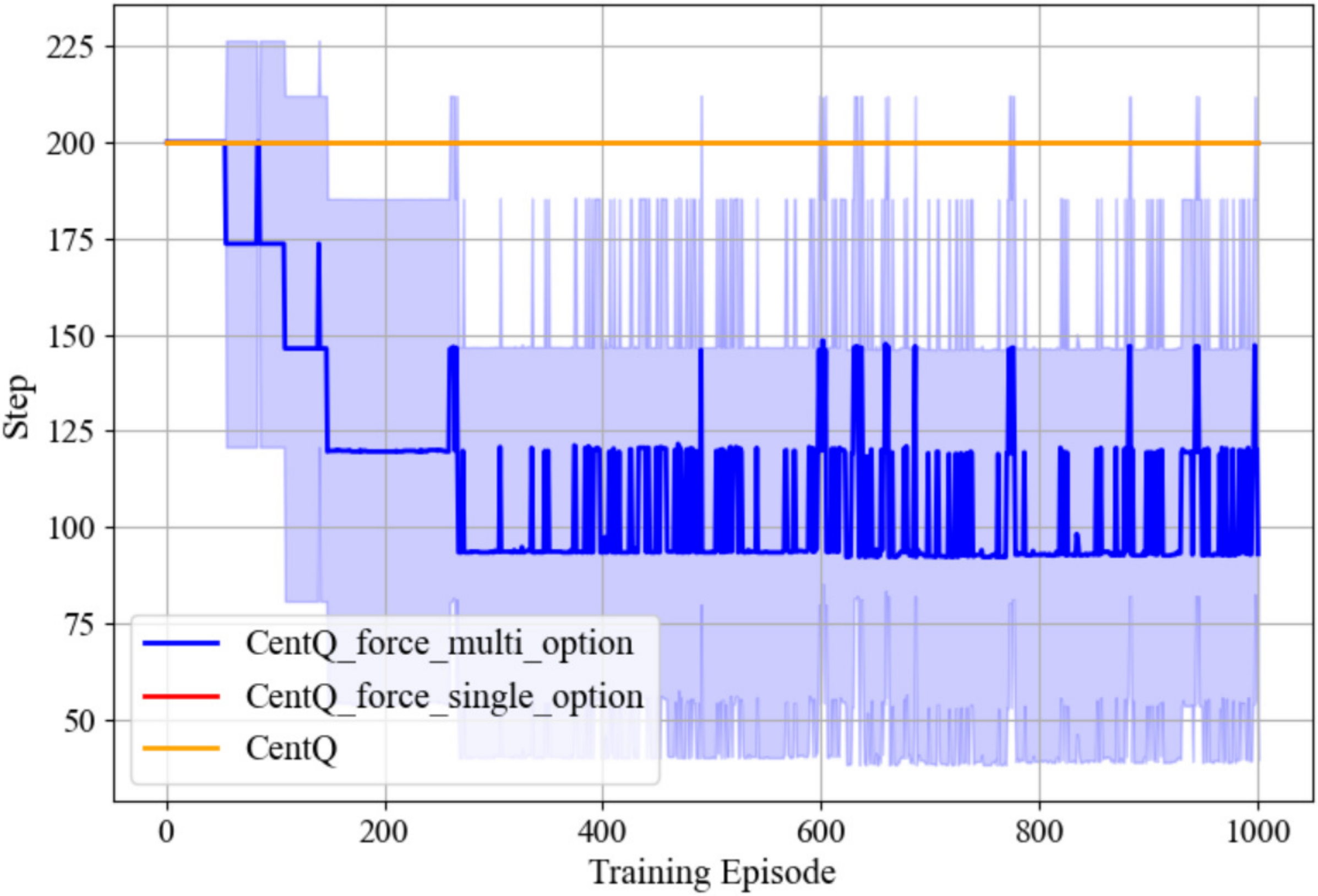}}

\caption{Comparisons on the $m \times n$ Grid Maze tasks: (a)-(b) Distributed Q-Learning; (c)-(d) Centralized Q-Learning + Force. Agents with pairwise options can learn these tasks much faster than the baselines, even when both the baselines fail on the $3\times2$ Grid Maze task. Also, agents trained with Centralized Q-Learning + Force have higher convergence speed and better final performance.}
\label{fig:6} %% label for entire figure

\end{figure}

\begin{figure}[t]
\centering
\subfigure[4 agents]{
\label{fig:4(a)} 
\includegraphics[width=1.31in, height=0.85in]{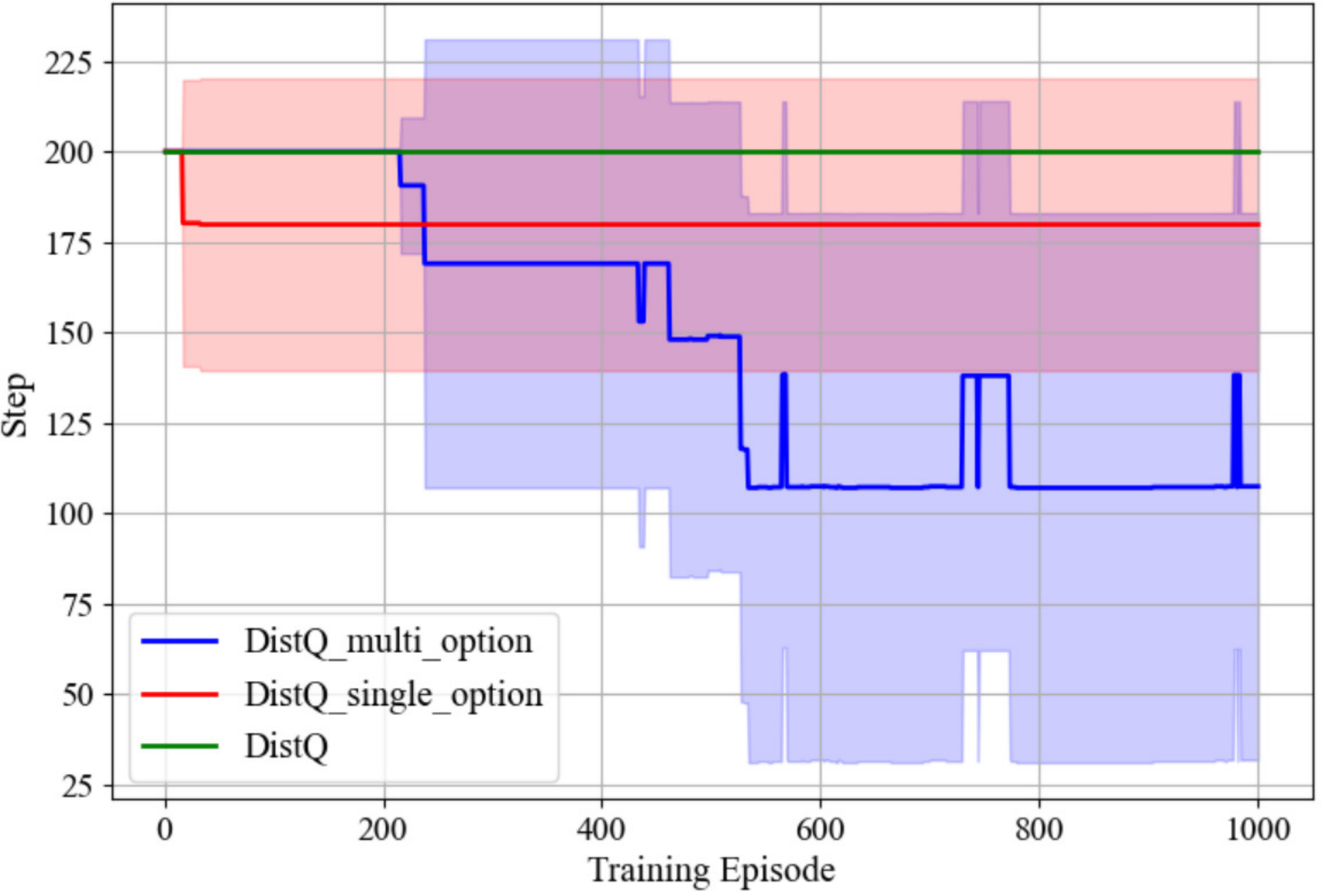}}
\subfigure[ 6 agents]{
\label{fig:4(b)} 
\includegraphics[width=1.31in, height=0.85in]{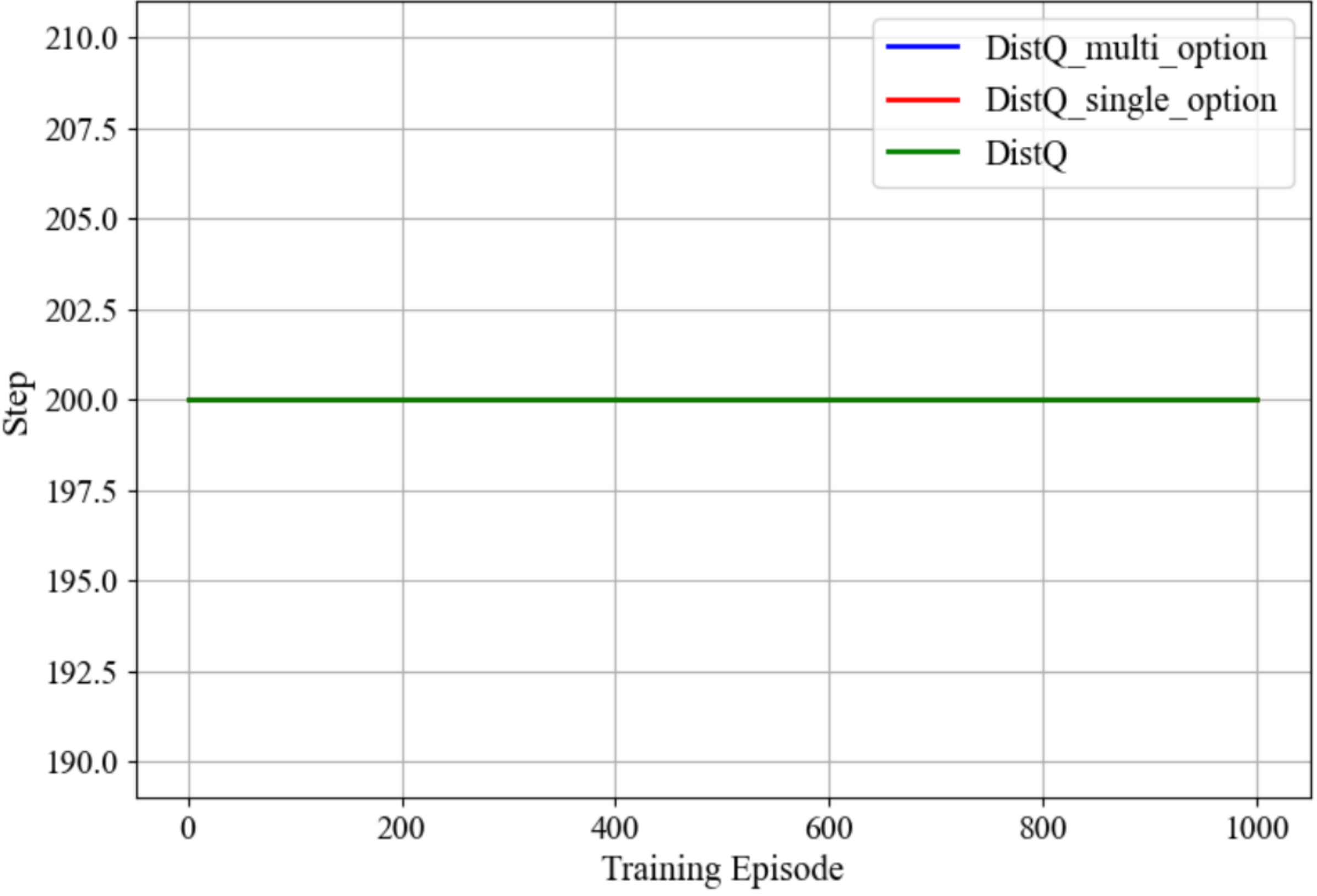}}
\subfigure[4 agents]{
\label{fig:4(d)} 
\includegraphics[width=1.31in, height=0.85in]{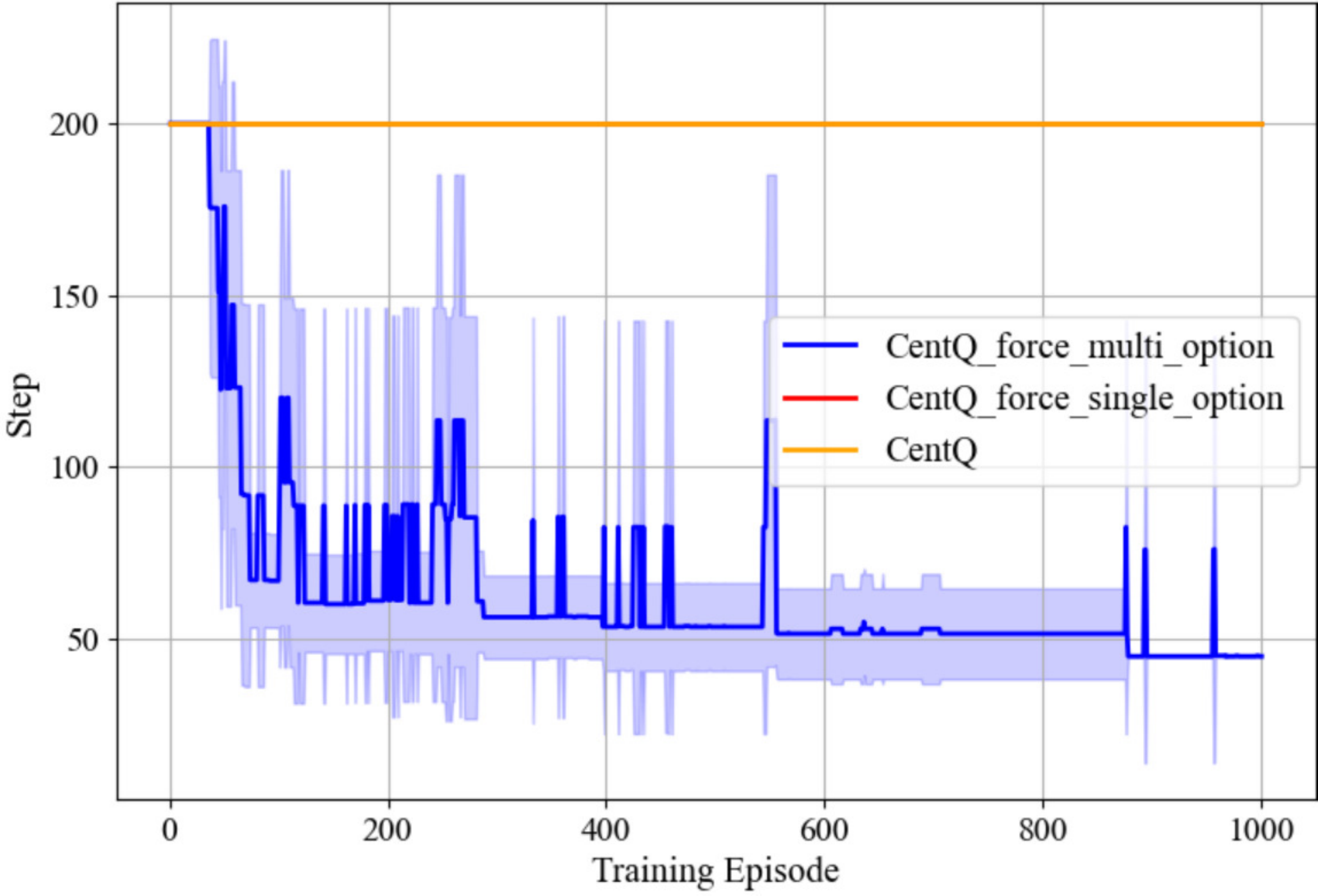}}
\subfigure[6 agents]{
\label{fig:4(e)} 
\includegraphics[width=1.31in, height=0.85in]{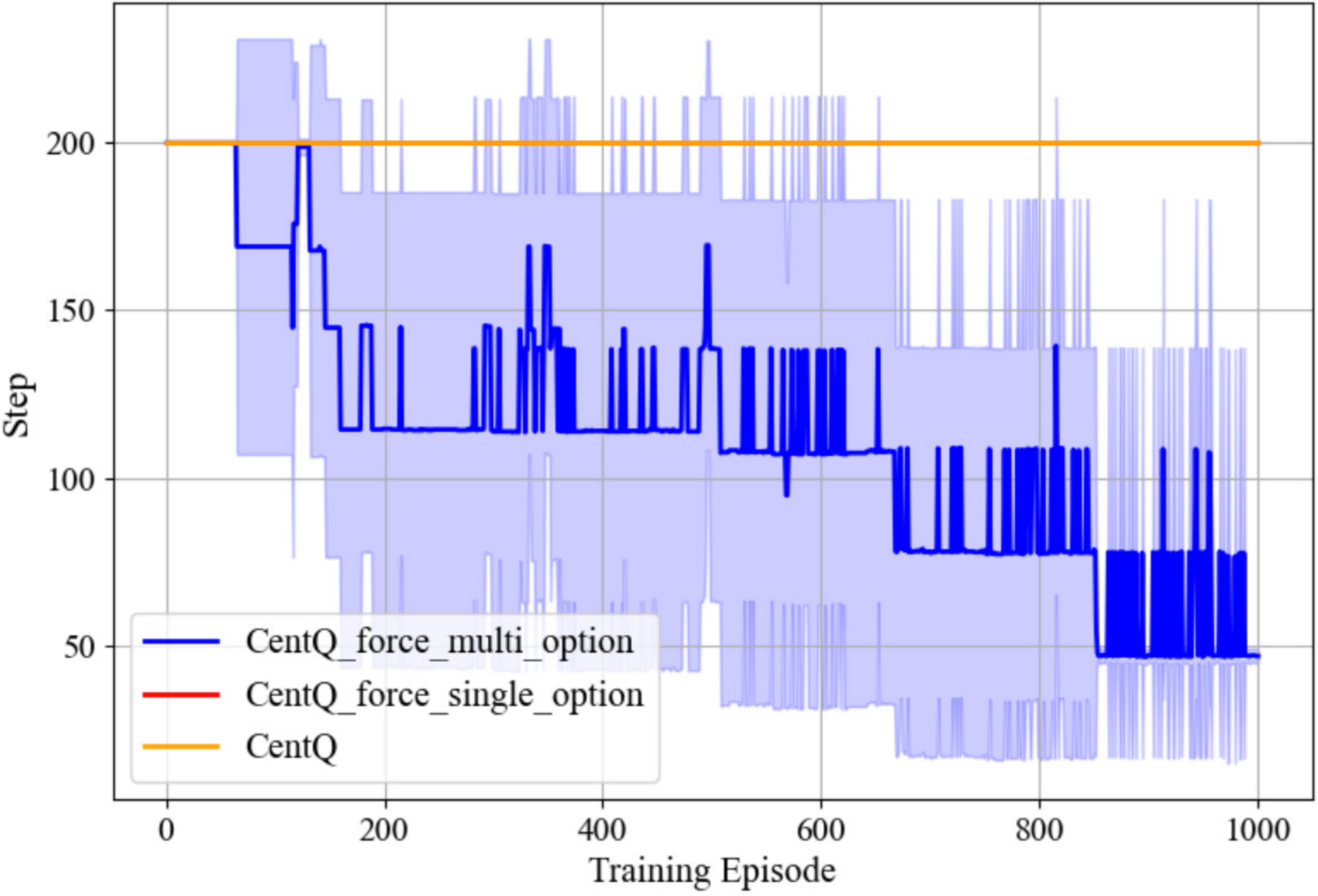}}

\caption{Comparisons on the $n$-agent Grid Maze tasks with random grouping: (a)-(b) Distributed Q-Learning; (c)-(d) Centralized Q-Learning + Force. When $n$-agent options are not available, we can still get a significant performance improvement with only pairwise options. }
\label{fig:4} %% label for entire figure

\end{figure}
% Adopting Centralized Q-Learning + Force can further improve the convergence speed and value.

\begin{figure}[t]

\centering
\subfigure[Grid Room with 4 agents]{
\label{fig:5(b)} 
\includegraphics[width=1.7in, height=0.9in]{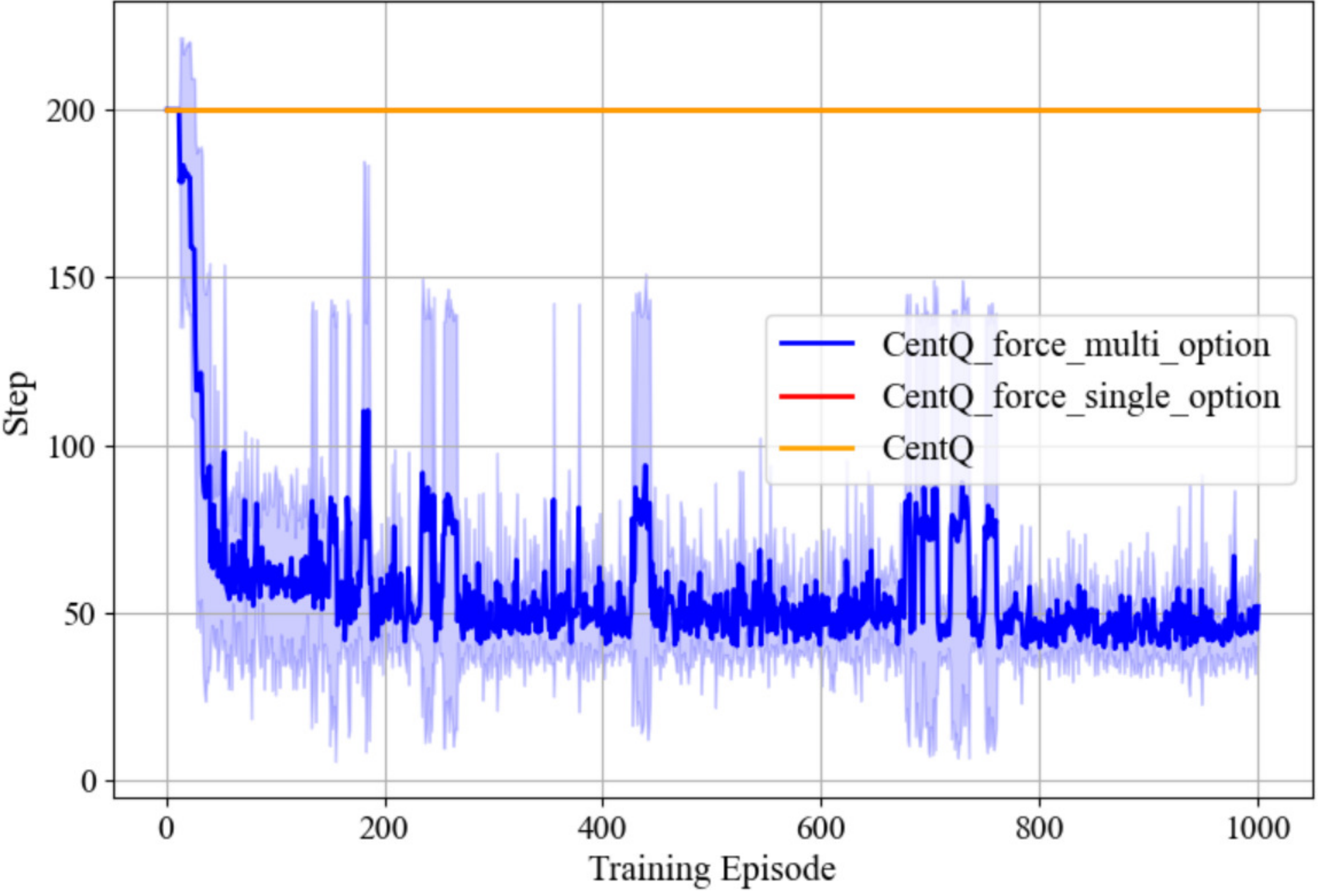}}
\subfigure[Grid Room with 6 agents]{
\label{fig:5(c)} 
\includegraphics[width=1.7in, height=0.9in]{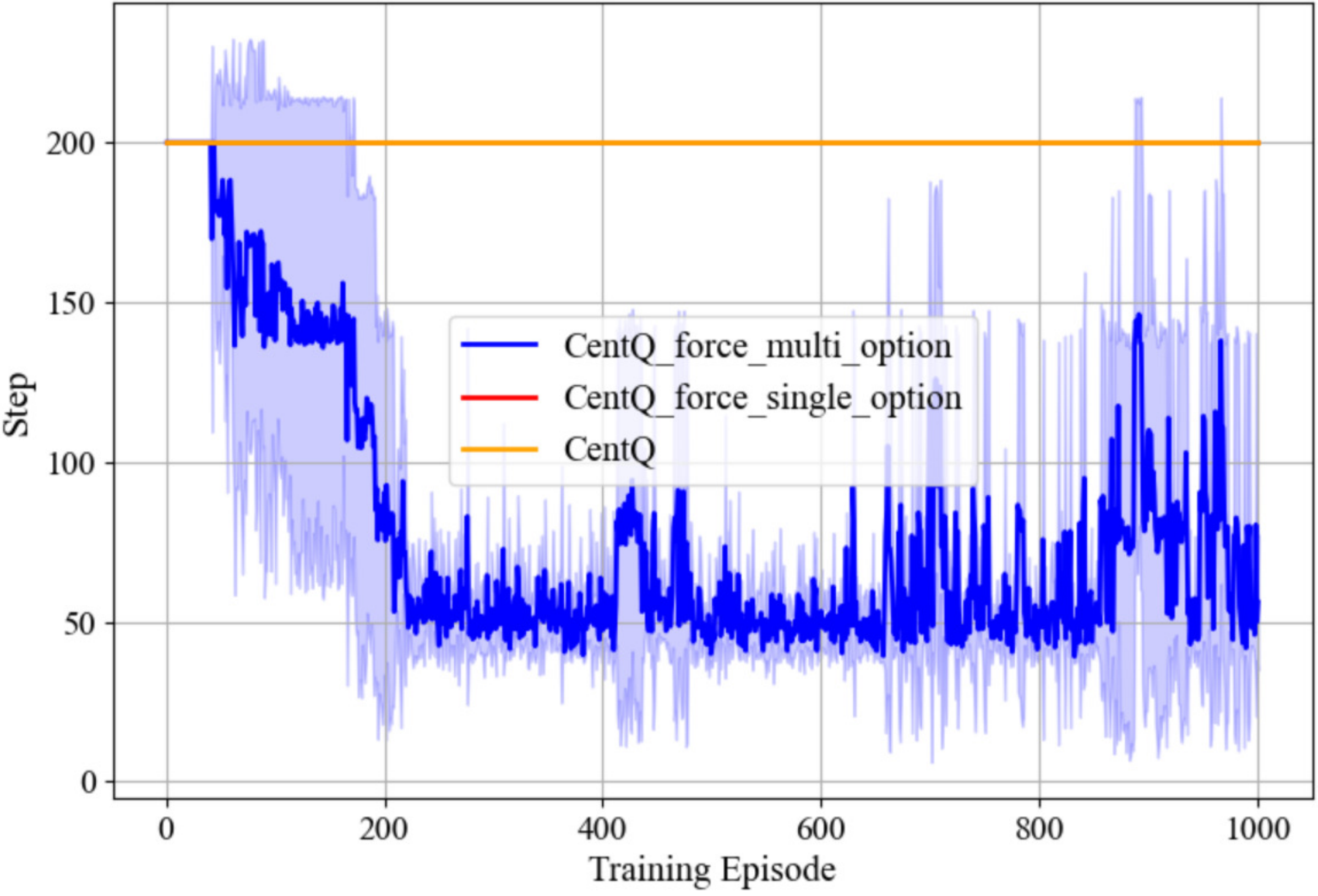}}
\subfigure[Change with collision numbers]{
\label{fig:5(a)} 
\includegraphics[width=1.7in, height=0.9in]{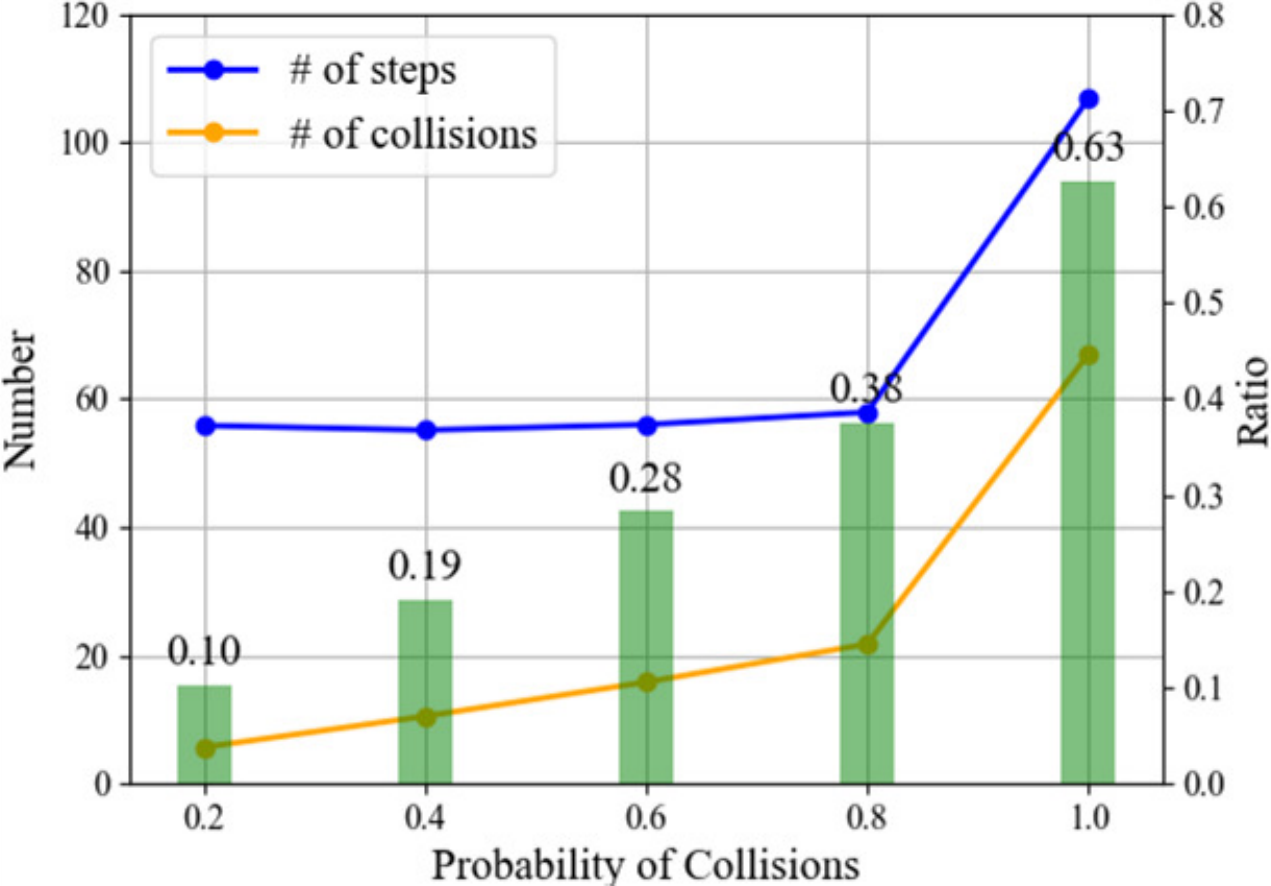}}

\caption{Comparisons on the $n$-agent Grid Room tasks where agent's state transitions can be influenced by the others.}
\label{fig:5} %% label for entire figure

\end{figure}
% We can still obtain good approximations of the multi-agent options based on the theory introduced in Section \ref{theory} and use them to get superior performance.

\subsection{Main Results} \label{results}

For each experiment, we present comparisons among agents with multi-agent options (blue line), single-agent options (red line) and no options. We use the number of time steps to complete the task as the metric (the lower the better). The maximum of steps that an agent can take is 200 for the discrete task and 500 for the continuous task. We run each experiment for five times with different random seeds and plot the mean and standard deviation during the training process. 

\textbf{$n$-agent Grid Maze task:} In Figure \ref{fig:3(b)}-\ref{fig:3(d)}, we test these methods with Distributed Q-learning as the high-level policy. The performance improvement brought by our approach are more and more significant as the number of agents increases. When $n=4$, both the baselines fail to complete the task, while agents with four-agent options can converge within $\sim400$ episodes. While, in Figure \ref{fig:3(e)}-\ref{fig:3(g)}, we show the results of using Centralized Q-Learning + Force. We can see that the centralized manner to utilize $n$-agent options leads to faster convergence, since the joint output space of the agents is pruned (Figure \ref{fig:1}). The results of the $n$-agent Grid Room task can be found in Appendix \ref{n-room}, based on which we can get the same conclusion. To justify the difficulty of the Grid tasks for evaluation, we provide the performance of COMA, MAVEN, Weighted QMIX on the most challenging 4-agent Grid Maze in \ref{n-room}. None of them can learn to complete the coordinated goal-achieving task due to the highly sparse reward setting in our evaluation tasks. Still, our method performs much better (Figure \ref{fig:3(d)}\ref{fig:3(g)}). Given that the neural network training in these baselines requires a longer time, we set the training horizon as 5000 episodes which is five times that of the tabular methods and the networks are trained for ten iterations in each episode, to maintain the fairness.

\textbf{Grid Maze task with sub-task grouping:} The size of the joint state space grows exponentially with the number of agents, making it infeasible to directly construct $n$-agent options for a large $n$. However, in practice, a multi-agent task can usually be divided into sub-tasks, and the agents can be divided into sub-groups based on the sub-tasks they are responsible for. Thus, we test our proposed method on the $m \times n$ Grid Maze tasks shown as Figure \ref{fig:2(c)}, where we divide the agents into $m$ sub-groups, each of which contains $n$ agents with the same goal area (labelled with the same color).  Note that, in the $2\times2$ ($3\times2$) task, we use two-agent (pairwise) options rather than four-agent (six-agent) options, and for the high-level policy, we only use the joint state of the two agents as input to decide on their joint option choice. We can see from Figure \ref{fig:6} that agents with pairwise options can learn to complete the tasks much faster than the baselines (e.g., improved by about two orders of magnitude in Figure \ref{fig:6(c)}). Also, we see that agents trained with Centralized Q-Learning + Force (Figure \ref{fig:6(c)}-\ref{fig:6(d)}) have faster convergence speed and better convergence value. The results for the $m \times n$ Grid Room tasks are in Appendix \ref{group-room}.

\textbf{Grid Maze task with random grouping:} Further, we note that our method also works with random grouping when sub-task grouping may not work. The intuition is that adopting two-agent or three-agent options can encourage the joint exploration of the agents in small sub-groups, which can increase the overall performance compared with only utilizing single-agent explorations. As shown in Figure \ref{fig:4}, we compare the performance of agents using pairwise options with the baselines on the $n$-agent Grid Maze tasks.  When $n=6$, agents with single-agent options or no options can't complete this task, while we can get a significant performance improvement with only pairwise options. On the other hand, agents with pairwise options can't complete the most challenging six-agent task, when using Distributed Q-Learning (Figure \ref{fig:4(b)}). However, if we adopt Centralized Q-Learning + Force, agents with pairwise options can still complete this challenging task with satisfaction (Figure \ref{fig:4(e)}). The same set of results for the Grid Room task are in Appendix \ref{pair-room}.

\textbf{Grid Room task with random grouping and dynamic influences among agents:} Further, we show that even if in environments where an agent's state transitions can be strongly influenced by the others, we can still obtain good approximations of the multi-agent options to encourage joint exploration using Theorem \ref{thm:2}. For this new setting, different agents cannot share the same grid so that an agent may be blocked (controlled with a probability parameter) by others when moving ahead, and this influence is highly dynamic. We use the Centralized Q-Learning + Force as the high-level policy, of which the results are shown as Figure \ref{fig:5}. Compared to the results shown in Appendix \ref{pair-room}, this modification leads to slight instability in the training process, but we can still get significant performance improvement with the identified pairwise options. Especially, in \ref{fig:5(a)}, our solution still perform well (i.e., reach the target in $60-100$ steps) in cases where the collision influence is heavy (i.e., collisions occur in up to 63\% of the steps).
% we make some modifications on the $n$-agent Grid Room --

\begin{wrapfigure}{r}{6.1cm}
\vspace{-.15in}
\centering
\subfigure[Mujoco Room]{
\label{fig:11(a)} 
\includegraphics[width=1.8in, height=0.85in]{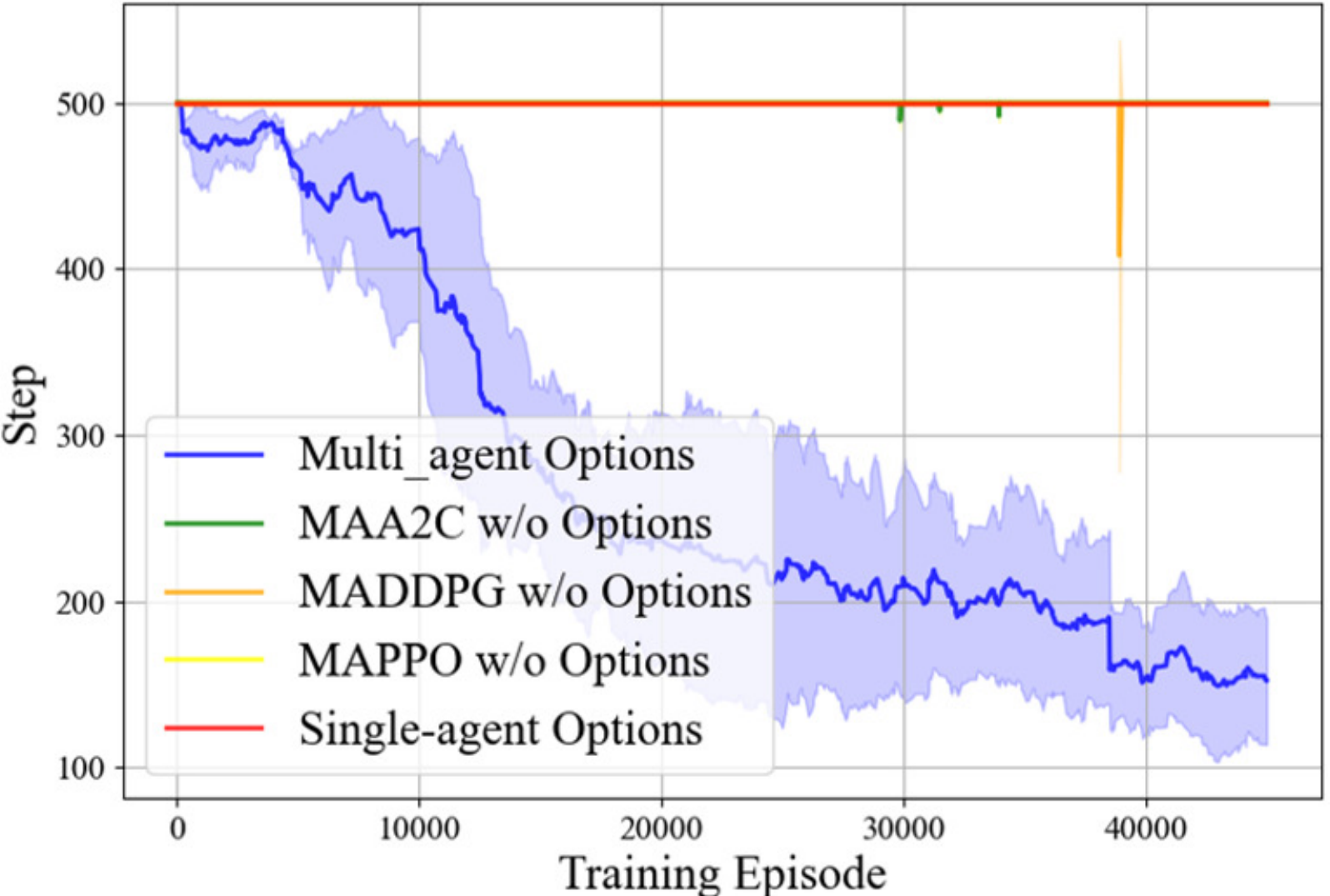}}

\subfigure[Mujoco Maze]{
\label{fig:11(b)} 
\includegraphics[width=1.8in, height=0.85in]{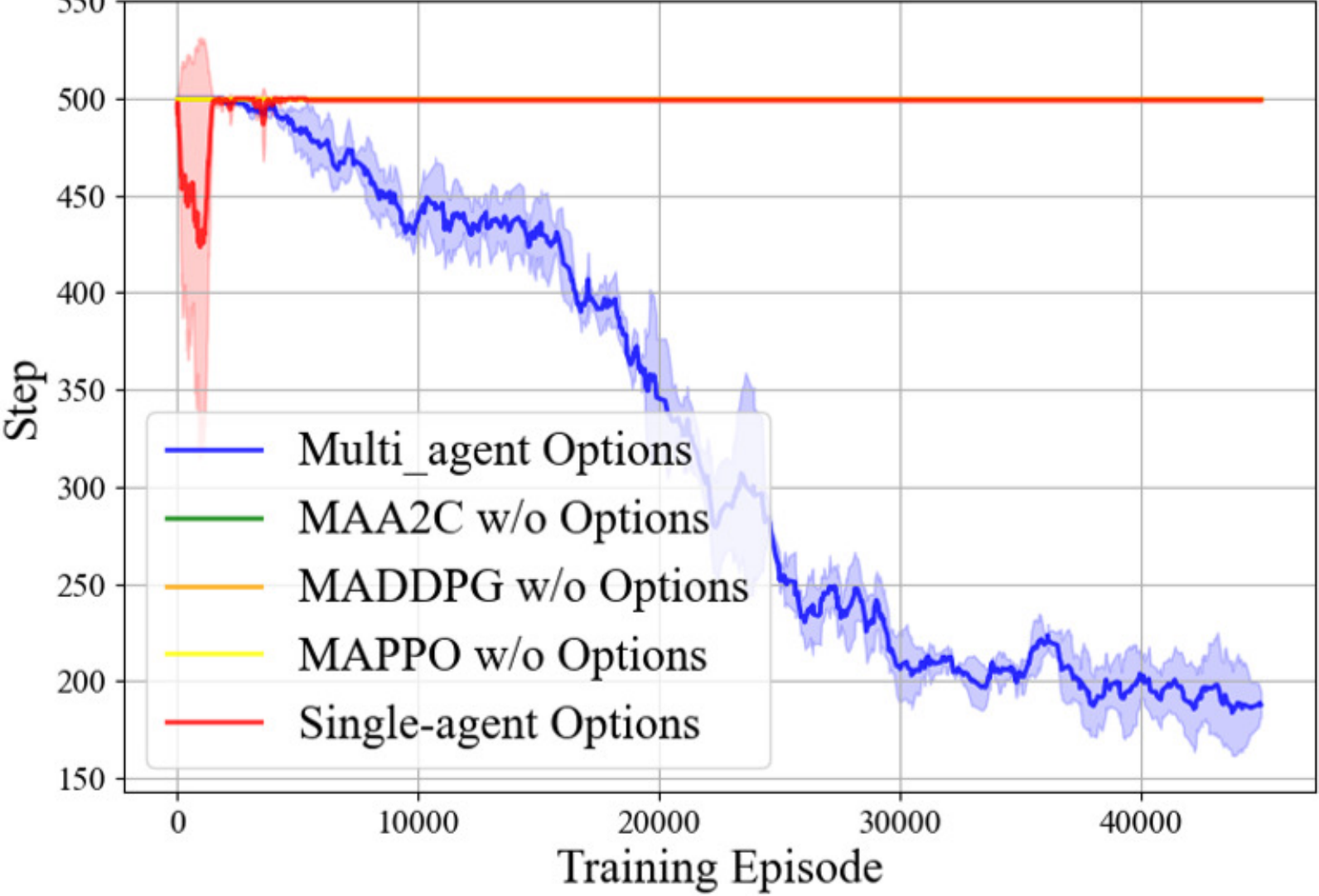}}

\caption{Comparisons on Mujoco tasks.}
\label{fig:11} %% label for entire figure
\vspace{-.1in}
\end{wrapfigure}

\textbf{Mujoco continuous control tasks:} Finally, in order to show the effectiveness and generality of the deep learning extension of our approach (introduced in Section \ref{smacod}) on tasks with infinite-scale state space, we compare it with the baselines on the Mujoco tasks shown as Figure \ref{fig:2(d)} and \ref{fig:2(e)}, of which the results are shown in Figure \ref{fig:11}. On one hand, the comparison between our method and MARL (including MAPPO, MADDPG, MAA2C) without using options shows that abstracting some control series into options and integrating them with a hierarchical training framework can reduce the learning difficulty of continuous control tasks. On the other hand, the comparison with agents using single-agent options shows that discovery and centralized adoption of multi-agent options can aid the exploration in their joint state space, and lead them toward the ``bottleneck" joint states which are essential to complete the cooperation tasks. Note that the options adopted are pre-trained with SAC for $1\times10^4$ episodes, so we have pretrained the baselines: MAA2C, MADDPG, and MAPPO, for the same number of episodes to maintain fairness.

\section{Conclusion And Future Works} \label{conc}

In this paper, we propose to approximate the joint state space in MARL as a Kronecker graph and estimate its Fiedler vector using the Laplacian spectrum of the individual agents' state transition graphs. Based on the approximation of the Fiedler vector, multi-agent covering options are constructed, containing multiple agents’ temporal action sequence towards the sub-goal joint states which are usually infrequently visited, so as to accelerate the joint exploration in the environment. Moreover, we propose a scalable extension of our algorithm so that it can work on tasks with large-scale or continuous state spaces, in order to improve the generality. Further, we propose algorithms to adopt these multi-agent options in MARL, using centralized, decentralized, and group-based strategies, respectively. We empirically show that agents with multi-agent options have significantly superior performance than agents relying on single-agent options or no options. 

One limitation of our method is that there will be non-negligible differences between our approximation $\otimes_{i=1}^{n}G_{i}$ and the true joint state transition graph $\widetilde{G}$, if the state transitions of an agent are hugely influenced by the others. Therefore, developing a specific approximation error bound as a supplement to Theorem 4.1 can be an interesting future direction. On the other hand, to utilize the multi-agent options, agents need to share their views (i.e., observations) since the initiation and termination conditions of the multi-agent options are defined based on the joint observations. Communication among the agents is required in cases where the joint observations are not directly available. Thus, integrating our method with efficient communication strategies in MARL \cite{DBLP:conf/nips/FoersterAFW16} to mitigate the communication complexity can be meaningful future works.

%%%%%%%%%%%%%%%%%%%%%%%%%%%%%%%%%%%%%%%%%%%%%%%%%%%%%%%%%%%%
\clearpage
\bibliography{references}

\begin{thebibliography}{10}

\bibitem{DBLP:journals/ai/SuttonPS99}
Sutton, R.~S., D.~Precup, S.~P. Singh.
\newblock Between mdps and semi-mdps: {A} framework for temporal abstraction in
  reinforcement learning.
\newblock \emph{Artificial Intelligence}, 112(1-2):181--211, 1999.

\bibitem{DBLP:conf/icml/JinnaiAHLK19}
Jinnai, Y., D.~Abel, D.~E. Hershkowitz, M.~L. Littman, G.~D. Konidaris.
\newblock Finding options that minimize planning time.
\newblock In \emph{Proceedings of the 36th International Conference on Machine
  Learning, {ICML} 2019}, vol.~97 of \emph{Proceedings of Machine Learning
  Research}, pages 3120--3129. {PMLR}, 2019.

\bibitem{DBLP:conf/icml/JinnaiPAK19}
Jinnai, Y., J.~W. Park, D.~Abel, G.~D. Konidaris.
\newblock Discovering options for exploration by minimizing cover time.
\newblock In \emph{Proceedings of the 36th International Conference on Machine
  Learning, {ICML} 2019}, vol.~97 of \emph{Proceedings of Machine Learning
  Research}, pages 3130--3139. {PMLR}, 2019.

\bibitem{fiedler1973algebraic}
Fiedler, M.
\newblock Algebraic connectivity of graphs.
\newblock \emph{Czechoslovak mathematical journal}, 23(2):298--305, 1973.

\bibitem{fast_graphs}
Ghosh, A., S.~Boyd.
\newblock Growing well-connected graphs, 2006.

\bibitem{DBLP:conf/atal/AmatoKK14}
Amato, C., G.~D. Konidaris, L.~P. Kaelbling.
\newblock Planning with macro-actions in decentralized pomdps.
\newblock In \emph{Proceedings of the 13rd International conference on
  Autonomous Agents and Multiagent Systems, {AAMAS}, 2014}, pages 1273--1280.
  {IFAAMAS/ACM}, 2014.

\bibitem{amato2019modeling}
Amato, C., G.~Konidaris, L.~P. Kaelbling, J.~P. How.
\newblock Modeling and planning with macro-actions in decentralized pomdps.
\newblock \emph{Journal of Artificial Intelligence Research}, 64:817--859,
  2019.

\bibitem{DBLP:conf/atal/ChakravortyWRCB20}
Chakravorty, J., P.~N. Ward, J.~Roy, M.~Chevalier{-}Boisvert, S.~Basu, A.~Lupu,
  D.~Precup.
\newblock Option-critic in cooperative multi-agent systems.
\newblock In \emph{Proceedings of the 19th International Conference on
  Autonomous Agents and Multiagent Systems, {AAMAS} 2020}, pages 1792--1794.
  2020.

\bibitem{DBLP:conf/iclr/LeeYL20}
Lee, Y., J.~Yang, J.~J. Lim.
\newblock Learning to coordinate manipulation skills via skill behavior
  diversification.
\newblock In \emph{Proceedings of the 8th International Conference on Learning
  Representations, {ICLR} 2020}. OpenReview.net, 2020.

\bibitem{DBLP:conf/icml/McGovernB01}
McGovern, A., A.~G. Barto.
\newblock Automatic discovery of subgoals in reinforcement learning using
  diverse density.
\newblock In \emph{Proceedings of the 18th International Conference on Machine
  Learning, {ICML} 2001}, pages 361--368. Morgan Kaufmann, 2001.

\bibitem{Menache02q-cut-}
Menache, I., S.~Mannor, N.~Shimkin.
\newblock Q-cut - dynamic discovery of sub-goals in reinforcement learning.
\newblock In \emph{Proceedings of the 13th European Conference on Machine
  Learning, {ECML} 2002}, pages 295--306. Springer, 2002.

\bibitem{DBLP:conf/nips/MankowitzMM16}
Mankowitz, D.~J., T.~A. Mann, S.~Mannor.
\newblock Adaptive skills adaptive partitions {(ASAP)}.
\newblock In \emph{Advances in Neural Information Processing Systems 29, {NIPS}
  2016}, pages 1588--1596. 2016.

\bibitem{Harb2018WhenWI}
Harb, J., P.-L. Bacon, M.~Klissarov, D.~Precup.
\newblock When waiting is not an option : Learning options with a deliberation
  cost.
\newblock In \emph{Proceedings of the 32nd {AAAI} Conference on Artificial
  Intelligence, {AAAI} 2018}. 2018.

\bibitem{stolle2002learning}
Stolle, M., D.~Precup.
\newblock Learning options in reinforcement learning.
\newblock In \emph{International Symposium on abstraction, reformulation, and
  approximation}, pages 212--223. Springer, 2002.

\bibitem{DBLP:conf/icml/SimsekWB05}
Simsek, {\"{O}}., A.~P. Wolfe, A.~G. Barto.
\newblock Identifying useful subgoals in reinforcement learning by local graph
  partitioning.
\newblock In \emph{Proceedings of the 22nd International Conference on Machine
  Learning, {ICML} 2005}, vol. 119, pages 816--823. 2005.

\bibitem{DBLP:conf/nips/SimsekB08}
Simsek, {\"{O}}., A.~G. Barto.
\newblock Skill characterization based on betweenness.
\newblock In \emph{Advances in Neural Information Processing Systems 21, {NIPS}
  2008}, pages 1497--1504. 2008.

\bibitem{DBLP:journals/corr/MachadoBB17}
Machado, M.~C., M.~G. Bellemare, M.~H. Bowling.
\newblock A laplacian framework for option discovery in reinforcement learning.
\newblock \emph{CoRR}, abs/1703.00956, 2017.

\bibitem{DBLP:conf/iclr/MachadoRGLTC18}
Machado, M.~C., C.~Rosenbaum, X.~Guo, M.~Liu, G.~Tesauro, M.~Campbell.
\newblock Eigenoption discovery through the deep successor representation.
\newblock In \emph{Proceedings of the 6th International Conference on Learning
  Representations, {ICLR} 2018}. OpenReview.net, 2018.

\bibitem{DBLP:conf/iclr/JinnaiPMK20}
Jinnai, Y., J.~W. Park, M.~C. Machado, G.~D. Konidaris.
\newblock Exploration in reinforcement learning with deep covering options.
\newblock In \emph{Proceedings of the 8th International Conference on Learning
  Representations, {ICLR} 2020}. OpenReview.net, 2020.

\bibitem{shen2006multi}
Shen, J., G.~Gu, H.~Liu.
\newblock Multi-agent hierarchical reinforcement learning by integrating
  options into maxq.
\newblock In \emph{First international multi-symposiums on computer and
  computational sciences (IMSCCS'06)}, vol.~1, pages 676--682. IEEE, 2006.

\bibitem{DBLP:conf/atal/YangBZ20}
Yang, J., I.~Borovikov, H.~Zha.
\newblock Hierarchical cooperative multi-agent reinforcement learning with
  skill discovery.
\newblock In \emph{Proceedings of the 19th International Conference on
  Autonomous Agents and Multiagent Systems, {AAMAS} 2020}, pages 1566--1574.
  2020.

\bibitem{9847387}
Chen, J., J.~Chen, T.~Lan, V.~Aggarwal.
\newblock Learning multi-agent options for tabular reinforcement learning using
  factor graphs.
\newblock \emph{IEEE Transactions on Artificial Intelligence}, pages 1--13,
  2022.

\bibitem{weichsel1962kronecker}
Weichsel, P.~M.
\newblock The kronecker product of graphs.
\newblock \emph{Proceedings of the American mathematical society},
  13(1):47--52, 1962.

\bibitem{basic2021estimation}
Ba{\v{s}}i{\'c}, M., B.~Arsi{\'c}, Z.~Obradovi{\'c}.
\newblock Another estimation of laplacian spectrum of the kronecker product of
  graphs.
\newblock \emph{Information Sciences}, 609:605--625, 2022.

\bibitem{DBLP:conf/iclr/WuTN19}
Wu, Y., G.~Tucker, O.~Nachum.
\newblock The laplacian in {RL:} learning representations with efficient
  approximations.
\newblock In \emph{Proceedings of the 7th International Conference on Learning
  Representations, {ICLR} 2019}. OpenReview.net, 2019.

\bibitem{DBLP:conf/icml/WangZZSHF21}
Wang, K., K.~Zhou, Q.~Zhang, J.~Shao, B.~Hooi, J.~Feng.
\newblock Towards better laplacian representation in reinforcement learning
  with generalized graph drawing.
\newblock In \emph{Proceedings of the 38th International Conference on Machine
  Learning, {ICML} 2021}, vol. 139 of \emph{Proceedings of Machine Learning
  Research}, pages 11003--11012. {PMLR}, 2021.

\bibitem{chung1997spectral}
Chung, F.~R., F.~C. Graham.
\newblock \emph{Spectral graph theory}.
\newblock 92. American Mathematical Society, 1997.

\bibitem{DBLP:conf/nips/VaswaniSPUJGKP17}
Vaswani, A., N.~Shazeer, N.~Parmar, J.~Uszkoreit, L.~Jones, A.~N. Gomez,
  L.~Kaiser, I.~Polosukhin.
\newblock Attention is all you need.
\newblock In \emph{Advances in Neural Information Processing Systems 30, {NIPS}
  2017}, pages 5998--6008. 2017.

\bibitem{todorov2012mujoco}
Todorov, E., T.~Erez, Y.~Tassa.
\newblock Mujoco: A physics engine for model-based control.
\newblock In \emph{2012 IEEE/RSJ International Conference on Intelligent Robots
  and Systems}, pages 5026--5033. IEEE, 2012.

\bibitem{DBLP:conf/icml/LauerR00}
Lauer, M., M.~A. Riedmiller.
\newblock An algorithm for distributed reinforcement learning in cooperative
  multi-agent systems.
\newblock In \emph{Proceedings of the 17th International Conference on Machine
  Learning, {ICML} 2000}, pages 535--542. Morgan Kaufmann, 2000.

\bibitem{sutton2018reinforcement}
Sutton, R.~S., A.~G. Barto.
\newblock \emph{Reinforcement learning: An introduction}.
\newblock MIT press, 2018.

\bibitem{DBLP:journals/corr/abs-2103-01955}
Yu, C., A.~Velu, E.~Vinitsky, Y.~Wang, A.~M. Bayen, Y.~Wu.
\newblock The surprising effectiveness of {MAPPO} in cooperative, multi-agent
  games.
\newblock \emph{CoRR}, abs/2103.01955, 2021.

\bibitem{DBLP:conf/icml/HaarnojaZAL18}
Haarnoja, T., A.~Zhou, P.~Abbeel, S.~Levine.
\newblock Soft actor-critic: Off-policy maximum entropy deep reinforcement
  learning with a stochastic actor.
\newblock In \emph{Proceedings of the 35th International Conference on Machine
  Learning, {ICML} 2018}, vol.~80 of \emph{Proceedings of Machine Learning
  Research}, pages 1856--1865. {PMLR}, 2018.

\bibitem{DBLP:conf/aaai/FoersterFANW18}
Foerster, J.~N., G.~Farquhar, T.~Afouras, N.~Nardelli, S.~Whiteson.
\newblock Counterfactual multi-agent policy gradients.
\newblock In \emph{Proceedings of the 32nd {AAAI} Conference on Artificial
  Intelligence, {AAAI} 2018}, pages 2974--2982. {AAAI} Press, 2018.

\bibitem{DBLP:conf/nips/MahajanRSW19}
Mahajan, A., T.~Rashid, M.~Samvelyan, S.~Whiteson.
\newblock {MAVEN:} multi-agent variational exploration.
\newblock In \emph{Advances in Neural Information Processing Systems 32, {NIPS}
  2019}, pages 7611--7622. 2019.

\bibitem{DBLP:conf/nips/RashidFPW20}
Rashid, T., G.~Farquhar, B.~Peng, S.~Whiteson.
\newblock Weighted {QMIX:} expanding monotonic value function factorisation for
  deep multi-agent reinforcement learning.
\newblock In \emph{Advances in Neural Information Processing Systems 33, {NIPS}
  2020}. 2020.

\bibitem{DBLP:conf/nips/LoweWTHAM17}
Lowe, R., Y.~Wu, A.~Tamar, J.~Harb, P.~Abbeel, I.~Mordatch.
\newblock Multi-agent actor-critic for mixed cooperative-competitive
  environments.
\newblock In \emph{Advances in Neural Information Processing Systems 30, {NIPS}
  2017}, pages 6379--6390. 2017.

\bibitem{DBLP:conf/nips/PapoudakisC0A21}
Papoudakis, G., F.~Christianos, L.~Sch{\"{a}}fer, S.~V. Albrecht.
\newblock Benchmarking multi-agent deep reinforcement learning algorithms in
  cooperative tasks.
\newblock In \emph{Proceedings of the Neural Information Processing Systems
  Track on Datasets and Benchmarks 1}. 2021.

\bibitem{DBLP:journals/corr/abs-2006-07869}
Papoudakis, G., F.~Christianos, L.~Schafer, S.~V. Albrecht.
\newblock Comparative evaluation of multi-agent deep reinforcement learning
  algorithms.
\newblock \emph{CoRR}, abs/2006.07869, 2020.

\bibitem{DBLP:conf/iclr/EysenbachGIL19}
Eysenbach, B., A.~Gupta, J.~Ibarz, S.~Levine.
\newblock Diversity is all you need: Learning skills without a reward function.
\newblock In \emph{Proceedings of the 7th International Conference on Learning
  Representations, {ICLR} 2019}. OpenReview.net, 2019.

\bibitem{DBLP:conf/nips/FoersterAFW16}
Foerster, J.~N., Y.~M. Assael, N.~de~Freitas, S.~Whiteson.
\newblock Learning to communicate with deep multi-agent reinforcement learning.
\newblock In \emph{Advances in Neural Information Processing Systems 29, {NIPS}
  2016}, pages 2137--2145. 2016.

\bibitem{DBLP:journals/dam/Sayama16}
Sayama, H.
\newblock Estimation of laplacian spectra of direct and strong product graphs.
\newblock \emph{Discrete Applied Mathematics}, 205:160--170, 2016.

\end{thebibliography}
\bibliographystyle{nips}

% \clearpage
% \input{checklist}
\clearpage
\appendix

\section{Proof of Theorem \ref{thm:2}} \label{thmproof}

For convenience, we use $G_i$ to represent the $i$-th factor graph and its adjacency matrix. Also, we denote the number of nodes in $G_i$ as $K_i$ and an identity matrix with $K_i$ diagonal elements as $I_{K_i}$. 
\begin{proof}
The normalized laplacian matrix of the Kronecker product of $n$ factor graphs $\otimes_{i=1}^{n}G_{i}$ can be written as:
\begin{eqnarray}
\mathcal{L}_{\otimes_{i=1}^{n}G_{i}} =& \otimes_{i=1}^{n}I_{K_i}-(\otimes_{i=1}^{n}D_{G_i}^{-\frac{1}{2}}) (\otimes_{i=1}^{n}G_{i}) (\otimes_{i=1}^{n} D_{G_i}^{-\frac{1}{2}}).
\end{eqnarray}
Using the property of the Kronecker product of matrices, $(A\otimes B)(C\otimes D)=AC\otimes BD$, we can obtain that:
\begin{equation}
    \begin{aligned}
    \mathcal{L}_{\otimes_{i=1}^{n}G_{i}} &=\otimes_{i=1}^{n}I_{K_i}-\otimes_{i=1}^{n}(D_{G_i}^{-\frac{1}{2}} G_i D_{G_i}^{-\frac{1}{2}})\\
    &=\otimes_{i=1}^{n}I_{K_i}-\otimes_{i=1}^{n}(I_{K_i}-\mathcal{L}_{G_i}).
    \end{aligned}
\end{equation}

Let $\{\lambda_{k_1}^{G_1}\}, \{\lambda_{k_2}^{G_2}\}, \ldots,  \{\lambda_{k_n}^{G_n}\}$ be the eigenvalues of matrices $\mathcal{L}_{G_1}, \mathcal{L}_{G_2}$ $,\ldots,\mathcal{L}_{G_n}$, with the corresponding orthonormal eigenvectors $\{v_{k_1}^{G_1}\}, \{v_{k_2}^{G_2}\},$ $\ldots,\{v_{k_n}^{G_n}\}$, where $k_i=1,2,\ldots,K_i$. Also, denote the diagonal matrices, whose diagonal elements are the values $\{1-\lambda_{k_1}^{G_1}\}, \{1-\lambda_{k_2}^{G_2}\},\ldots,\{1-\lambda_{k_n}^{G_n}\}$, as $\Lambda_{G_1},\Lambda_{G_2},\ldots,\Lambda_{G_n}$, and the square matrices containing the eigenvectors $\{v_{k_1}^{G_1}\}, \{v_{k_2}^{G_2}\},\ldots, \{v_{k_n}^{G_n}\}$ as the column vectors as $V_{G_1}, V_{G_2},\ldots,V_{G_n}$. Using the spectral decomposition of the matrix $I_{K_i}-\mathcal{L}_{G_i}\ (i=1,\ldots,n)$, we can obtain that:
\begin{equation}
    \begin{aligned}
    \mathcal{L}_{\otimes_{i=1}^{n}G_{i}}&=\otimes_{i=1}^{n}I_{K_i}-\otimes_{i=1}^{n}(V_{G_i}\Lambda_{G_i}V_{G_i}^T)\\
    &=\otimes_{i=1}^{n}I_{K_i}-(\otimes_{i=1}^{n}V_{G_i})(\otimes_{i=1}^{n}\Lambda_{G_i})(\otimes_{i=1}^{n}V_{G_i})^T\\
    &=(\otimes_{i=1}^{n}V_{G_i})(\otimes_{i=1}^{n}I_{K_i}-\otimes_{i=1}^{n}\Lambda_{G_i})(\otimes_{i=1}^{n}V_{G_i})^T,
    \end{aligned}
\end{equation}
since $\otimes_{i=1}^{n}I_{K_i} = \otimes_{i=1}^{n}[(V_{G_i})(V_{G_i})^T]= (\otimes_{i=1}^{n}V_{G_i})(\otimes_{i=1}^{n}V_{G_i})^T$. This implies that $\mathcal{L}_{\otimes_{i=1}^{n}G_{i}}$ has eigenvalues $\{[1-\prod_{i=1}^{n}(1-\lambda_{k_i}^{G_i})]\} $ and corresponding eigenvectors $\{\otimes_{i=1}^{n}v_{k_i}^{G_i}\}$.

Then, we let $\Lambda=\otimes_{i=1}^{n}I_{K_i}-\otimes_{i=1}^{n}\Lambda_{G_i}$ and $D=\otimes_{i=1}^{n}D_{G_i}$. Since the normalized Laplacian could be expressed in terms of Laplacian matrix as $\mathcal{L}=D^{-\frac{1}{2}}LD^{-\frac{1}{2}}$, we can get $L_{\otimes_{i=1}^{n}G_{i}}(\otimes_{i=1}^{n}V_{G_i})=D^{\frac{1}{2}}\mathcal{L}_{\otimes_{i=1}^{n}G_{i}}D^{\frac{1}{2}}(\otimes_{i=1}^{n}V_{G_i})$. By making assumption (used and testified in \cite{basic2021estimation, DBLP:journals/dam/Sayama16}) that $D^{\frac{1}{2}}_{G_i}V_{G_i}\approx V_{G_i}D^{\frac{1}{2}}_{G_i}$, for $i=1,2,\ldots,n$, we can derive that:
\begin{equation}
    \begin{aligned}
    L_{\otimes_{i=1}^{n}G_{i}}(\otimes_{i=1}^{n}V_{G_i})& \approx D^{\frac{1}{2}}\mathcal{L}_{\otimes_{i=1}^{n}G_{i}}(\otimes_{i=1}^{n}V_{G_i})D^{\frac{1}{2}}\\
    &=D^{\frac{1}{2}}\Lambda (\otimes_{i=1}^{n}V_{G_i}) D^{\frac{1}{2}}.
    \end{aligned}
\end{equation}
After applying the same assumption again, we finally obtain that:
\begin{equation} \label{equ:final}
    \begin{aligned}
    L_{\otimes_{i=1}^{n}G_{i}}(\otimes_{i=1}^{n}V_{G_i})&\approx (D \Lambda)(\otimes_{i=1}^{n}V_{G_i}). 
    \end{aligned}
\end{equation}

Based on Equation (\ref{equ:final}), we can get an approximation of the Laplacian spectrum, including the eigenvalues and corresponding eigenvectors, of the Kronecker product of $n$ factor graphs, shown as Theorem \ref{thm:2}.

Next, we will prove that the estimated eigenvalues $\mu_{k_1k_2,\ldots,k_n}$ in Theorem \ref{thm:2} are non-negative. It is obvious that $d_{k_i}^{G_i}$ and $\prod_{i=1}^{n}d_{k_i}^{G_i}$ are non-negative. Then, we need to prove $[1-\prod_{i=1}^{n}(1-\lambda_{k_i}^{G_i})]$ is non-negative. We know that if $\lambda$ is an eigenvalue of a normalized Laplacian matrix, we can get $0 \leq \lambda \leq 2 $. Hence, $-1 \le 1-\lambda_{k_i}^{G_i} \le 1$, for $i=1,2,\ldots,n$. Based on this, we can get that $\left|\prod_{i=1}^{n}(1-\lambda_{k_i}^{G_i})\right| \leq 1$ and thus $[1-\prod_{i=1}^{n}(1-\lambda_{k_i}^{G_i})]$ is non-negative. 
\end{proof}
\clearpage

\section{Basic Conceptions and Notations} \label{notation}

\noindent \textbf{Markov Decision Process (MDP): }The RL problem can be described with an MDP, denoted by $\mathcal{M}=(\mathcal{S}, \mathcal{A}, \mathcal{P}, \mathcal{R}, \mathcal{\gamma})$, where $\mathcal{S}$ is the state space, $\mathcal{A}$ is the action space, $\mathcal{P}:\mathcal{S} \times \mathcal{A} \times \mathcal{S} \rightarrow [0,1]$ is the state transition function, $\mathcal{R}:\mathcal{S} \times \mathcal{A} \rightarrow R^{1}$ is the reward function, and $\mathcal{\gamma} \in (0,1]$ is the discount factor. 

\noindent \textbf{State transition graph in an MDP: }The state transitions in $\mathcal{M}$ can be modelled as a state transition graph $G=(V_G, E_G)$, where $V_G$ is a set of vertices representing the states in $\mathcal{S}$, and $E_G$ is a set of undirected edges representing state adjacency in $\mathcal{M}$. We note that:
\begin{remark} \label{rem:2}
There is an edge between state $s$ and $s'$ (i.e., $s$ and $s'$ are adjacent) if and only if $\exists \  a \in \mathcal{A},\  s.t.\  \mathcal{P}(s,a,s')>0\ \lor \ \mathcal{P}(s',a,s)>0$. 
\end{remark}

The adjacency matrix $A$ of $G$ is an $|\mathcal{S}| \times |\mathcal{S}|$ matrix whose $(i,j)$ entry is 1 when $s_{i}$ and $s_{j}$ are adjacent, and 0 otherwise. The degree matrix $D$ is a diagonal matrix whose entry $(i, i)$ equals the number of edges incident to $s_{i}$. The Laplacian matrix of $G$ is defined as $L=D-A$. Its second smallest eigenvalue $\lambda_{2}(L)$ is called the algebraic connectivity of the graph $G$, and the corresponding normalized eigenvector is called the Fiedler vector \cite{fiedler1973algebraic}. Last, the normalized Laplacian matrix is defined as $\mathcal{L}=D^{-\frac{1}{2}}LD^{-\frac{1}{2}}$.

\section{Finding the Fiedler vector for the illustrative example shown in Figure \ref{fig:-1(a)}} \label{example2}

\noindent(1) Compute the normalized Laplacian matrix of $G_1$ and $G_2$, namely $\mathcal{L}_{1}$ and $\mathcal{L}_{2}$:
\begin{eqnarray}
\mathcal{L}_{1} = \left[ \begin{array}{cc} 1 &-1 \\ -1 & 1 \end{array} \right] \ \ {\rm ,} \ \ \mathcal{L}_{2} =\left[  \begin{array}{cccc} 1 & -\frac{1}{\sqrt{2}} & 0 & 0 \\ -\frac{1}{\sqrt{2}} & 1 & -\frac{1}{2} & 0 \\ 0 & -\frac{1}{2} & 1 & -\frac{1}{\sqrt{2}} \\ 0 & 0 & -\frac{1}{\sqrt{2}} & 1 \end{array}  \right].
\end{eqnarray}
(2) Compute the eigenvalues and eigenvectors of $\mathcal{L}_{1}$ and $\mathcal{L}_{2}$:
\begin{eqnarray}
\lambda_1^{G_1} = 0,  \ \  \lambda_2^{G_1} = 2, \ \ v_{1:2}^{G_1} = \frac{1}{\sqrt{2}} \left[\left[\begin{array}{c} 1 \\ 1 \end{array} \right], \left[\begin{array}{c} -1 \\ 1 \end{array} \right]\right].  
\end{eqnarray}
\begin{eqnarray}
 \lambda_1^{G_2} = 0, \ \ \lambda_2^{G_2} = 0.5, \ \  \lambda_3^{G_2}  = 1.5, \ \  \lambda_4^{G_2}  = 2, \ \ \\ v_{1:4}^{G_2}=  \frac{1}{\sqrt{3}}\left[\left[\begin{array}{c} \frac{1}{\sqrt{2}} \\ 1 \\ 1 \\ \frac{1}{\sqrt{2}} \end{array} \right], \left[  \begin{array}{c} -1 \\  -\frac{1}{\sqrt{2}} \\ \frac{1}{\sqrt{2}} \\ 1 \end{array} \right], \left[\begin{array}{c} 1 \\ -\frac{1}{\sqrt{2}} \\ -\frac{1}{\sqrt{2}} \\ 1 \end{array} \right], \left[  \begin{array}{c} \frac{1}{\sqrt{2}} \\  -1 \\ 1\\ -\frac{1}{\sqrt{2}}  \end{array} \right] \right].
\end{eqnarray}

\noindent(3) Compute the degree list of $G_1$ and $G_2$ (sorted in ascending order), namely $d^{G_{1}}$ and $d^{G_{2}}$:
\begin{eqnarray}
d^{G_{1}} = [1,\ 1]^{T},\ d^{G_{2}} = [1,\ 1,\ 2,\ 2]^{T}.
\end{eqnarray}
(4) According to Theorem \ref{thm:2}, we can get two approximations of the Fiedler vector:
\begin{eqnarray}
v_{11}=v_{1}^{G_{1}} \otimes v_{1}^{G_{2}}=\frac{1}{\sqrt{6}}\left[\frac{1}{\sqrt{2}},\ 1,\ 1,\ \frac{1}{\sqrt{2}},\ \frac{1}{\sqrt{2}},\ 1,\ 1,\ \frac{1}{\sqrt{2}}\right]^{T},
\\
v_{24}=v_{2}^{G_{1}} \otimes v_{4}^{G_{2}}=\frac{1}{\sqrt{6}}\left[-\frac{1}{\sqrt{2}},\ 1,\ -1,\ \frac{1}{\sqrt{2}},\ \frac{1}{\sqrt{2}},\ -1,\ 1,\ -\frac{1}{\sqrt{2}}\right]^{T}.
\end{eqnarray}

\section{Pseudo Code of Multi-agent Covering Option Discovery} \label{pseudo-code}

\begin{algorithm}[htbp]
\caption{Multi-agent Covering Option Discovery}\label{alg:1}
{\begin{algorithmic}[1]
\State \textbf{Input}: \# of agents $n$, list of adjacency matrices $A_{1:n}$, \# of options to generate $tot\_num$
\State \textbf{Output}: list of multi-agent options $\Omega$
\State $\Omega \leftarrow \emptyset$, $cur\_num \leftarrow 0$
\While{$cur\_num < tot\_num$}
\State \textbf{Collect} the degree list $D_{1:n}$ of each individual state transition graph according to $A_{1:n}$
\State \textbf{Obtain} the list of normalized laplacian matrices $\mathcal{L}_{1:n}$ corresponding to $A_{1:n}$
\State \textbf{Calculate} the eigenvalues $U_{i}$ and corresponding eigenvectors $V_{i}$ for each $\mathcal{L}_{i}$ and collect them 
\Statex \quad\ \  as $U_{1:n}$ and $V_{1:n}$
\State \textbf{Obtain} the Fielder vector $F$ of the joint state space using Theorem \ref{thm:2} with $D_{1:n}$, $U_{1:n}$, $V_{1:n}$
\State \textbf{Collect} the list of joint states corresponding to the minimum or maximum in $F$, named $MIN$ 
\Statex \quad\ \ and $MAX$ respectively
\State \textbf{Convert} each joint state $s_{joint}$ in $MIN$ 
and $MAX$ to $(s_{1}, \cdots, s_{n})$, where $s_{i}$ is the corre-
\Statex \quad\ \ sponding individual state of agent $i$, based on the equation: 
\\ \quad\quad$ind(s_{joint}) = ((ind(s_{1})*dim(A_{2})+\cdots+ind(s_{n-1}))*dim(A_{n})+ind(s_{n})$
\Statex \quad\ \ where $dim(A_{i})$ is the dimension of $A_{i}$, $ind(s_{i})$ is the index of $s_{i}$ (indexed from 0) in the 
\Statex \quad\ \ state space of agent $i$
\State \textbf{Generate} a new list of options $\Omega'$ through \textit{GenerateOptions}
\State $\Omega \leftarrow \Omega \cup \Omega'$, $cur\_num \leftarrow cur\_num + len(\Omega')$
\State \textbf{Update} $A_{1:n}$ through \textit{UpdateAdjacencyMatrices}

\EndWhile
\State \textbf{Return} $\Omega$
\State
\Function{\textit{GenerateOptions}}{$MIN$, $MAX$}
\State $\Omega' \leftarrow \emptyset$
\For{$s=(s_{1}, \cdots, s_{n})$ in ($MIN \cup MAX$)}
\State \textbf{Define} the initiation set $I_{\omega}$ as the joint states in the known region of the joint state space
\State \textbf{Define} the termination condition as:
 
\Statex \quad\quad\quad\quad  $ \beta_{\omega}(s_{cur}) \leftarrow \left\{
                \begin{aligned}
                1 &  & if \ (s_{cur}==s) \ or\  (s_{cur}\ is\ unknown) \\
                0 &  & otherwise
                \end{aligned}
                \right.
$
\Statex \quad\quad\quad where $s_{cur}$ is the current joint state
\State\textbf{Train} the intra-option policy $\pi_{\omega}=(\pi_{\omega}^{1}, \cdots, \pi_{\omega}^{n})$, where $\pi_{\omega}^{i}$ maps the individual state of 
\Statex \quad\quad\quad agent $i$ to its action aiming at leading agent $i$ from any state in its initiation set to its 
\Statex \quad\quad\quad termination state $s_{i}$
\State $\Omega' \leftarrow \Omega' \cup \{<I_{\omega}, \pi_{\omega}, \beta_{\omega}>\}$
\EndFor
\State \textbf{Return} $\Omega'$
\EndFunction
\State 
\Function{\textit{UpdateAdjacencyMatrices}}{$MIN$, $MAX$}
\For{$s_{min}=(s_{min}^{1}, \cdots, s_{min}^{n})$ in $MIN$}
    \For{$s_{max}=(s_{max}^{1}, \cdots, s_{max}^{n})$ in $MAX$}
        \For{$i=1$ to $n$}
        \State $A_{i}[ind(s_{min}^{i})][ind(s_{max}^{i})]=1$
        \State $A_{i}[ind(s_{max}^{i})][ind(s_{min}^{i})]=1$
        \EndFor
    \EndFor
\EndFor
\EndFunction

\end{algorithmic}}
\end{algorithm}

\clearpage

\section{Additional Evaluation Results}
\subsection{$n$-agent Grid Room Task} \label{n-room}

\begin{figure}[htbp]
\centering
\subfigure[Grid Room with 2 agents]{
\label{fig:a1(b)} 
\includegraphics[width=1.7in, height=1.1in]{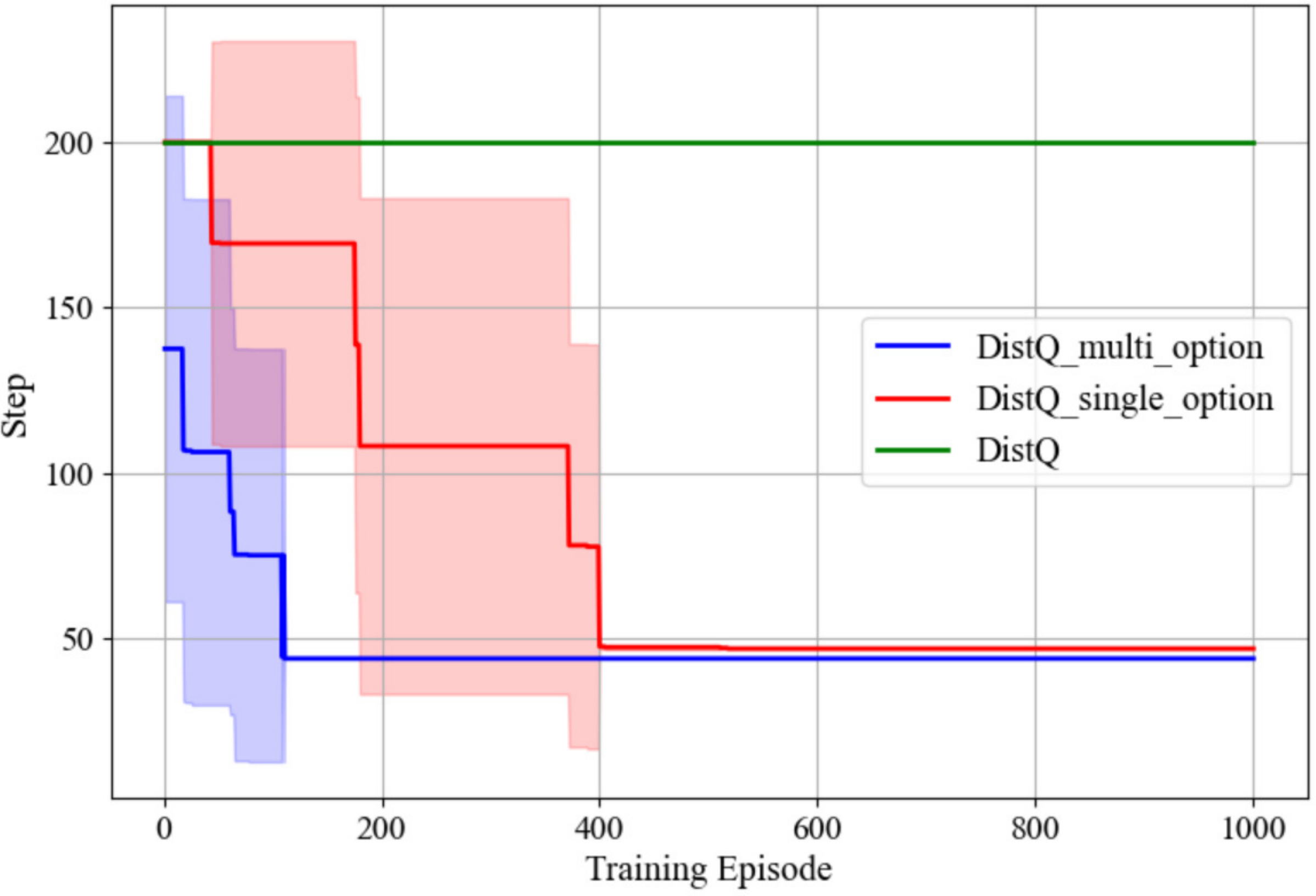}}
\subfigure[Grid Room with 3 agents]{
\label{fig:a1(c)} 
\includegraphics[width=1.7in, height=1.1in]{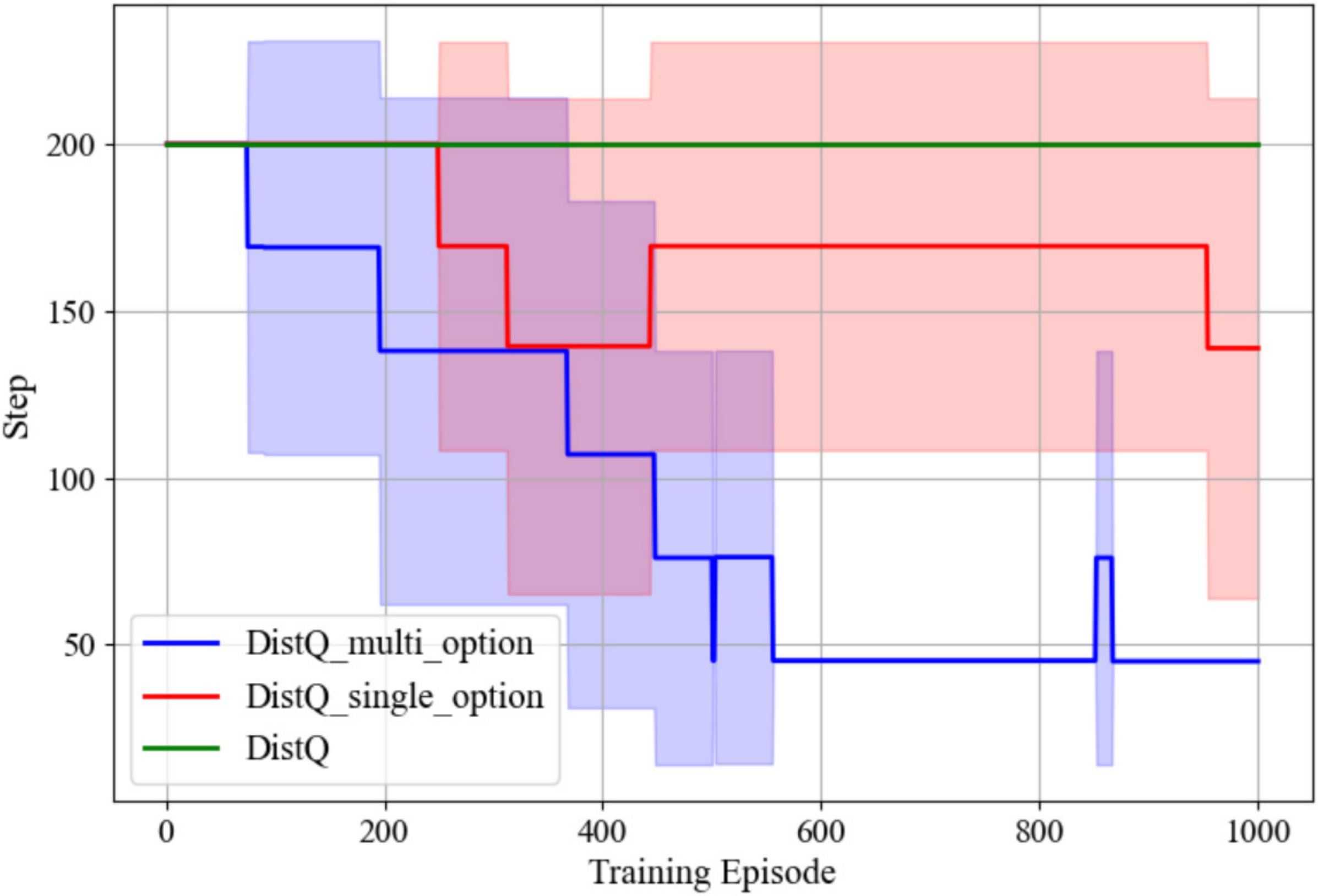}}
\subfigure[Grid Room with 4 agents]{
\label{fig:a1(d)} 
\includegraphics[width=1.7in, height=1.1in]{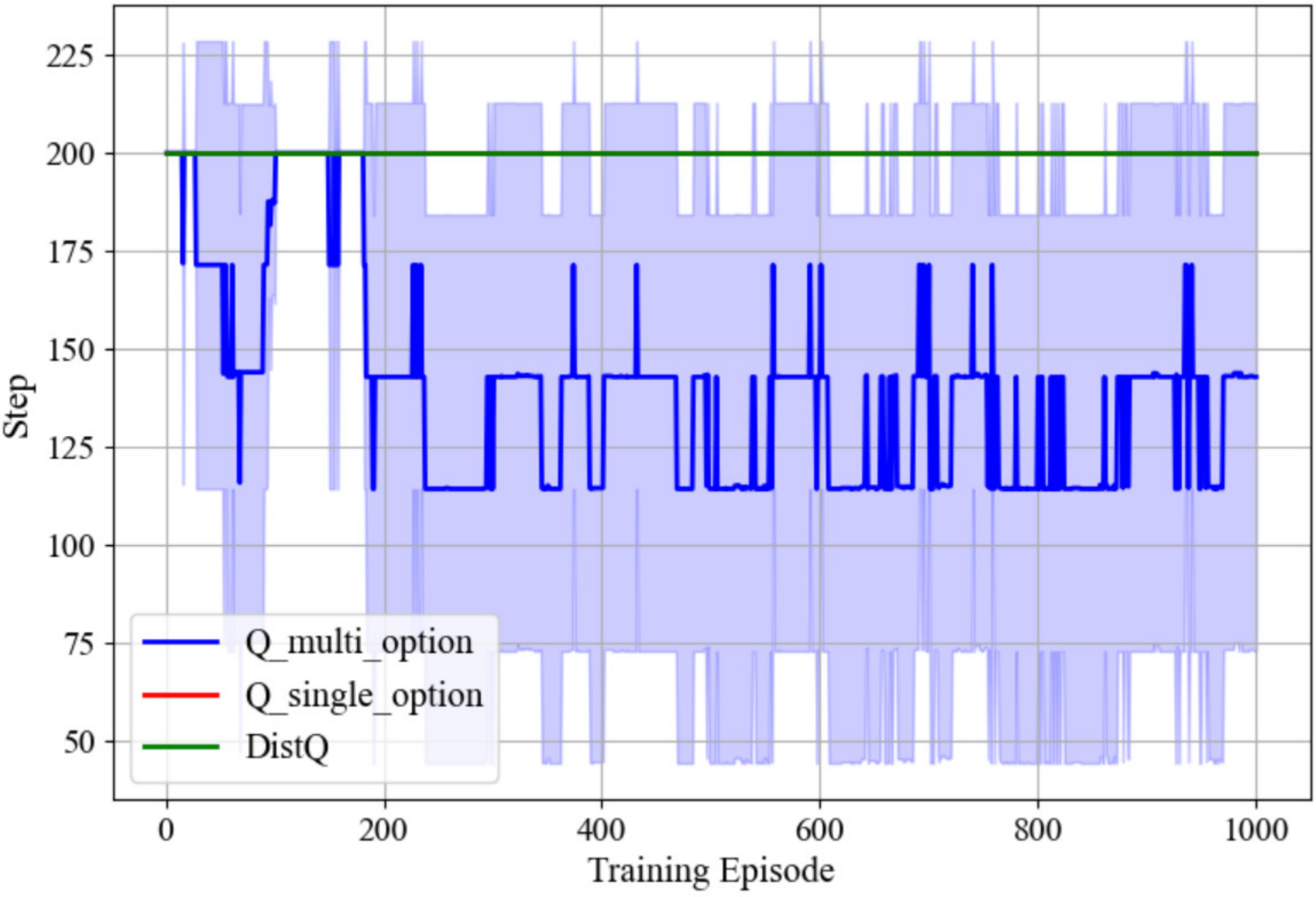}}
\subfigure[Grid Room with 2 agents]{
\label{fig:a1(e)} 
\includegraphics[width=1.7in, height=1.1in]{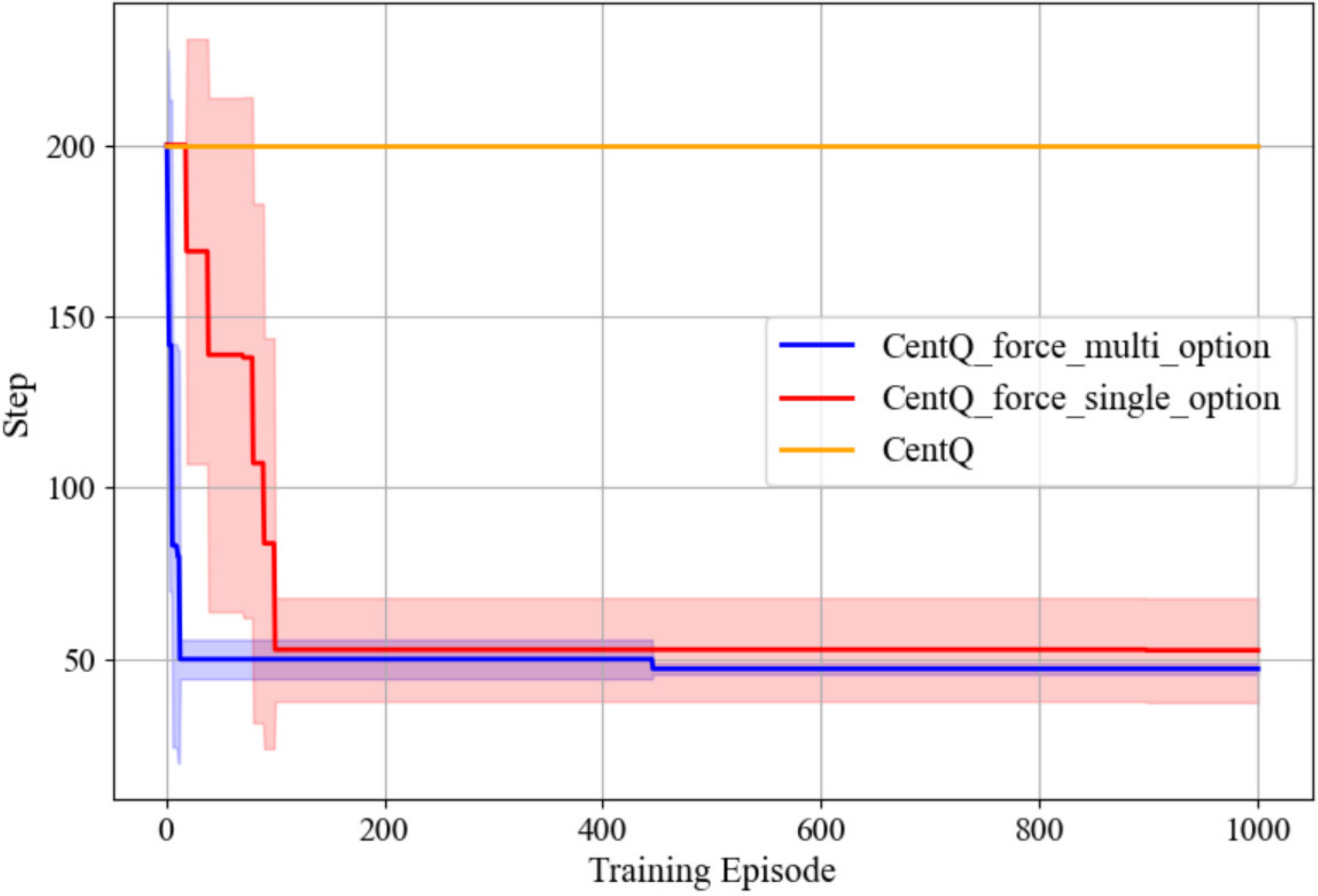}}
\subfigure[Grid Room with 3 agents]{
\label{fig:a1(f)} 
\includegraphics[width=1.7in, height=1.1in]{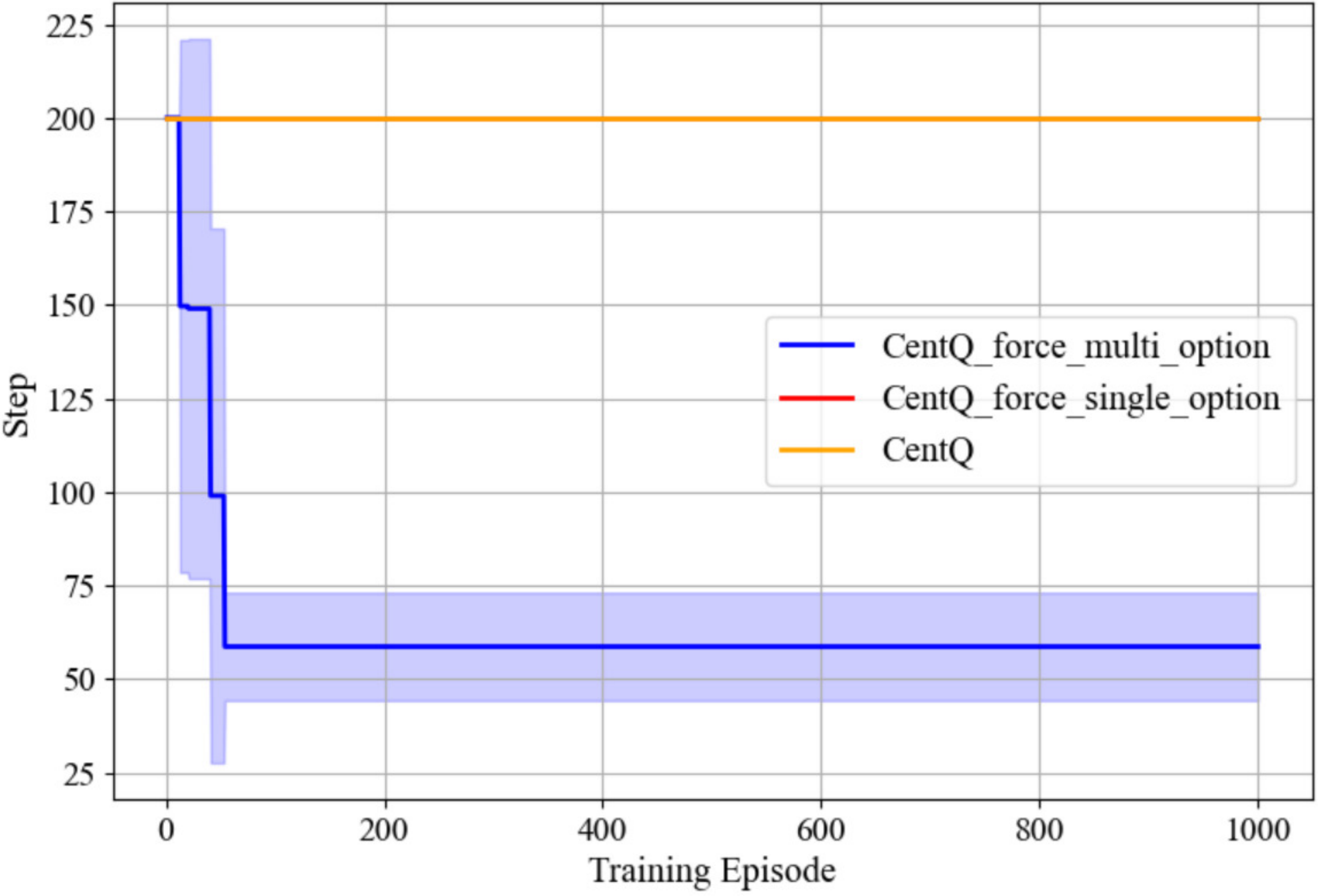}}
\subfigure[Grid Room with 4 agents]{
\label{fig:a1(g)} 
\includegraphics[width=1.7in, height=1.1in]{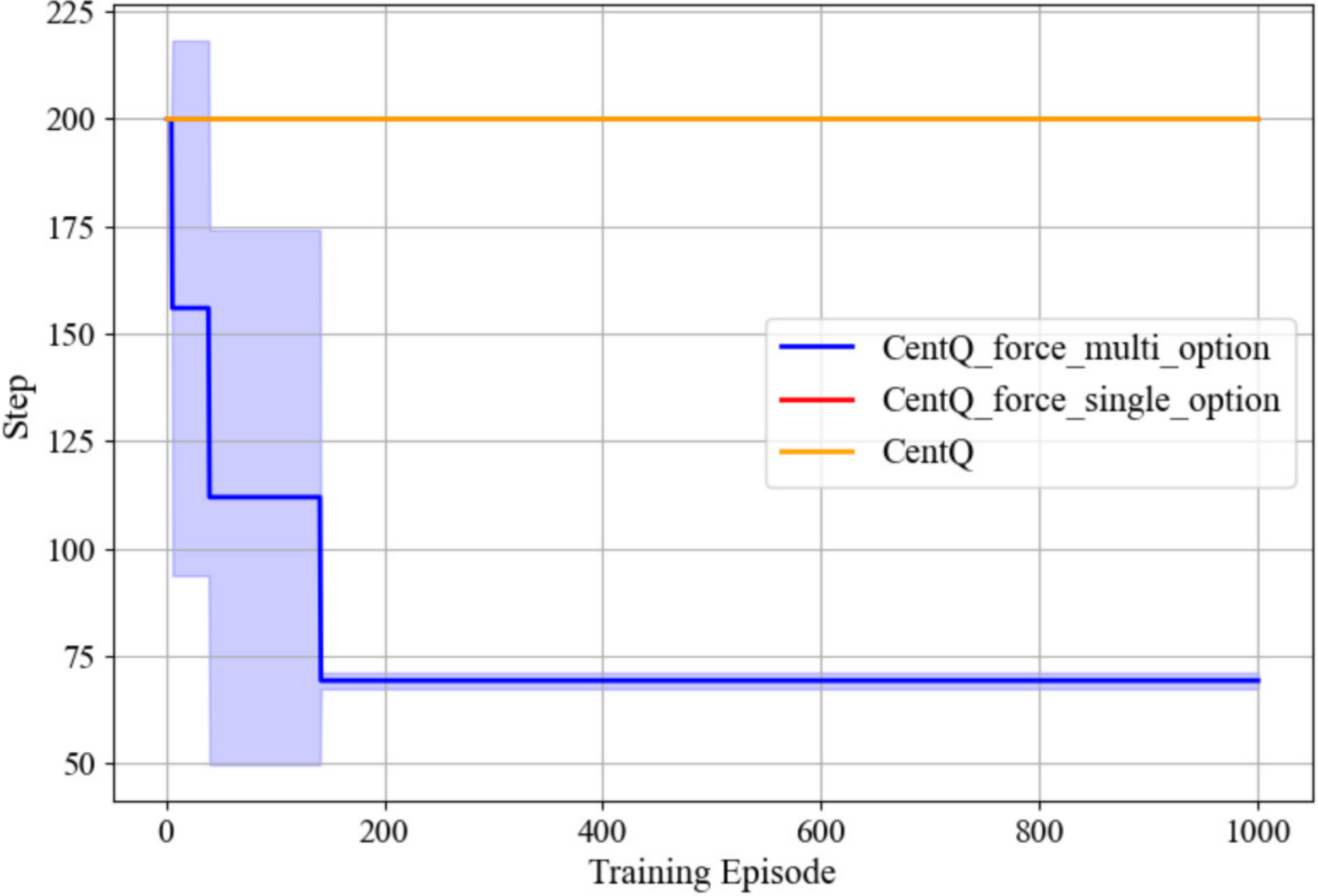}}
\caption{Evaluation on $n$-agent Grid Room tasks: (a)-(c) using Distributed Q-Learning as the high-level policy. The performance improvement of our approach is more significant as the number of agents increases. (d)-(f) using Centralized Q-Learning + Force as the high-level policy. Agents with single-agent options start to fail since the 3-agent case. Also, it can be observed that the centralized way to utilize the $n$-agent options leads to faster convergence.}
\label{fig:a1} 
\end{figure}

\begin{figure}[htbp]
\centering
\includegraphics[width=2.2in, height=1.2in]{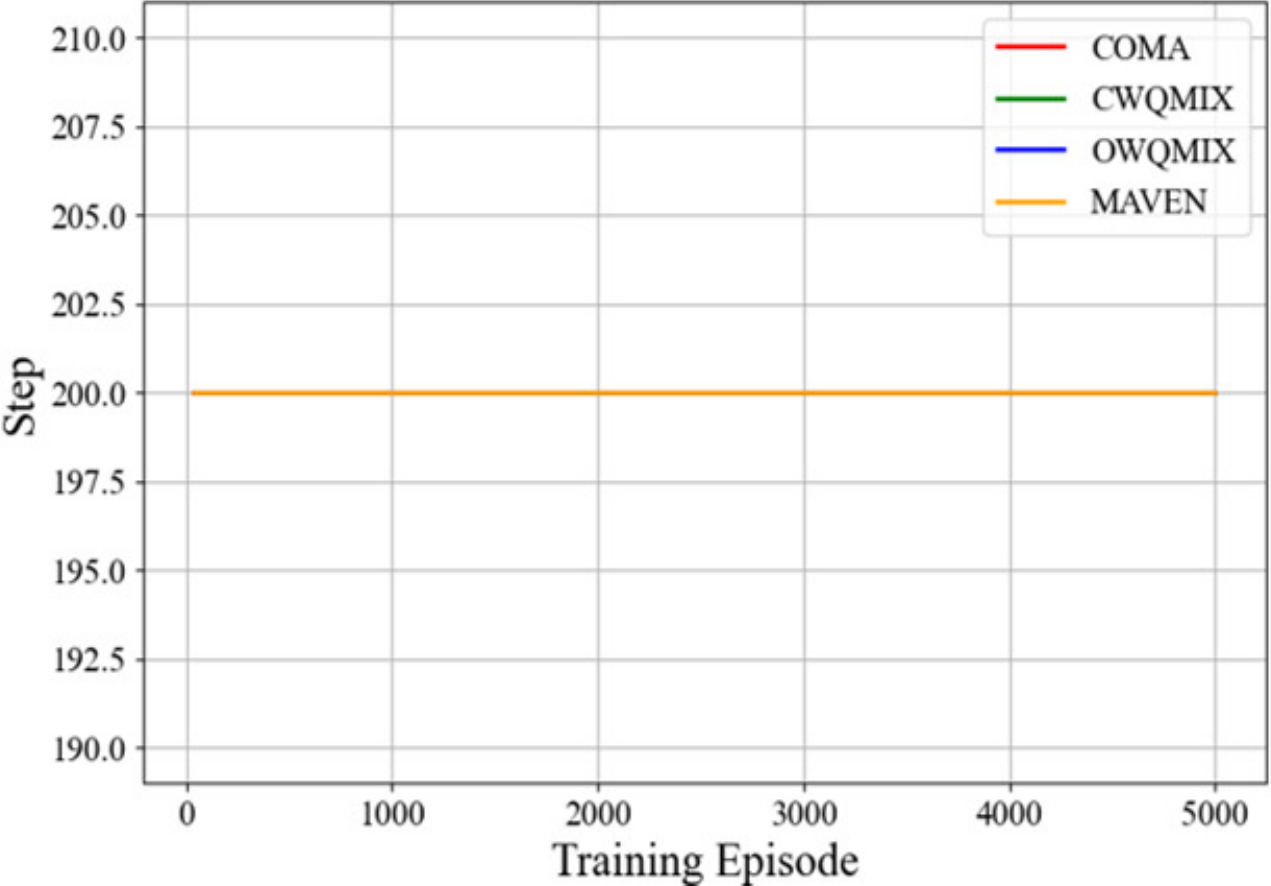}
\caption{Performance of SOTA MARL algorithms: COMA, MAVEN, Weighted QMIX, on the 4-agent Grid Maze Task (Figure \ref{fig:3(d)}\ref{fig:3(g)}). For each algorithm, the experiment is repeated three times with different random seeds (codes are available in the provided link). On this discrete problem setting, these SOTA algorithms do not show better performance than the tabular Q-learning we use as baselines. Also, our method performs much better on the same task.}
\label{fig:a11} 
\end{figure}

\clearpage

\subsection{$m \times n$-agent Grid Room Task} \label{group-room}

\begin{figure}[htbp]
\centering
\subfigure[2$\times$2 agents]{
\label{fig:a2(a)} 
\includegraphics[width=2.2in, height=1.0in]{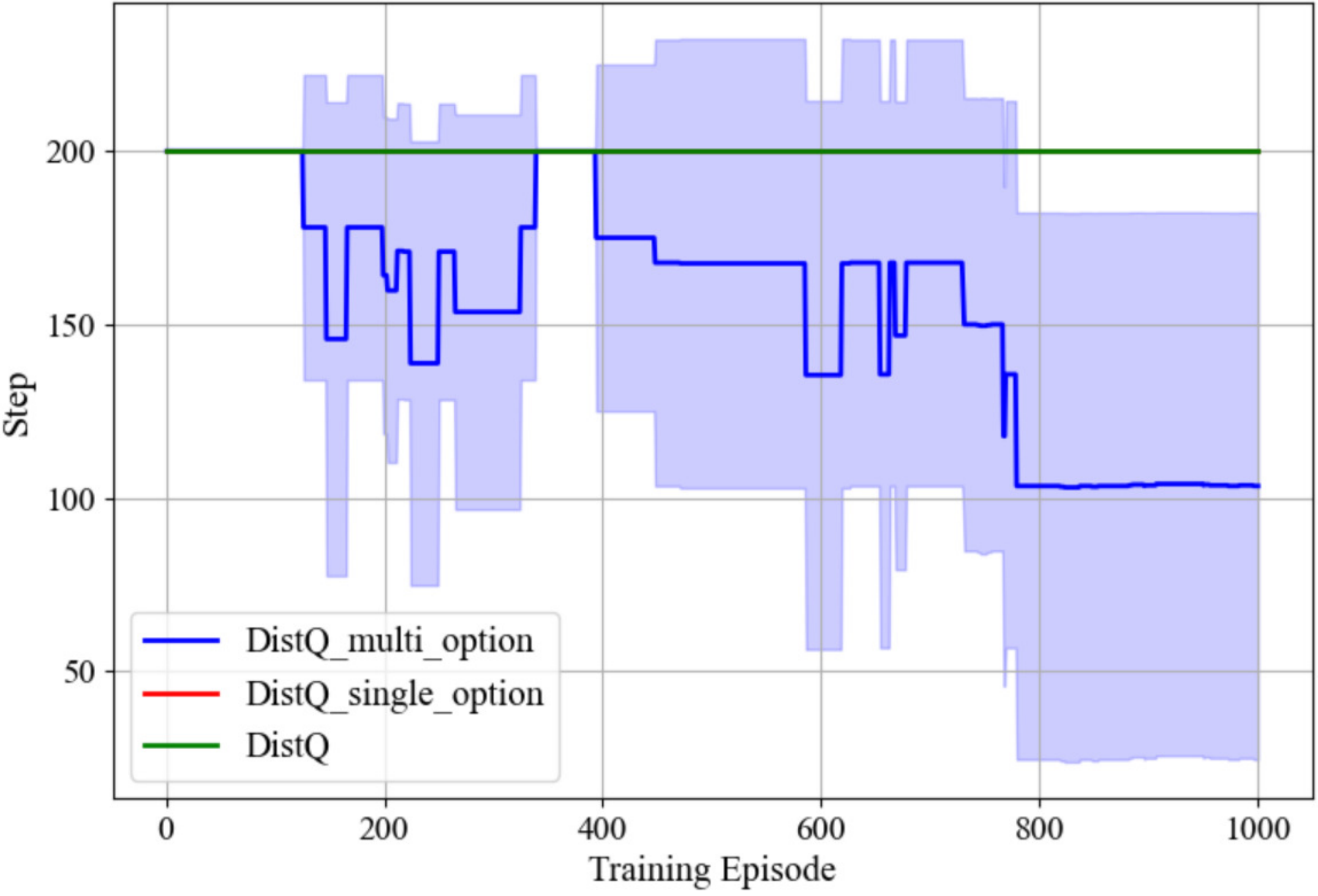}}
\subfigure[3$\times$2 agents]{
\label{fig:a2(b)} 
\includegraphics[width=2.2in, height=1.0in]{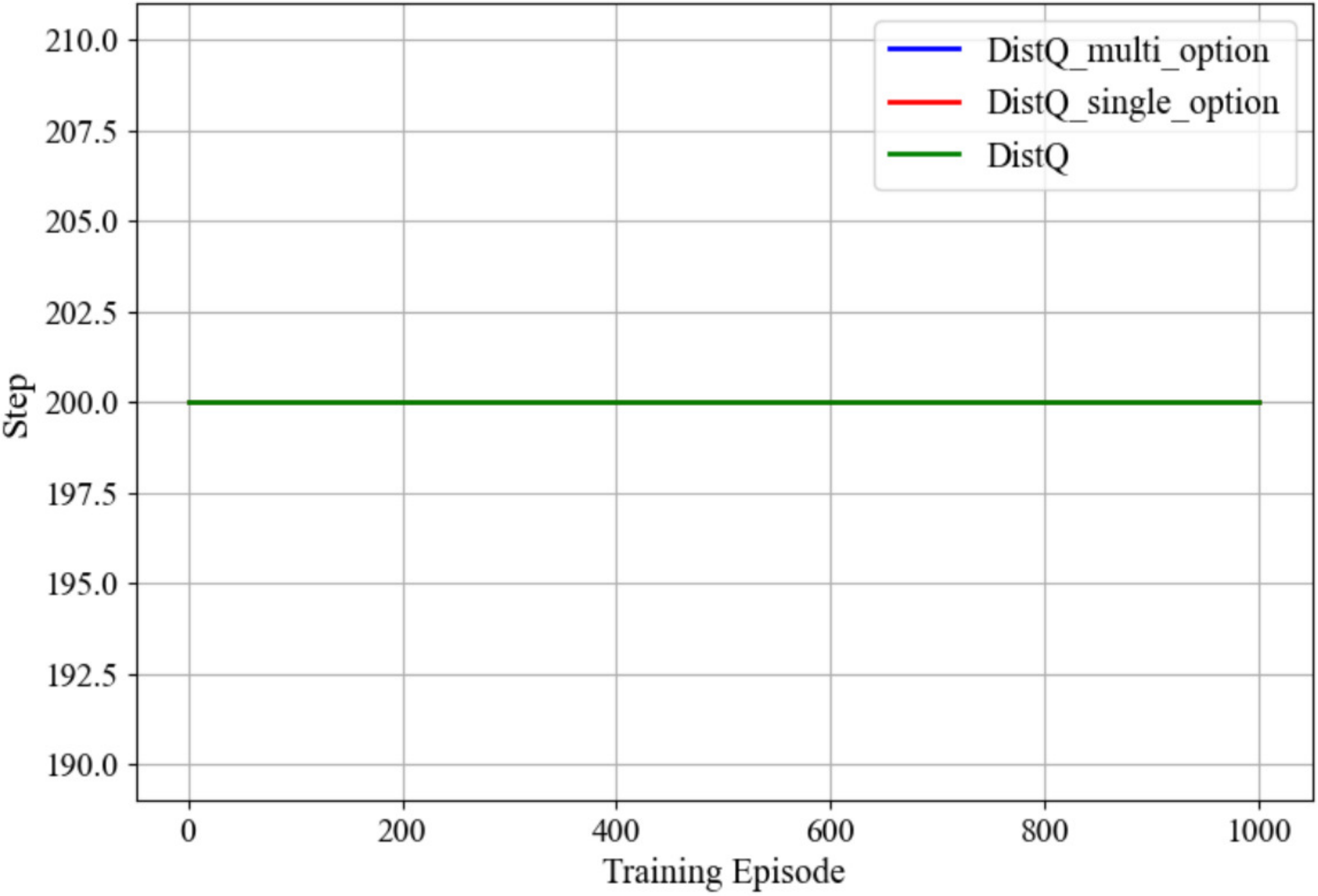}}
\subfigure[2$\times$2 agents]{
\label{fig:a2(c)} 
\includegraphics[width=2.2in, height=1.0in]{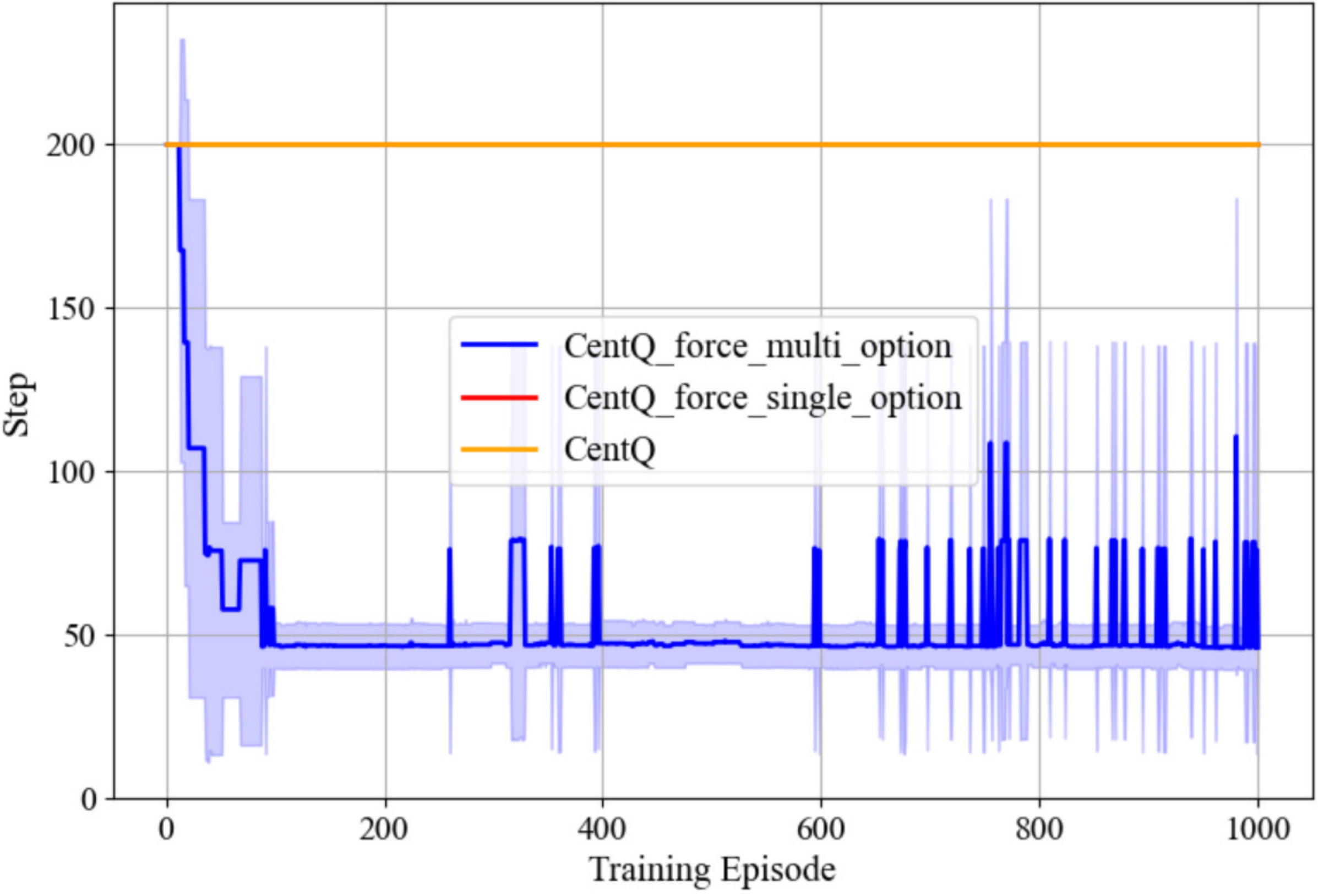}}
\subfigure[3$\times$2 agents]{
\label{fig:a2(d)} 
\includegraphics[width=2.2in, height=1.0in]{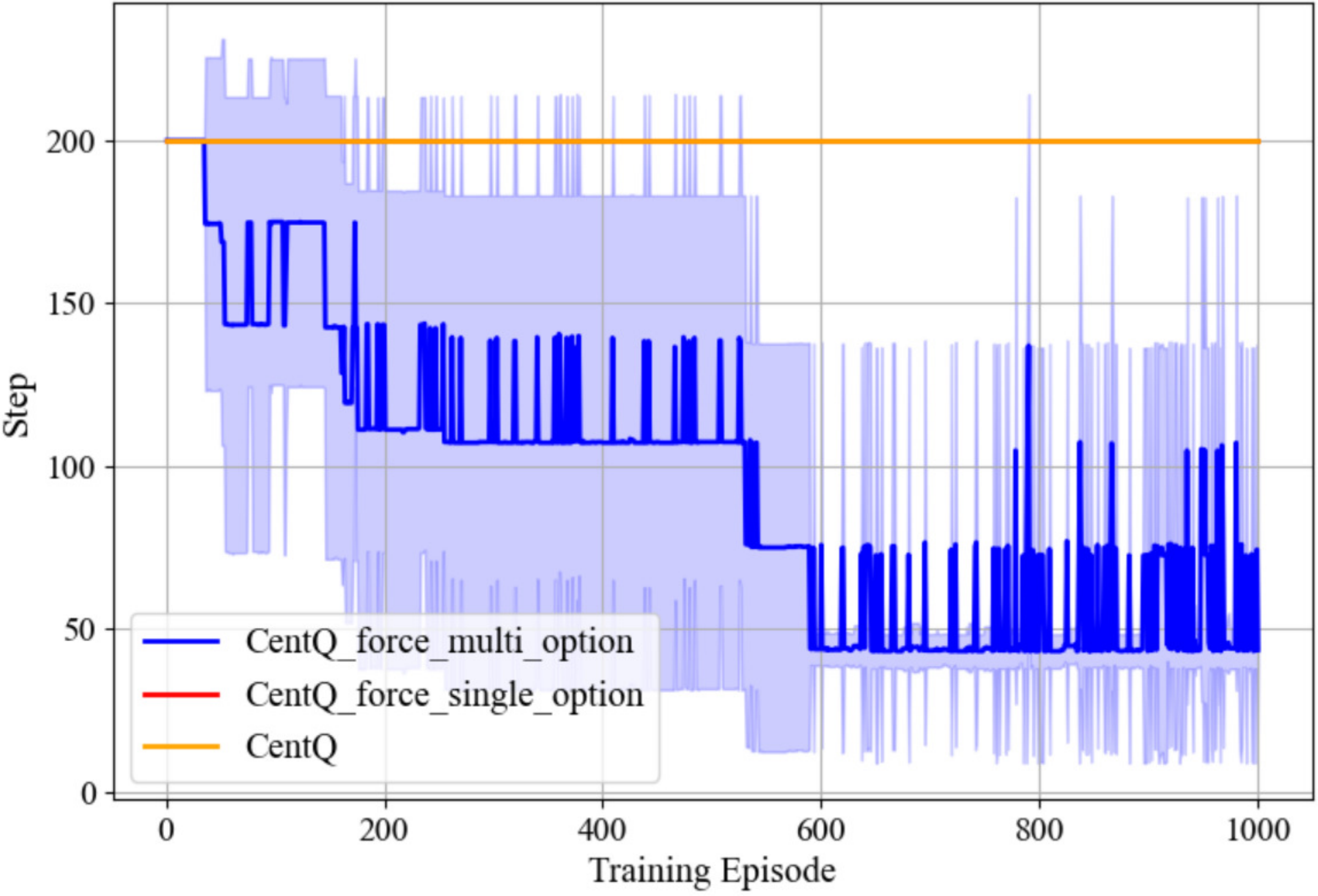}}
\caption{Comparisons on the $m \times n$ Grid Room tasks: (a)(b) Distributed Q-Learning; (c)(d) Centralized Q-Learning + Force.}
\label{fig:a2} %% label for entire figure
\end{figure}

\vspace{-.1in}

\subsection{$n$-agent Grid Room Task with random grouping} \label{pair-room}

\begin{figure}[htbp]
\centering
\subfigure[Grid Room with 4 agents]{
\label{fig:a3(a)} 
\includegraphics[width=2.2in, height=1.0in]{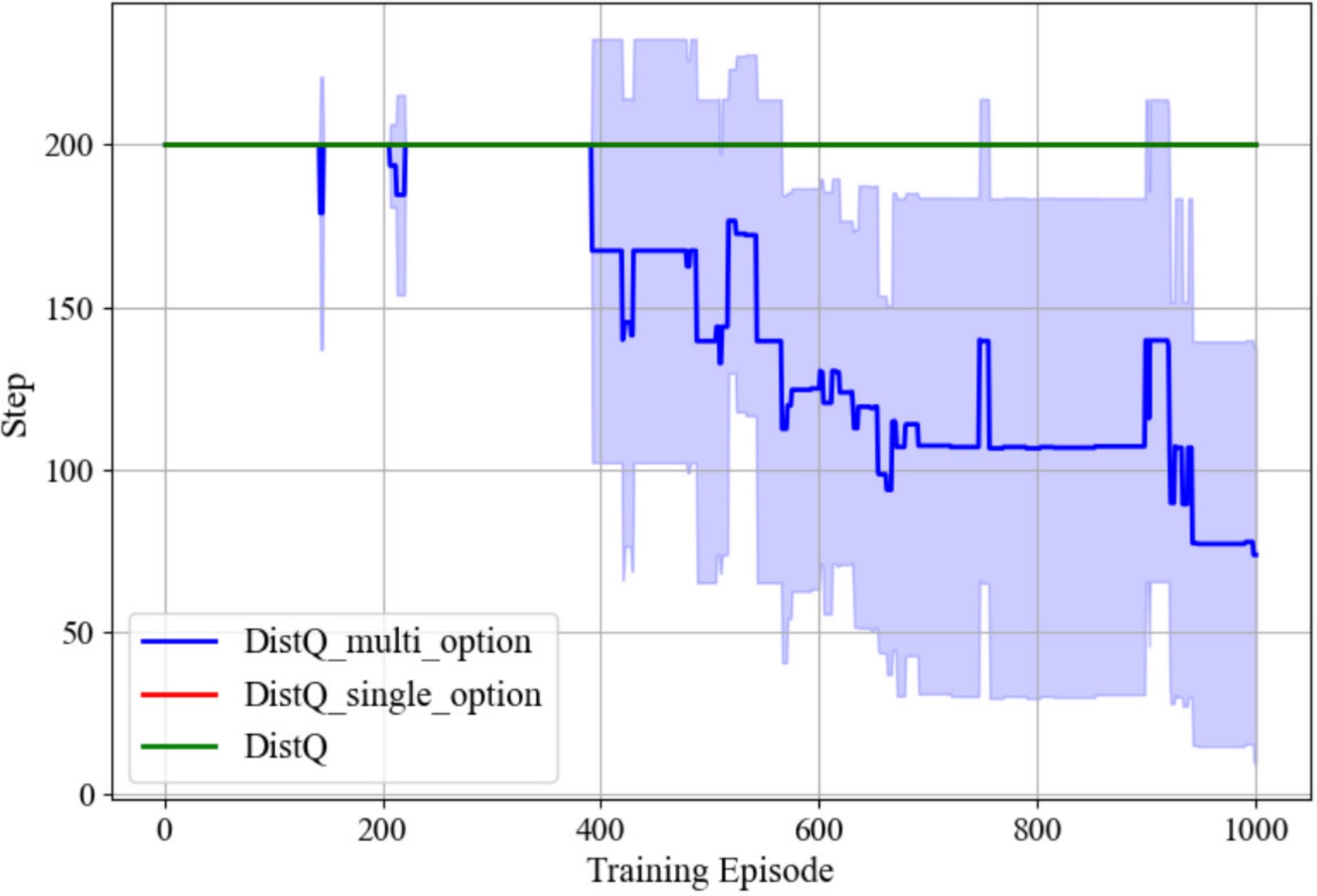}}
\subfigure[Grid Room with 6 agents]{
\label{fig:a3(b)} 
\includegraphics[width=2.2in, height=1.0in]{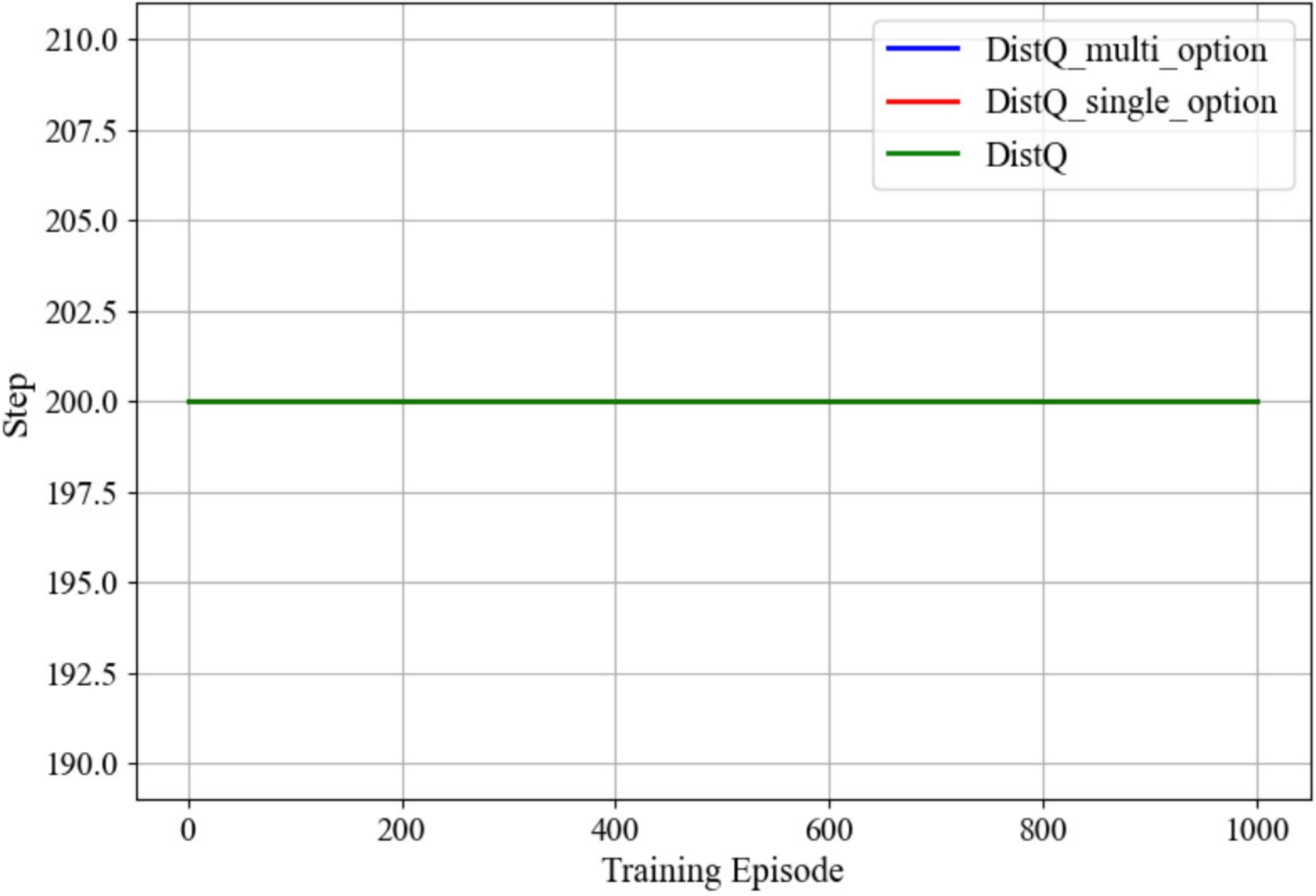}}
\subfigure[Grid Room with 4 agents]{
\label{fig:a3(c)} 
\includegraphics[width=2.2in, height=1.0in]{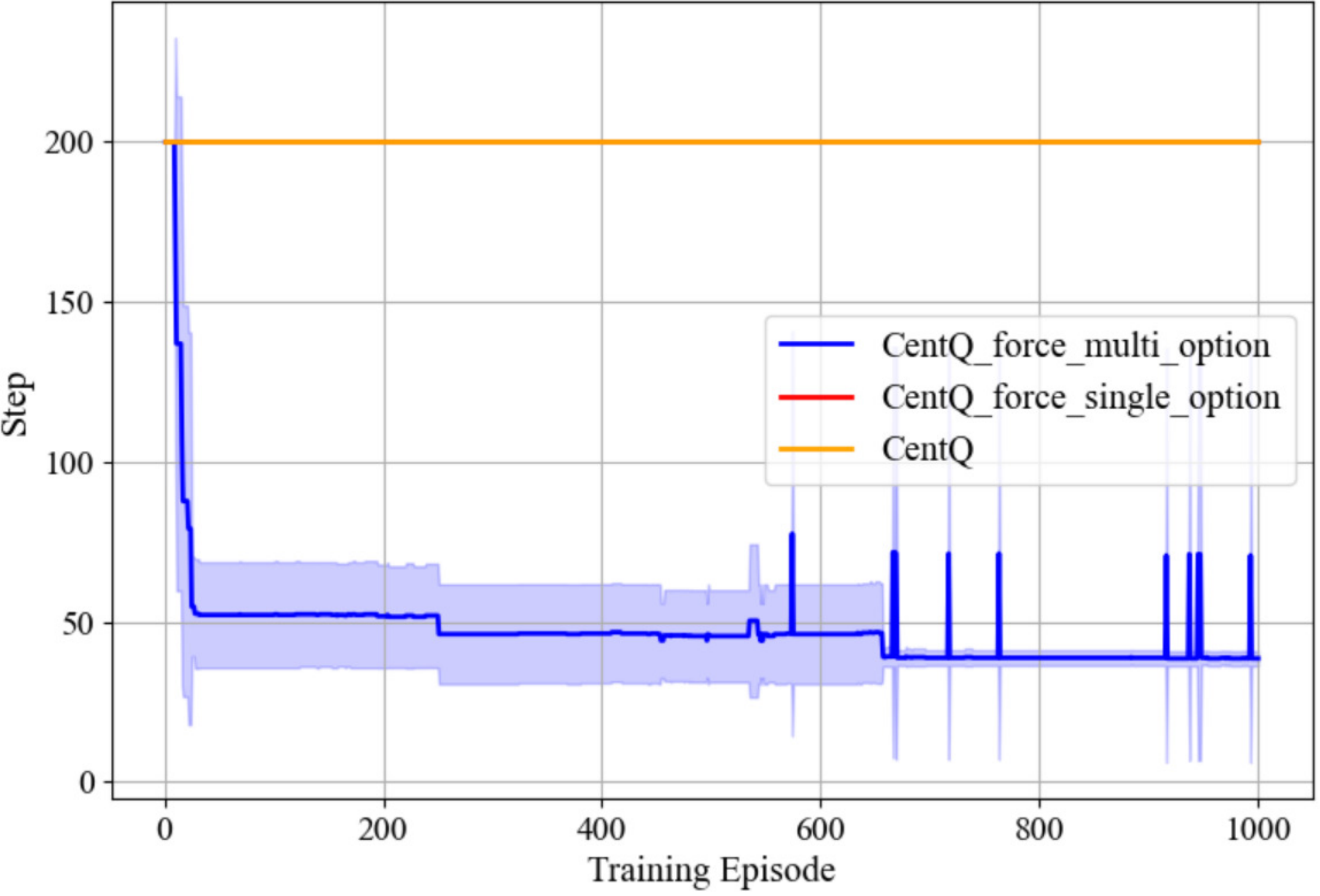}}
\subfigure[Grid Room with 6 agents]{
\label{fig:a3(d)} 
\includegraphics[width=2.2in, height=1.0in]{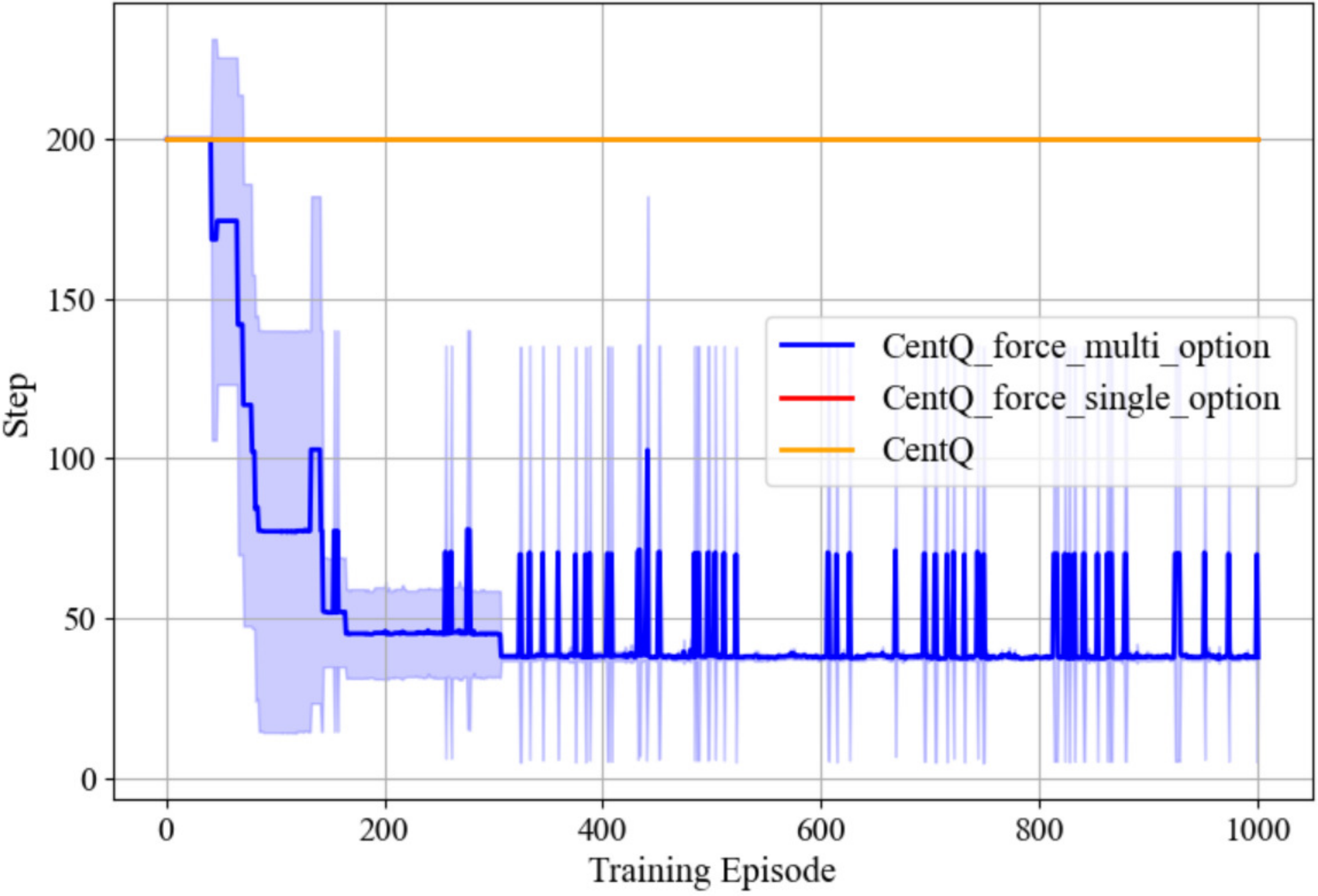}}
\caption{Comparisons on the $n$-agent Grid Room tasks with random grouping: (a)(b) Distributed Q-Learning; (c)(d) Centralized Q-Learning + Force.}
\label{fig:a3} %% label for entire figure
\end{figure}

\subsection{A quantitative study on the approximation error of the joint transition graph with Kronecker-product approximation} \label{QS}

\begin{wrapfigure}{r}{3.3cm}
    \vspace{-.15in}
	\centering
 \includegraphics[height=0.85in, width=0.95in]{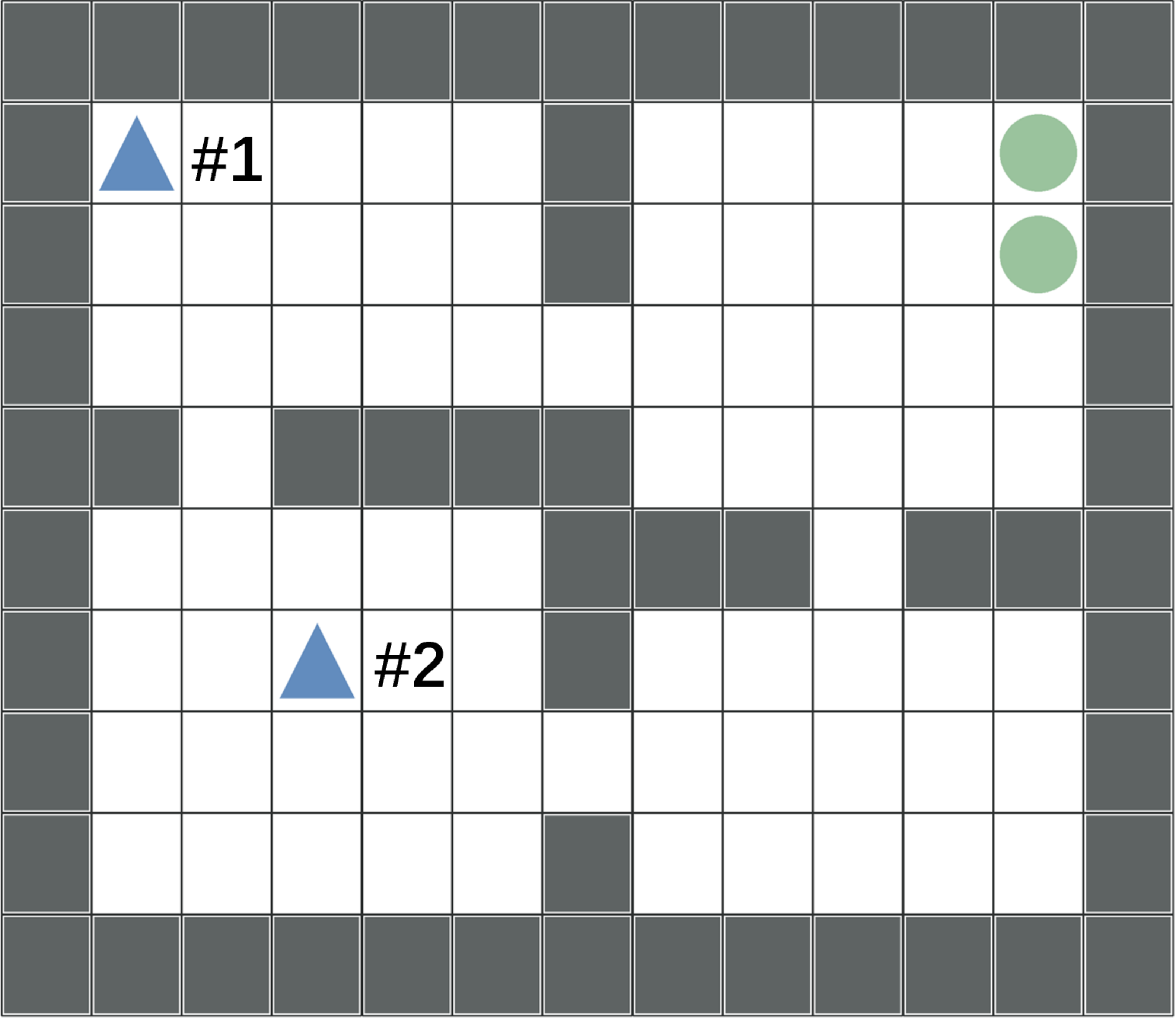}
	\caption{Simulator}
	\vspace{-.1in}
	\label{fig:qs2}
	\vspace{-.1in}
\end{wrapfigure}

In this section, we evaluate the approximation error when we use $\otimes_{i=1}^{n}G_{i}$ as a factorized approximation of $\widetilde{G}$, regarding option discovery. We test on a simplified Grid Room task shown as Figure \ref{fig:qs2}, where two agents are represented as triangles and the goal area is labelled as circles. The time complexity to compute the groundtruth of the Laplacian spectrum of the joint state transition graph is cubic with the number of the joint states which grows exponentially with the number of agents. For example, there are 74 states for each agent in Figure 8, and the computation complexity is already $\mathcal{O}(10^{11})$ (i.e., $(74^2)^3$). 

\begin{figure}[t]
\centering
\subfigure[Eigenvalue when $\alpha$=0.3]{
\label{fig:qs1(b)} 
\includegraphics[width=1.75in, height=1.1in]{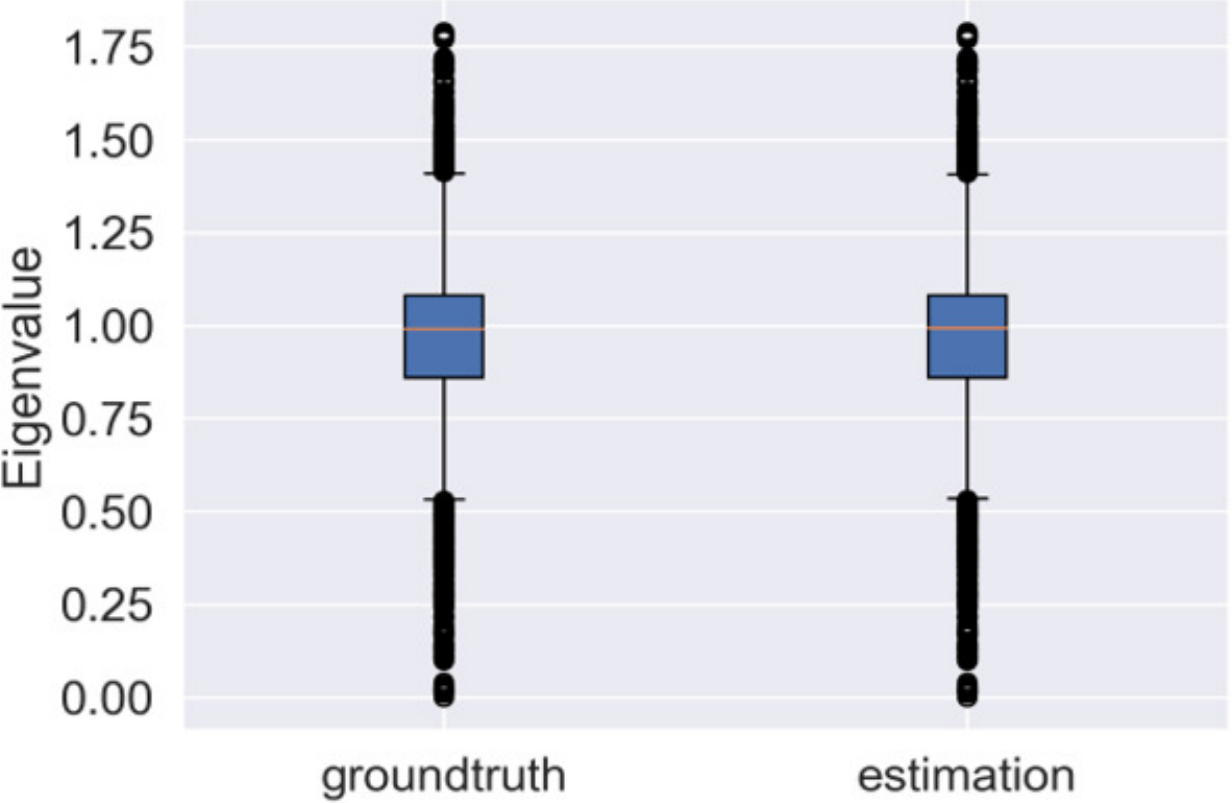}}
\subfigure[Fielder when $\alpha$=0.3, $\rho$=0.5]{
\label{fig:qs1(c)} 
\includegraphics[width=1.75in, height=1.1in]{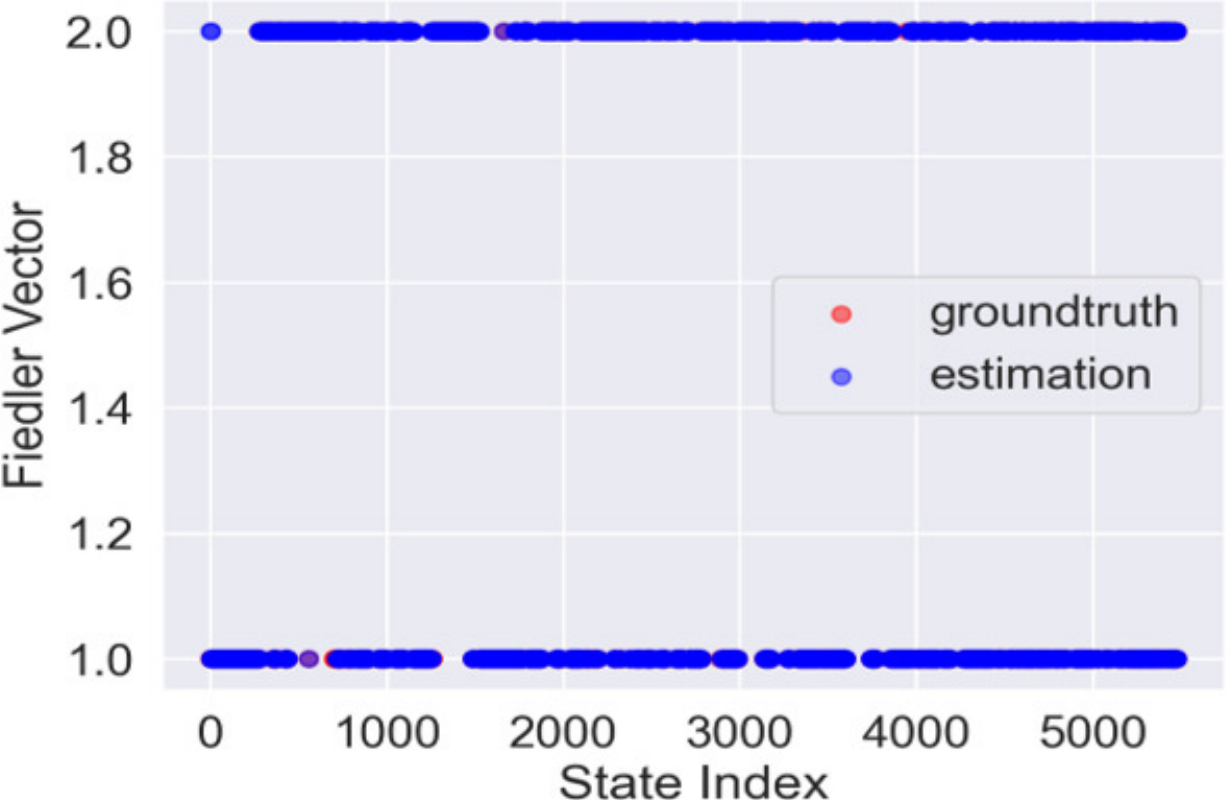}}
\subfigure[Fielder when $\alpha$=0.3, $\rho$=0.2]{
\label{fig:qs1(d)} 
\includegraphics[width=1.75in, height=1.1in]{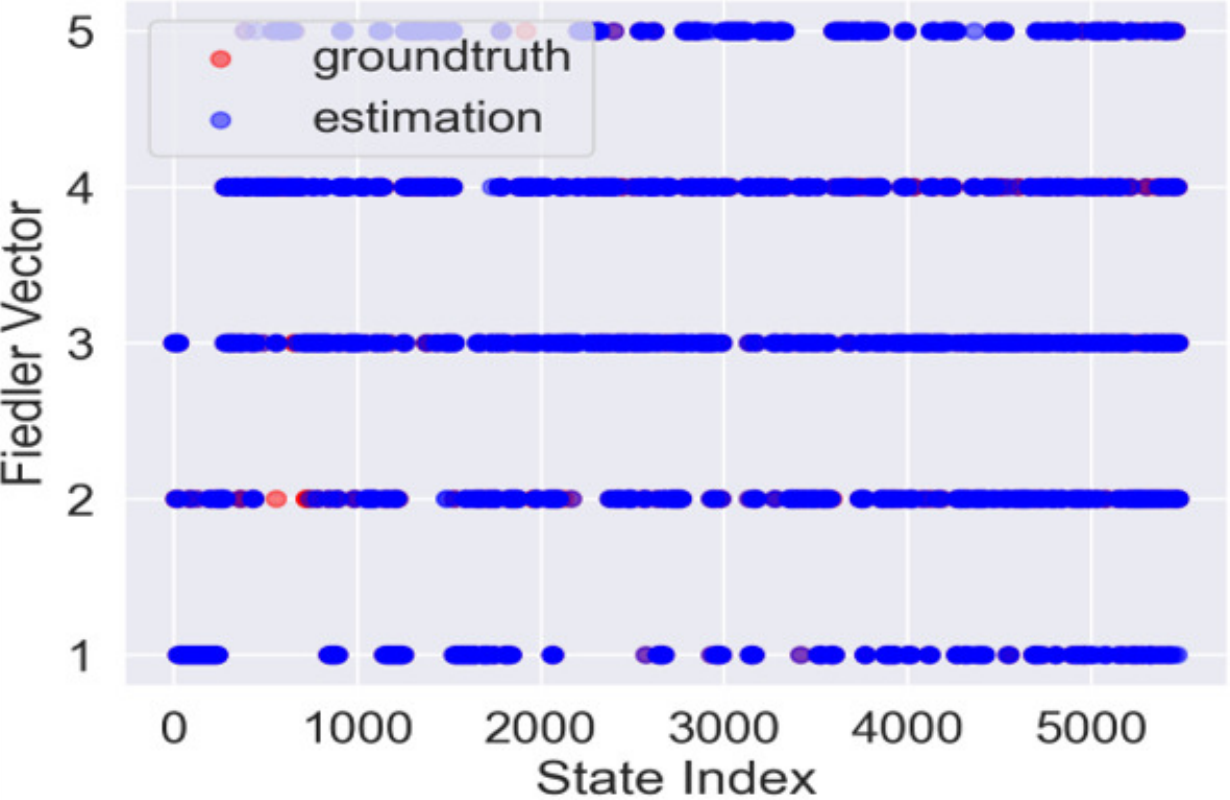}}
\subfigure[Eigenvalue when $\alpha$=0.5]{
\label{fig:qs1(e)} 
\includegraphics[width=1.75in, height=1.1in]{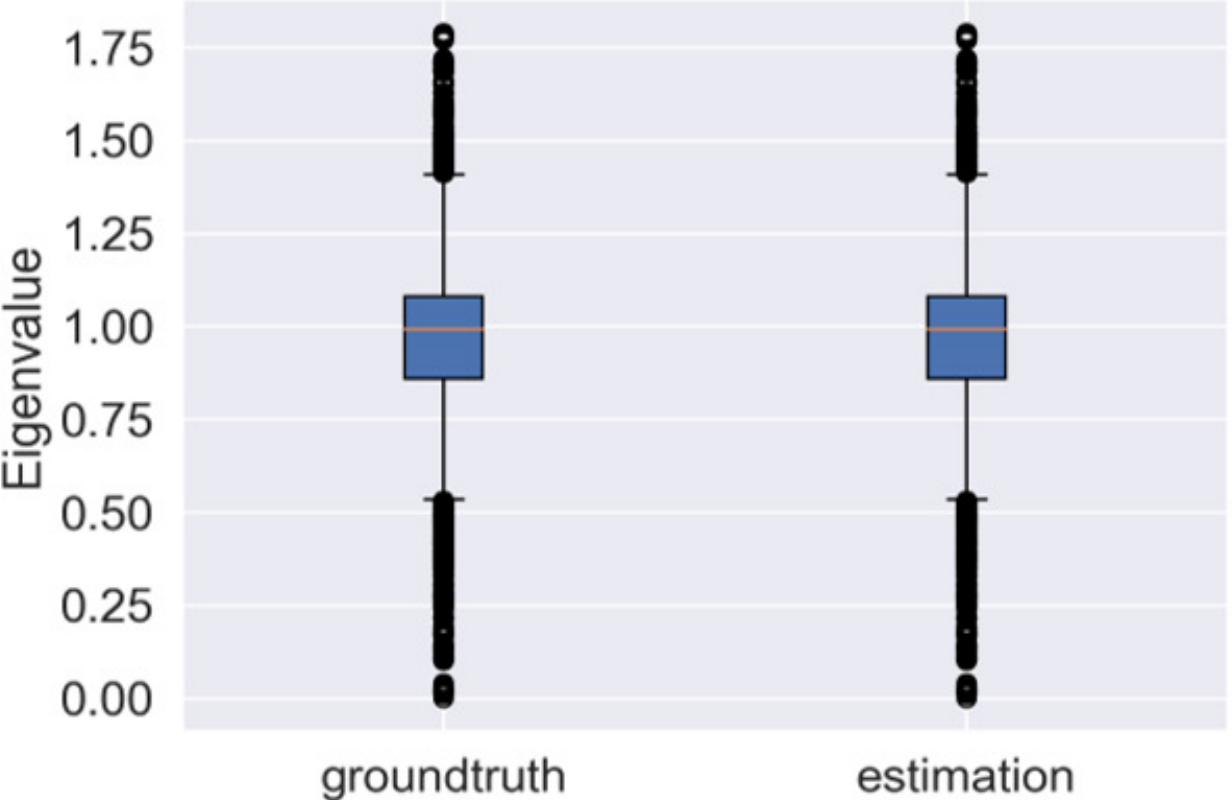}}
\subfigure[Fielder when $\alpha$=0.5, $\rho$=0.5]{
\label{fig:qs1(f)} 
\includegraphics[width=1.75in, height=1.1in]{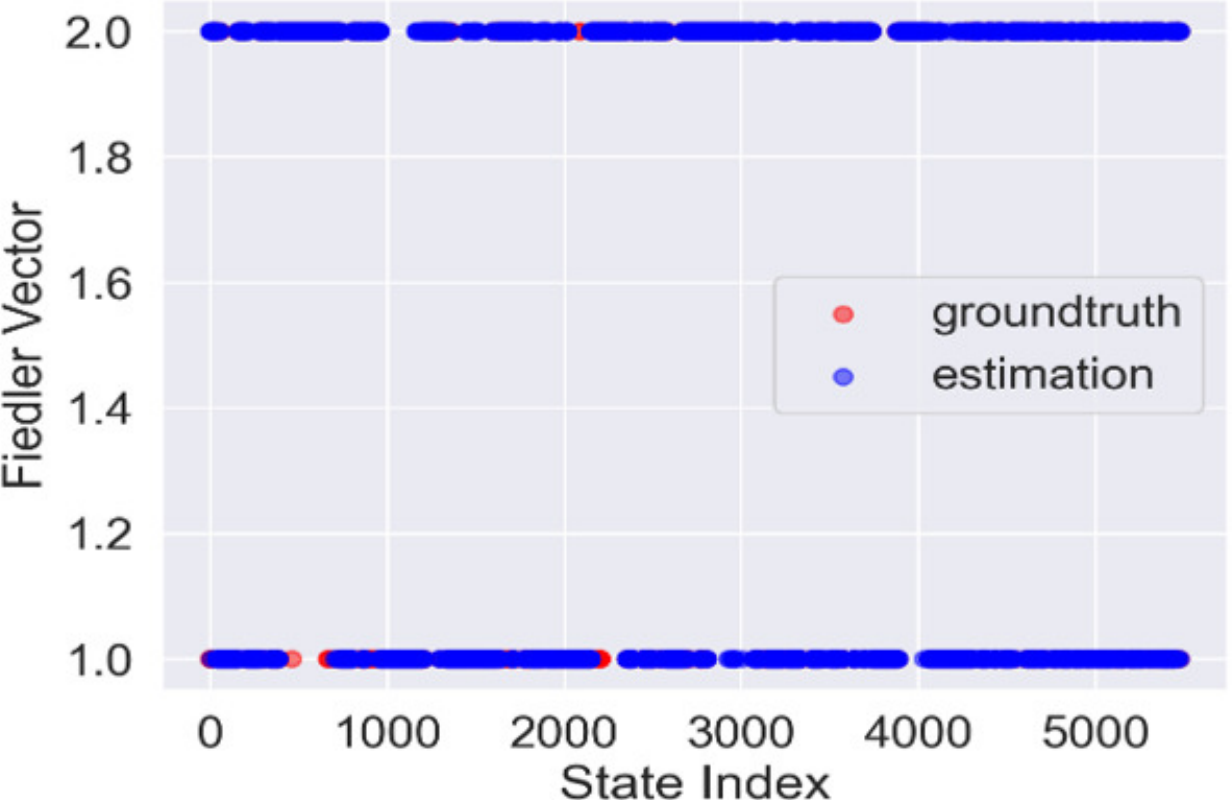}}
\subfigure[Fielder when $\alpha$=0.5, $\rho$=0.2]{
\label{fig:qs1(g)} 
\includegraphics[width=1.75in, height=1.1in]{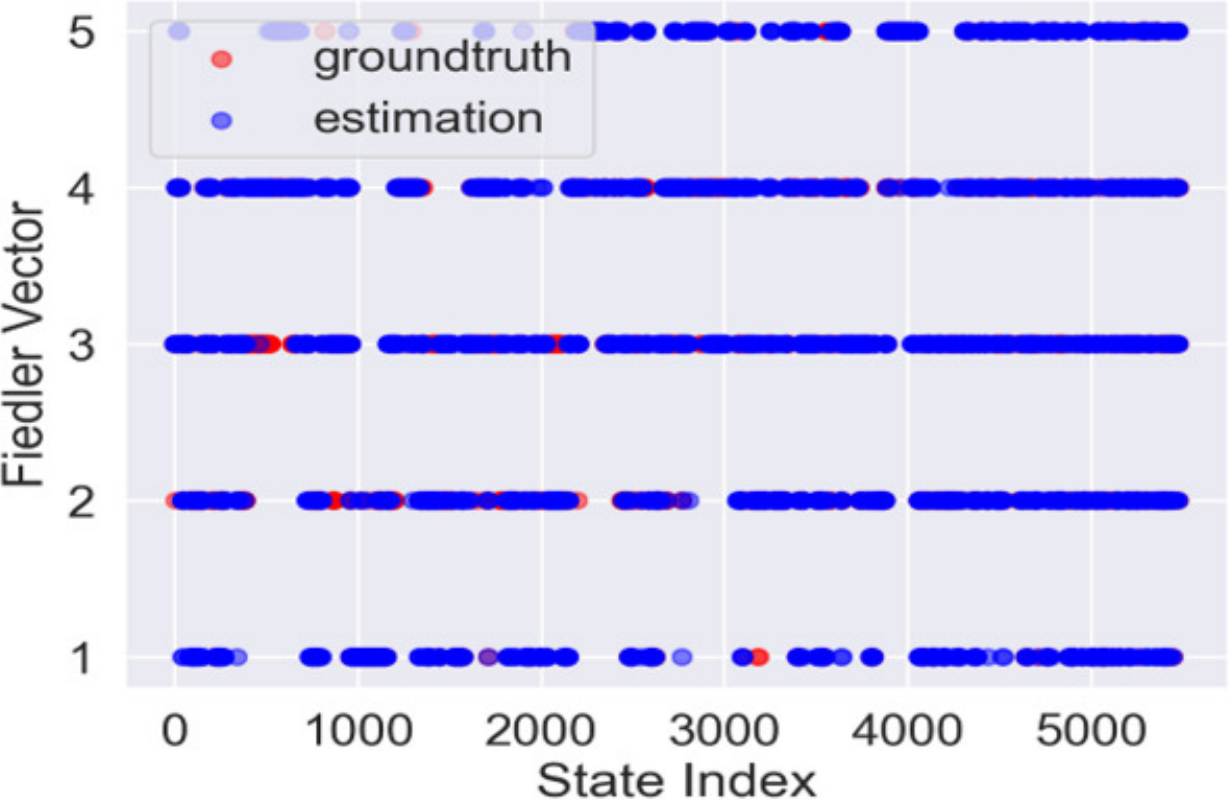}}
\subfigure[Eigenvalue when $\alpha$=0.7]{
\label{fig:qs1(h)} 
\includegraphics[width=1.75in, height=1.1in]{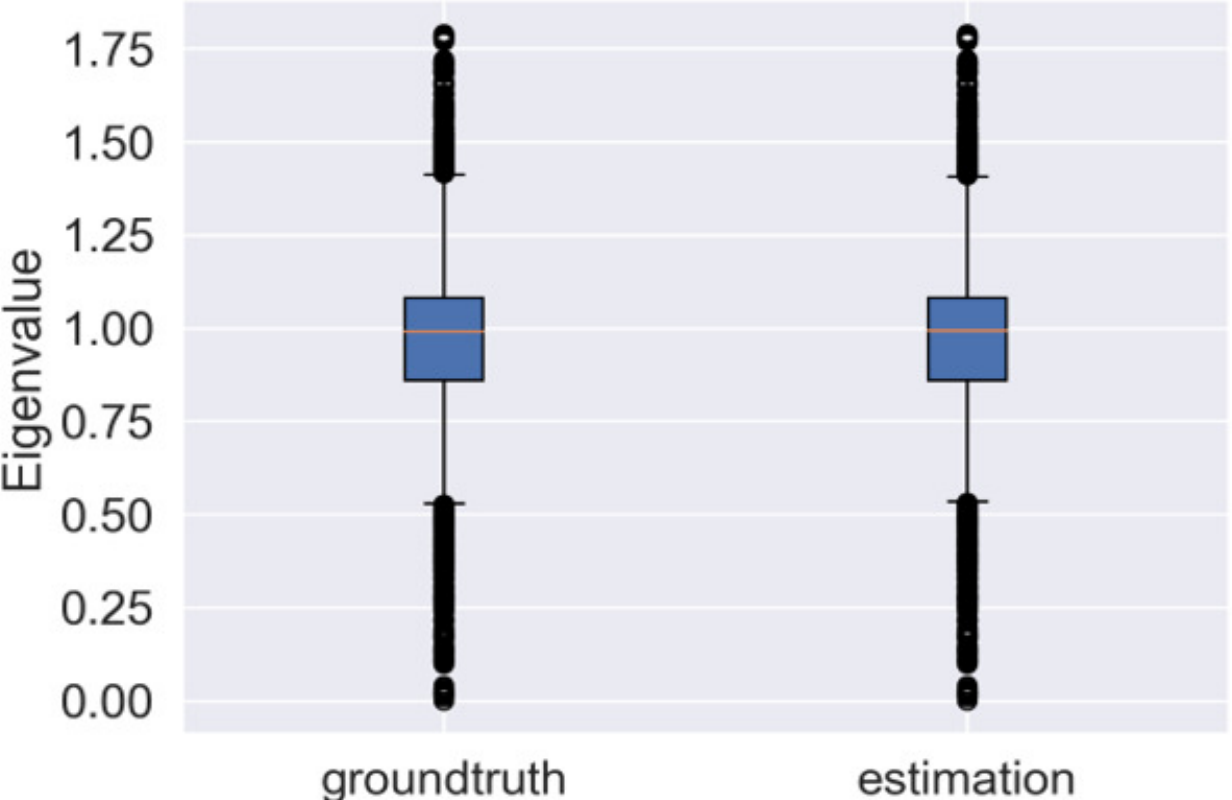}}
\subfigure[Fielder when $\alpha$=0.7, $\rho$=0.5]{
\label{fig:qs1(i)} 
\includegraphics[width=1.75in, height=1.1in]{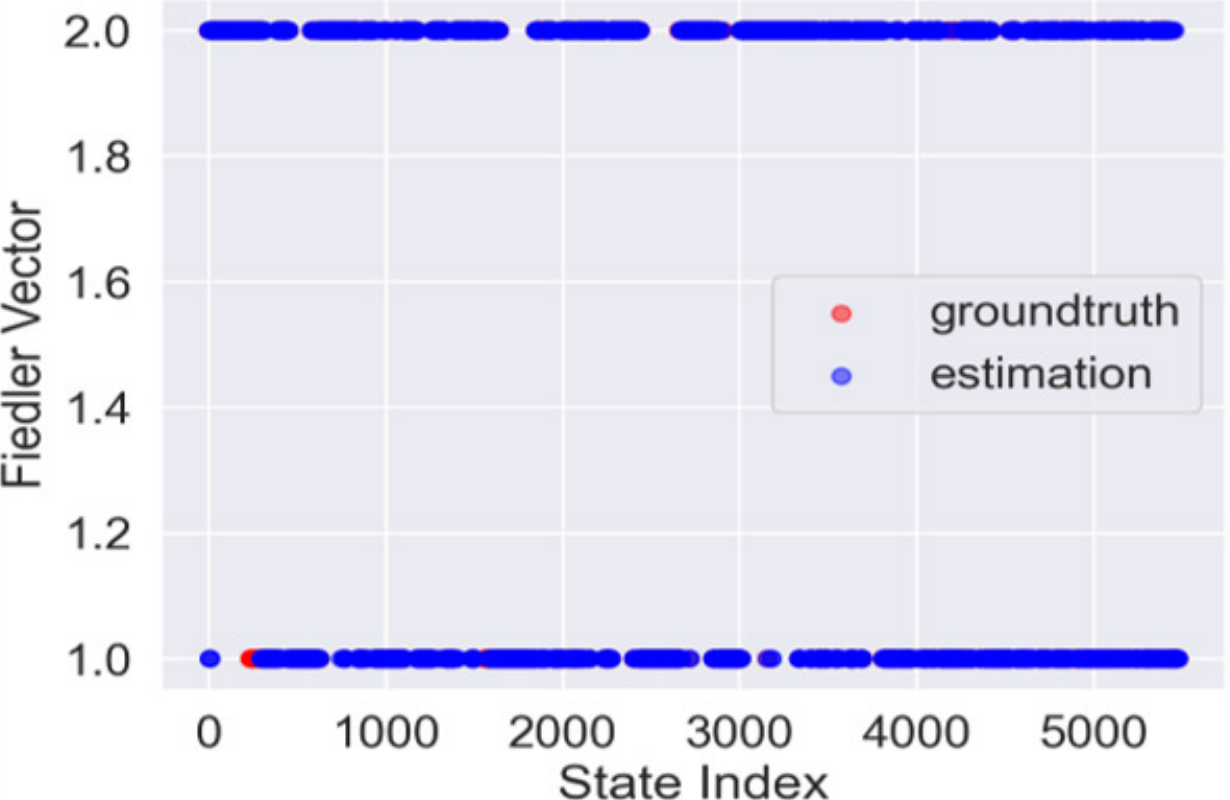}}
\subfigure[Fielder when $\alpha$=0.7, $\rho$=0.2]{
\label{fig:qs1(j)} 
\includegraphics[width=1.75in, height=1.1in]{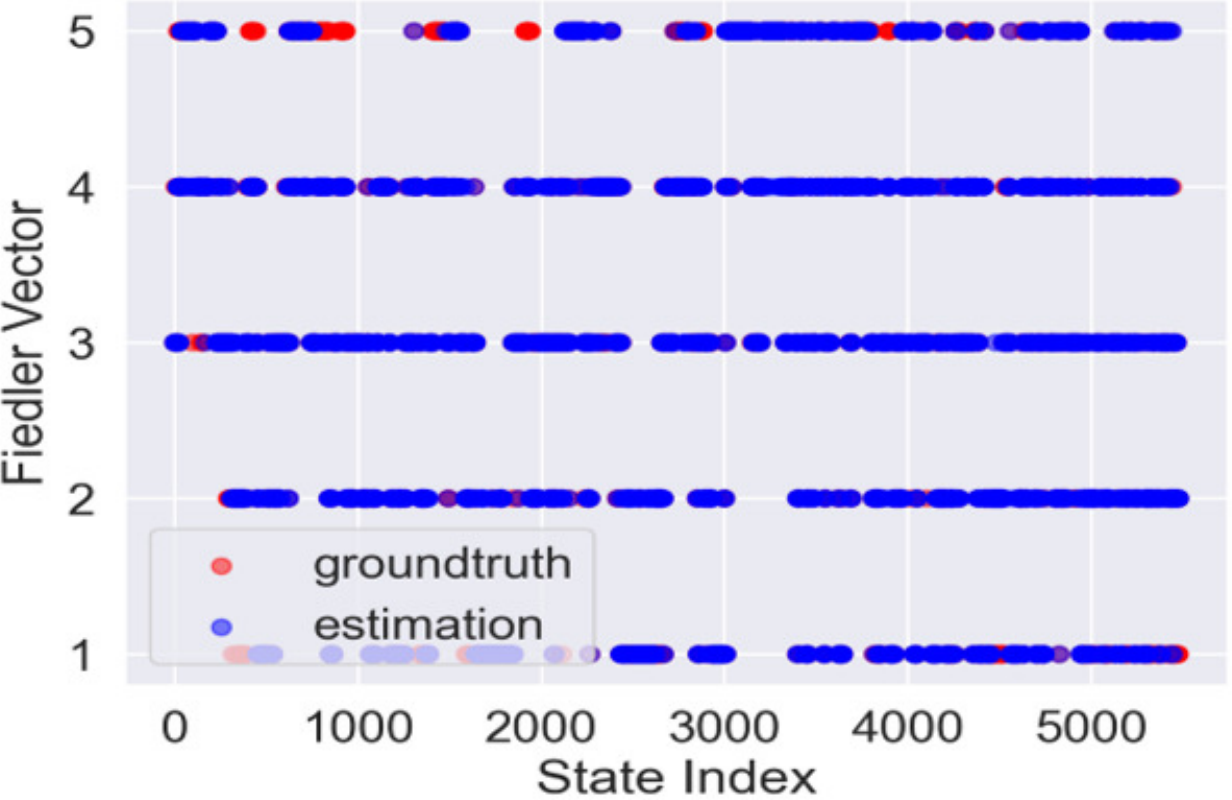}}
\subfigure[Eigenvalue when $\alpha$=0.9]{
\label{fig:qs1(k)} 
\includegraphics[width=1.75in, height=1.1in]{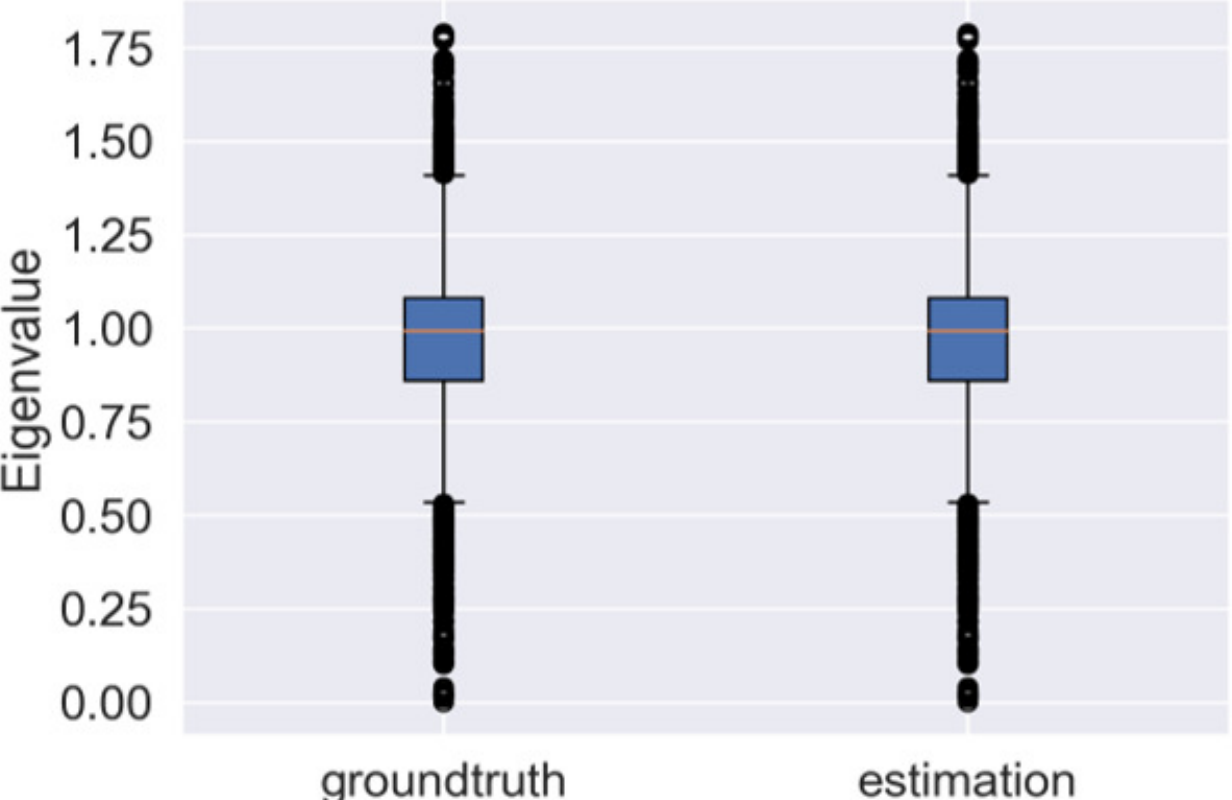}}
\subfigure[Fielder when $\alpha$=0.9, $\rho$=0.5]{
\label{fig:qs1(l)} 
\includegraphics[width=1.75in, height=1.1in]{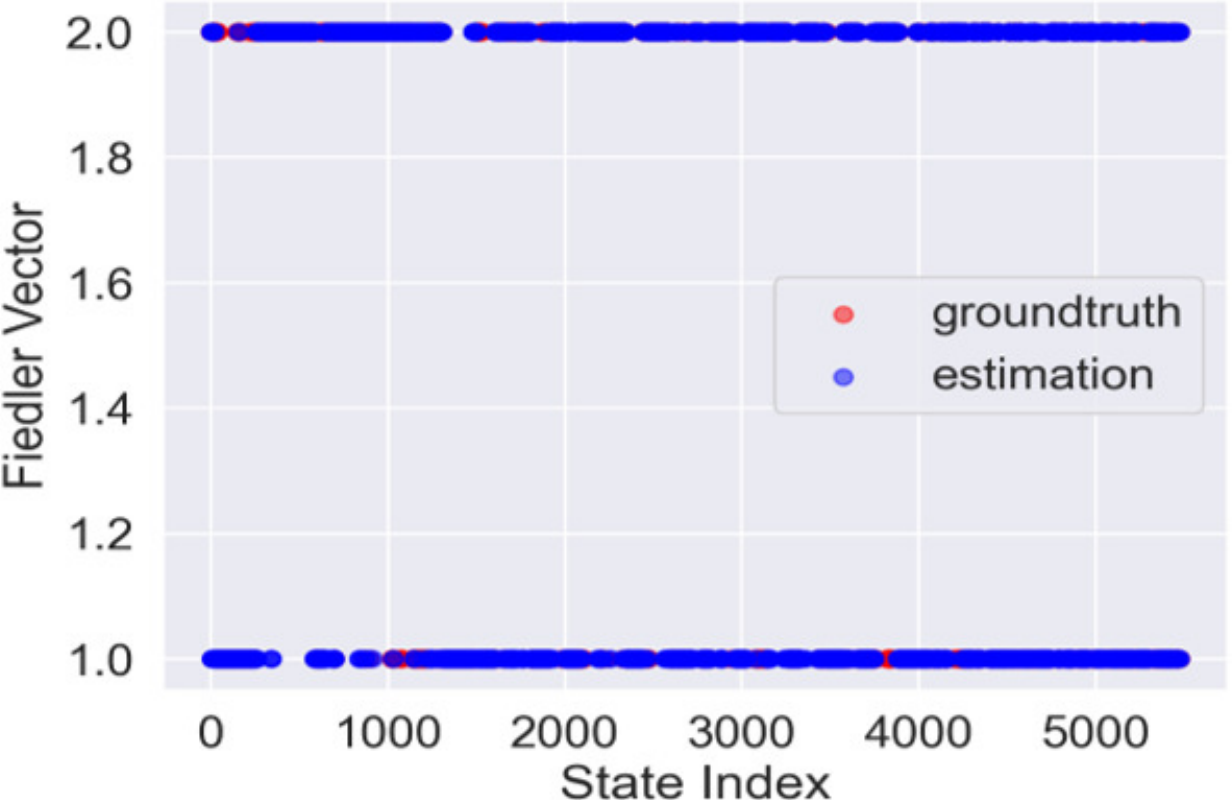}}
\subfigure[Fielder when $\alpha$=0.9, $\rho$=0.2]{
\label{fig:qs1(m)} 
\includegraphics[width=1.75in, height=1.1in]{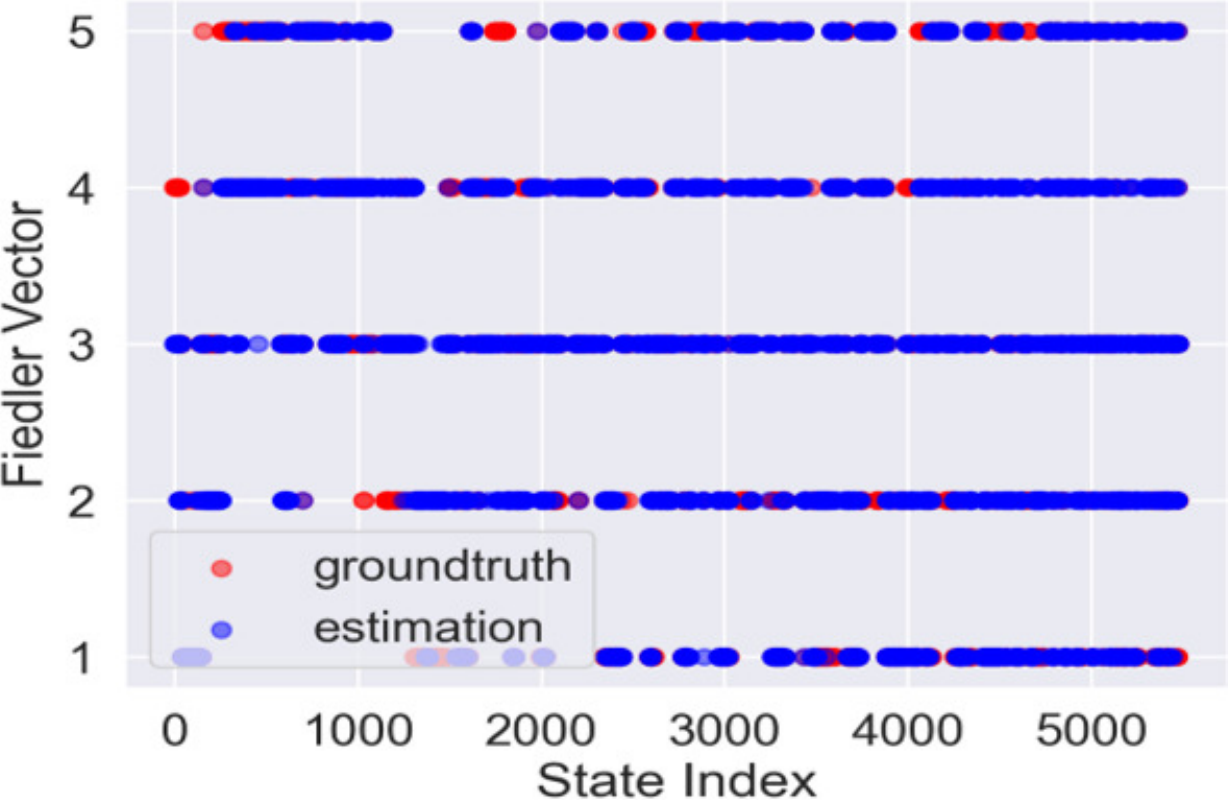}}
\caption{Comparison between the groundtruth and estimation of the Laplacian spectrum of the joint state transition graph as transition influence increases. The first column shows the distribution of the eigenvalues, from which we can see the distribution of the estimated eigenvalues is very close to the groundtruth. The second column shows the Fiedler vector on the joint state space, where we partition the states into 2 clusters (i.e., $\rho=0.5$) and the states with the value in the Fiedler vector that is lower than the median is labelled as 1 and the others are labelled as 2. Similarly, in the third column, the states are partitioned into 5 clusters (i.e., $\rho=0.2$), where the value of the states labeled as $i$ is between the $(i-1)$-th and $i$-th quintile of the values in the Fielder vector. In the third column, the number of the unmatched groundtruth (i.e., red points) goes up as $\alpha$ increases, showing that approximation error increases with $\alpha$.}
\label{fig:qs1} 
\end{figure}

\begin{table}[t]
\centering
\begin{tabular}{|c||c|c|c|c|}
\hline
$\alpha$ & 0.3 & 0.5 & 0.7 & 0.9 \\
\hline
\hline
Algebraic Connectivity ($\times10^{-3}$) & 8.0988 & 8.1153 & 8.0996 & 7.9763 \\
\hline
\makecell[c]{Estimation Accuracy of \\
Fielder when $\rho$=0.5 (\%)}
 & 100 & 100 & 100 & 100 \\  
\hline
\makecell[c]{Estimation Accuracy of \\
Fielder when $\rho$=0.2 (\%)} & 99.9 & 96.2 & 89.8 & 79.4 \\
\hline
\end{tabular}
\caption{Numeric results on the groundtruth of the algebraic connectivity and accuracy of the Fielder estimation, as the transition influence increases.}
\label{table:1}
\end{table}

As mentioned in Section \ref{theory}, the approximation error occurs when the state transitions of an agent are influenced by others. However, the state transition influence among agents, e.g., collisions and blocking, would most likely result in local perturbations of the transition graph and thus is inconsequential to global properties of $\widetilde{G}$. Therefore, approximating $\widetilde{G}$ by $\otimes_{i=1}^{n}G_{i}$ allows efficient option discovery. In Figure \ref{fig:5}, we have evaluated on the case where an agent's state transitions will be influenced by the others' states (i.e., blocking by other agents when going ahead). However, the transition influence for an agent may also come from the action choices of the other agents. Thus, in this scenario (i.e., Figure \ref{fig:qs2}), we set Agent \#1 as the leading agent and Agent \#2 will follow the moving direction of Agent \#1 with the probability $\alpha$, so the state transition of Agent \#2 can be influenced by the action choice of Agent \#1. With a certain $\alpha$, we collect a million state transitions (i.e., $\{(s, a, s')\}$) through Monte Carlo sampling, based on which we can build the joint state transition graph $\widetilde{G}$ and the individual state transition graphs $G_{i}$ ($i=1, 2$) and then get $\otimes_{i=1}^{2}G_{i}$. 

As shown in Figure \ref{fig:qs1} and Table \ref{table:1}, we set $\alpha$ as 0.3, 0.5, 0.7, and 0.9, respectively, to show the approximation error as the transition influence goes up. For the covering option discovery, we only care about the Laplacian spectrum of the state transition graph, especially the algebraic connectivity and Fielder vector. We validate through experiments that the approximation error on the algebraic connectivity and Fielder vector caused by the transition influence among the agents is minor, and thus we can still accurately identify multi-agent options.

In the first column of Figure \ref{fig:qs1}, we visualize the distribution of the eigenvalues corresponding to the Laplacian matrix of $\widetilde{G}$ (i.e., groundtruth) and $\otimes_{i=1}^{2}G_{i}$ (i.e., estimation). It can be observed that the estimated distribution is very close to the groundtruth. Further, we show the algebraic connectivity of $\widetilde{G}$ when setting $\alpha$ as 0.3, 0.5, 0.7, 0.9 in Table \ref{table:1}. The algebraic connectivity of our estimation $\otimes_{i=1}^{2}G_{i}$ is $8.1131\times10^{-3}$ (invariant to $\alpha$), which is close to the groundtruth values.

In the second and third column of Figure \ref{fig:qs1}, we compare the estimated Fiedler vector with the groundtruth. As mentioned in Section \ref{theory}, we only need to identify areas in the state space with relatively low or high values in the Fielder vector and connect them with options. Thus, we partition the states into 2 clusters (i.e., $\rho=1/2=0.5$) according to the median of the values in the Fiedler vector, or partition them into 5 clusters (i.e., $\rho=1/5=0.2$) based on the quintile. We use the Fiedler vector of $\widetilde{G}$ as the groundtruth and compare it with the estimated Fiedler vectors, by comparing the label (i.e., which cluster it belongs to) of each state. It can be observed that the number of the unmatched groundtruth (shown as red points) increases with $\alpha$. Further, we note that the states with the lowest or highest value in the Fiedler vector (i.e., $MIN$ or $MAX$) are the subgoals based on which we define the options, so the estimation accuracy of these states are directly related to the option discovery. In Table \ref{table:1}, we show the estimation accuracy of the subgoals. The third row corresponds to defining the states with the lowest or highest 20\% values in the Fielder vector as the subgoals, which is also the setup we use for option discovery. It can be observed that even if in conditions where the state transition influence is heavy (i.e., $\alpha=0.9$), we can still estimate about 80\% of the subgoals correctly and build options toward them accordingly. 

These results empirically validate our statement that approximating $\widetilde{G}$ with $\otimes_{i=1}^{n}G_{i}$ allows efficient option discovery in cases where transition influence exists. We will consider a theoretical characterization of the impact of approximation errors in the future work.

% However, it will be a good extension for future work to further reduce the difference between $\widetilde{G}$ and $\otimes_{i=1}^{n}G_{i}$ and estimate the subgoals more accurately in scenarios where the state transitions of an agent can be seriously influenced by the others.

\end{document}